\documentclass[10pt,journal,compsoc]{IEEEtran}

\usepackage{tabularx,ragged2e}
\usepackage{booktabs}
\usepackage[hidelinks]{hyperref}
\hypersetup{
    colorlinks,
    citecolor=black,
    filecolor=black,
    linkcolor=black,
    urlcolor=black
}

\usepackage{times}
\usepackage{epsfig}
\usepackage{graphicx}
\usepackage{amsmath}
\usepackage{amssymb}
\usepackage{tabulary}
\usepackage{multirow}
\usepackage[table]{xcolor}
\usepackage{epstopdf}
\usepackage[T1]{fontenc}
\usepackage[utf8]{inputenc}
\usepackage{xspace}
\usepackage[font=small,skip=2pt]{caption}
\usepackage{float}

\newcommand*{\eg}{e.g.\@\xspace}
\newcommand*{\ie}{i.e.\@\xspace}
\newcommand*{\etal}{et al.\@\xspace}

\newcommand{\kdtree}{$k$d-tree~}
\newcommand{\ptdn}{\mbox{$\mathbf{X}$}}

\newcommand{\nspacet}{\mbox{$\mathbf{\Psi}$}}
\newcommand{\nspace}{\mbox{$\mathbf{\psi}$}}

\newcommand{\rom}[1]{\textup{\uppercase\expandafter{\romannumeral#1}}
\newcommand*\samethanks[1][\value{footnote}]{\footnotemark[#1]}}

\newcolumntype{Y}{>{\centering\arraybackslash}X}
\newcolumntype{s}{>{\hsize=.1\hsize}Y}
\newcolumntype{t}{>{\hsize=.2\hsize}Y}
\newcolumntype{u}{>{\hsize=.1\hsize}X}

\DeclareMathOperator*{\argmin}{arg\,min}
\DeclareMathOperator*{\argmax}{arg\,max}

\def\footnoterule{\relax%
  \kern-5pt
  \hbox to \columnwidth{\vrule width 0.5\columnwidth height 0.4pt\hfill}
  \kern4.6pt}
\makeatother

\begin{document}

\title{A Dual-Source Approach for 3D Human Pose Estimation from a Single Image}

\author{Umar~Iqbal*\thanks{*authors contributed equally},
        Andreas~Doering*,
        Hashim Yasin, 
        Bj{\"o}rn Kr{\"u}ger,
        Andreas Weber,
        and Juergen~Gall
\IEEEcompsocitemizethanks{\IEEEcompsocthanksitem Umar Iqbal, Andreas Doering and Juergen Gall are 
with the Computer Vision Group, University of Bonn, Germany. 
Email: 
\href{mailto:uiqbal@iai.uni-bonn.de}{uiqbal@iai.uni-bonn.de}, 
\href{mailto:s6andoer@uni-bonn.de}{s6andoer@uni-bonn.de}, 
\href{mailto:gall@iai.uni-bonn.de}{gall@iai.uni-bonn.de}, 
\protect}
\IEEEcompsocitemizethanks{\IEEEcompsocthanksitem \vspace{-2mm}Hashim Yasin is with the National University of Science and Technology, Pakistan 
Email: \href{mailto:hashim.yasin@nu.edu.pk}{hashim.yasin@nu.edu.pk},
\protect}
\IEEEcompsocitemizethanks{\IEEEcompsocthanksitem \vspace{-2mm}Bj{\"o}rn Kr{\"u}ger is with the Gokhale Method Institute, Stanford, USA.
Email: \href{mailto:kruegerb@cs.uni-bonn.de}{kruegerb@cs.uni-bonn.de},
\protect}
\IEEEcompsocitemizethanks{\IEEEcompsocthanksitem \vspace{-2mm}Andreas Weber is with the Multimedia, Simulation, Virtual Reality Group, University of Bonn, Germany. 
Email: \href{mailto:weber@cs.uni-bonn.de}{weber@cs.uni-bonn.de}.
\protect}

}

\markboth{Submitted to Computer Vision and Image Understanding.}%
{Shell \MakeLowercase{\textit{et al.}}: A Dual-Source Approach for 3D Human Pose Estimation from Single Images}

\IEEEtitleabstractindextext{
\begin{abstract}
In this work we address the challenging problem of 3D human pose estimation from single images. Recent approaches learn deep neural networks to regress
3D pose directly from images. One major challenge for such methods, however, is the collection of training data. 
Specifically, collecting large amounts of training data containing unconstrained images annotated with 
accurate 3D poses is infeasible. We therefore propose to use two independent training sources. The first source consists 
of accurate 3D motion capture data, and the second source consists of unconstrained images with annotated 2D poses. 
To integrate both sources, we propose a dual-source approach that combines 2D pose estimation with efficient 
3D pose retrieval. To this end, we first convert the motion capture data into a normalized 2D pose space, and separately learn a 
2D pose estimation model from the image data. During inference, we estimate the 2D pose and efficiently retrieve the nearest 3D poses.
We then jointly estimate a mapping from the 3D pose space to the image and reconstruct the 3D pose. We provide a comprehensive evaluation of the proposed method 
and experimentally demonstrate the effectiveness of our approach, even when the skeleton structures of the two sources differ substantially. 
\end{abstract}

}

\maketitle

\IEEEdisplaynontitleabstractindextext
\IEEEpeerreviewmaketitle

\section{Introduction}
3D human pose estimation has a vast range of applications such as virtual reality, human-computer interaction, activity recognition, sports
video analytics, and autonomous vehicles. The problem has traditionally been tackled by utilizing multiple images captured by synchronized cameras 
capturing the person from multiple views~\cite{belagiannis20143d, sigal2012loose, yao2012coupled}. In many scenarios, however, capturing multiple views is infeasible 
which limits the applicability of such approaches. Since 3D human pose estimation from a single image is very difficult due to missing depth information, depth cameras
have been utilized for human pose estimation~\cite{Baak:2011,Shotton:2011,Grest:2005}. However, current depth sensors are also limited to indoor environments and cannot
be used in unconstrained scenarios. Therefore, estimating 3D pose from single, in particular, unconstrained images is a highly relevant task.  

One approach to address this problem is to follow a fully-supervised learning paradigm, where a regression model ~\cite{Bo-2010,h36m_pami, Ilya_2014,ics-cvpr14,Agarwal:2006, bo2008fast,LiC14, tekin2015predicting}
or a deep neural network \cite{Sijin2015iccv,tekin2016structured, tekin2016fusing, zhou2016deep, Moreno_arxiv2016, popa2017CVPRmultitask} 
can be learned to directly regress the 3D pose from single images. This approach, however, requires a large amount of training data where each
2D image is annotated with a 3D pose. In contrast to 2D pose estimation, manual annotation of such training data
is not possible due to ambiguous geometry and body part occlusions. On the other hand, automatic acquisition of accurate 3D pose
for an image requires a very sophisticated setup. The popular datasets like HumanEva~\cite{Sigal_2010} or Human3.6M~\cite{h36m_pami} 
synchronized cameras with a commercial marker-based system to obtain 3D poses for images. This requires a very expensive hardware setup and 
the requirements for marker-based system like studio environment and attached markers limits the applicability of such systems primarily
to indoor laboratory environments. Some recent approaches such as EgoCap~\cite{rhodin2016egocap} allows to capture 3D poses in outdoor environments, but image data in such cases is 
restricted only to ego-centric views of the person. 

\begin{figure*}[t]
\begin{center}
 \includegraphics[width=1\linewidth]{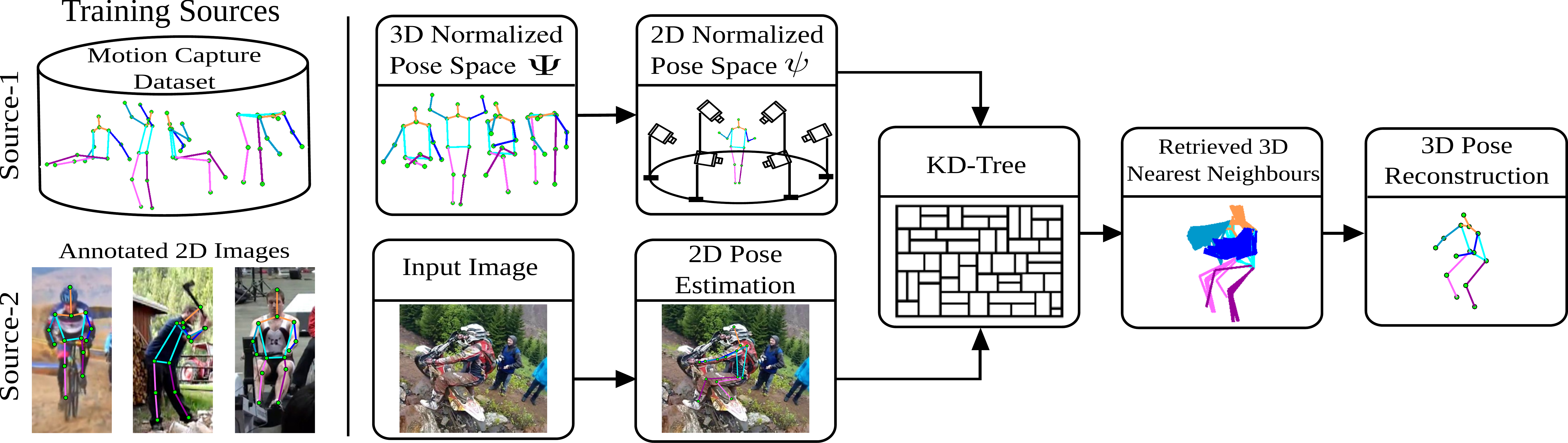}
\end{center}
\caption{
{\bf Overview.} Our approach relies on two training sources. The first source is a motion capture database that contains only 3D poses. 
The second source is an image database with annotated 2D poses. The motion capture data is processed by pose normalization and projecting 
the poses to 2D using several virtual cameras. This gives many 3D-2D pairs where the 2D poses serve as features. The image data is used
to learn a CNN model for 2D pose estimation. Given a test image, the CNN predicts the 2D pose which is then used to retrieve the normalized 
nearest 3D poses. The final 3D pose is then estimated by minimizing the projection error under the constraint that the solution is close to
the retrieved poses. 
}
\label{fig:sysflow}
\end{figure*}

In this work, we propose a dual-source method that does not require training data consisting of pairs of an image and a 3D pose, but rather utilize 2D and
3D information from two independent training sources as illustrated in Fig.~\ref{fig:sysflow}. 
The first source is accurate 3D motion capture data containing a large number of 3D poses, and is captured in a laboratory setup, 
e.g., as in the CMU motion capture dataset~\cite{cmu_mocap} or the Human3.6M dataset \cite{h36m_pami}. Whereas, the second source consists of images with annotated 2D poses as they are provided by 2D human pose datasets, \eg, MPII Human Pose~\cite{andriluka14cvpr},
Leeds Sports Pose~\cite{Johnson10} and MSCOCO~\cite{lin2014microsoft}. 
Since 2D poses can be manually annotated for images, they do not impose any restriction regarding the environment from where the 
images are taken. In fact any image from the Internet can be annotated and used. 
Since both sources are captured independently, we do not know the 3D pose for any training image.
In order to bring the two sources together, we map the motion capture data into a normalized 2D pose space to allow for an efficient retrieval based on 2D body joints. Concurrently, we learn a 2D pose estimation model from the 2D images based on convolutional neural networks. During inference, we first estimate the 2D pose and retrieve the nearest 3D poses using an effective approach that is robust 
to 2D pose estimation errors. We then jointly estimate the projection from the 3D pose space to the image and reconstruct the 3D pose. 

A preliminary version of this work was presented in \cite{Yasin_2016_CVPR}. In this work we leverage the recent progress in 2D 
pose estimation \cite{toshev2014deeppose, iqbalFG2017actionpose, carreira2015human, pishchulin2016deepcut, wei2016convolutional, hu2016bottom, 
insafutdinov2016deepercut, newell2016eccv, bulat2016human, georgia2016eccv, rafi2016bmvc, chu2017CVPRmulti}, and improve the performance of \cite{Yasin_2016_CVPR}
by a large margin. We further show that with the availability of better 2D pose estimates, the approach \cite{Yasin_2016_CVPR} can be largely
simplified. We extensively evaluate our approach on two popular datasets for 3D pose estimation namely Human3.6M \cite{h36m_pami} and HumanEva \cite{Sigal_2010}.
On both datasets, our approach performs better or on par with the state-of-the-art. We provide an in-depth analysis of the proposed approach. 
In particular, we analyze the impact of different MoCap datasets, the impact of the similarity of the training and test poses,
the impact of the accuracy of the used 2D pose estimator, and also the differences of the skeleton structure between the two training sources.
Finally, we also provide qualitative results for images taken from the MPII Human Pose dataset \cite{andriluka14cvpr}.

\section{Related Work}

Earlier approaches for 3D human pose estimation from single images~\cite{Bo-2010, AgarwalT-2004,Sminchisescu-2005,Agarwal:2006, bo2008fast, mori2006recovering} 
utilize discriminative methods to learn a mapping from hand-crafted local image features (e.g., HOG, SIFT, etc.) to 3D human pose. 
Since local features are sensitive to noise, \cite{Ilya_2014} proposed an approach based on a 3D pictorial structure model that 
combines generative and discriminative methods to obtain robustness to noise. For this, regression forests are trained to estimate
the probabilities of 3D joint locations and the final 3D pose is inferred by the pictorial structure model. 
Since inference is performed in 3D, the bounding volume of the 3D pose space needs to be known and the inference 
requires a few minutes per frame. In addition to the local image features, the approach~\cite{ics-cvpr14} also utilizes body part segmentation 
with a second order hierarchical pooling process to obtain robust image descriptors. Instead of computing low level image features,
the approach~\cite{Pons-Moll_2014_CVPR} uses boolean geometric relationships between body joints to encode body pose appearance. These features
are then used to retrieve semantically similar poses from a large corpus of 3D poses. 

With the advances in deep learning, more recent approaches learn end-to-end CNNs to regress 
the 3D joint locations directly from the images~\cite{LiC14, Sijin2015iccv,tekin2016structured, rogez2016mocap, chen20163dv,
zhou2016deep, Moreno_arxiv2016, parkECCV2016workshops, tekin2016fusing, popa2017CVPRmultitask}. In this direction, the work~\cite{LiC14} is 
one of the earliest methods that presents an end-to-end CNN architecture, where a multi-task loss 
is proposed to simultaneously detect body parts in 2D images and regress their locations in 3D space. In~\cite{Sijin2015iccv} a max-margin loss is 
incorporated with a CNN architecture to efficiently model joint dependencies. Similarly,~\cite{zhou2016deep} enforces kinematic constraints by introducing 
a differentiable kinematic function that can be combined with a CNN. The approach~\cite{tekin2016structured} uses auto-encoders to incorporate 
dependencies between body joints and combines them with a CNN architecture to regress 3D poses. Approaches for data augmentation 
have also been proposed in~\cite{rogez2016mocap} and \cite{chen20163dv} where synthetic training images are generated to enlarge the training data. 
\cite{Moreno_arxiv2016} proposes to encode 3D pose using an Euclidean distance matrix formulation that implicitly incorporates body joint 
relations and allows to regress 3D poses in form of a distance matrix using a very small network architecture. The approaches \cite{parkECCV2016workshops, tekin2016fusing, popa2017CVPRmultitask}
 leverage the information about the locations of 2D body joints to aid 3D human pose estimation. While~\cite{parkECCV2016workshops} directly uses the 
 2D joint coordinates to regularize the training of a CNN,~\cite{tekin2016fusing, popa2017CVPRmultitask} use confidence scoremaps of 2D body joints
 obtained using a CNN as additional features for 3D pose regression. All these approaches demonstrate very good performances for 3D pose 
estimation, but require a large amount of training data containing pairs of images and ground-truth 3D poses to train deep network architectures.
This limits their applicability to the environments of the training data.

Estimating 3D human pose from a given 2D pose by exploiting motion capture data has also been addressed in the literature
~\cite{SimoSerraCVPR2012,Ramakrishna_2012,yasin-2013,SimoSerraCVPR2013,Wang_2014_CVPR, zhou2015sparse, bogo2016keep, sanzari2016bayesian, chen2017matching, lassner2017unite, tome2017lifting}. 
While early approaches~\cite{Ramakrishna_2012, SimoSerraCVPR2012, yasin-2013} used 
manually annotated 2D joint locations,~\cite{SimoSerraCVPR2013, Wang_2014_CVPR} are one of the first approaches that estimate the 3D pose from estimated 2D poses. 
With the progress in 2D pose estimation methods 
\cite{toshev2014deeppose, pishchulin2016deepcut, carreira2015human, iqbalFG2017actionpose, wei2016convolutional, hu2016bottom, 
insafutdinov2016deepercut, newell2016eccv, bulat2016human, georgia2016eccv, rafi2016bmvc, chu2017CVPRmulti}, the number of approaches in this category 
also rose~\cite{zhou2015sparse, bogo2016keep, chen2017matching, lassner2017unite, tome2017lifting}. All these approaches have the 
benefit that they do not require training data containing images with annotated 3D poses, but rather only utilize 3D pose data to build their models.

In~\cite{yasin-2013}, the 2D pose is manually annotated in the first frame and tracked in a video. A nearest neighbor search is then
performed to retrieve the closest 3D poses. In~\cite{Ramakrishna_2012} a sparse representation of 3D human pose is constructed from a MoCap dataset and
fitted to manually annotated 2D joint locations. The approach has been extended in \cite{Wang_2014_CVPR} to
handle poses from an off-the-shelf 2D pose estimator~\cite{YiYang-2011}, and subsequently to videos in \cite{du2016marker}.
The information about the 2D body joints is used in \cite{SimoSerraCVPR2012,SimoSerraCVPR2013} to constrain the search space of 3D poses. In \cite{SimoSerraCVPR2012}
an evolutionary algorithm is used to sample poses from the pose space that correspond to the estimated 2D joint positions.
This set is then exhaustively evaluated according to some anthropometric constraints. The approach is extended
in~\cite{SimoSerraCVPR2013} such that the 2D pose estimation and 3D pose estimation are iterated. In contrast
to~\cite{Ramakrishna_2012,Wang_2014_CVPR,SimoSerraCVPR2012}, the approach \cite{SimoSerraCVPR2013} deals with 2D pose 
estimation errors. 

An expectation maximization algorithm is presented in~\cite{zhou2015sparse} to estimate 3D poses from monocular videos. 
Additional smoothness constraints are used to exploit the temporal information in videos. 
In addition to the 3D pose,~\cite{bogo2016keep} also estimates the 3D shape of the person. The approach exploits a high-quality 3D human body model and fits
it to estimated 2D joints using an energy minimization objective. The approach is improved further in~\cite{lassner2017unite} by introducing an extra fitting objective and generating additional training data. In~\cite{chen2017matching} a non-parametric nearest neighbor model is used to retrieve 3D exemplars 
that minimize the reprojection error from the estimated 2D joint locations. More recently,~\cite{tome2017lifting} proposes a probabilistic 3D pose model and combines it with a multi-staged 
CNN, where the CNN incorporates evidences from the 2D body part locations and projected 3D poses to sequentially improve 2D joint predictions which in turn also results in better 3D pose estimates. 

Action specific priors learned from motion capture data have also been proposed for 3D pose tracking ~\cite{UrtasunFF06,andrilukaCVPR2010}. 
These approaches, however, are more constrained by assuming that the type of motion is known in advance and 
therefore cannot deal with a large and diverse pose dataset. 

\section{Overview}\label{sec:overview}
In this work, we aim to predict the 3D pose from an RGB image. Since acquiring 3D pose data in natural 
environments is impractical and annotating 2D images with 3D pose data is infeasible, we do not assume that
our training data consists of images annotated with 3D pose. Instead, we propose an approach that utilizes two
independent sources of training data. The first source consists of motion capture data, which is publicly
available in large quantities and that can be recorded in controlled environments. The second source consists of 
images with annotated 2D poses, which is also available and can be easily provided by humans. Since we do not assume
that we know any relations between the sources except that the motion capture data includes the poses we are
interested in, we preprocess the sources first independently as illustrated in Fig.~\ref{fig:sysflow}. From the 
image data, we learn a CNN to predict 2D poses from images. This will be discussed 
in Section~\ref{sec:2DPose}. The motion capture data is prepared to efficiently retrieve 3D poses that could 
correspond to a 2D pose. This part is described in Section \ref{sec:similarity}. We then estimate the 3D pose by minimizing the projection 
error under the constraint that the solution is close to the retrieved poses (Section~\ref{sec:posrec}). The source code of the approach is publicly available.\footnote{\url{http://pages.iai.uni-bonn.de/iqbal_umar/ds3dpose/}}

\section{2D Pose Estimation}\label{sec:2DPose}
In this work, we use the convolutional pose machines (CPM) \cite{wei2016convolutional} for 2D pose estimation, but other CNN architectures, \eg stacked hourglass~\cite{newell2016eccv} or multi-context attention models~\cite{chu2017CVPRmulti}, could be used as well. Given an image $I$, we define the 2D pose of the person
 as $\bold{x} = \{x_j\}_{j\in \mathcal{J}}$, where $x_j \in \mathbb{R}^2$ is the 2D coordinates
of body joint $j$, and $\mathcal{J}$ is the set of all body joints. CPM consists of a multi-staged
CNN architecture, where each stage $t \in \{1 \dots T\}$ produces a set of confidence scoremaps $\bold{s_t} = \{s^j_t\}_{j \in \mathcal{J}}$, 
where  $s^j_t \in \mathbb{R}^{w \times h}$ is the confidence score map of body joint $j$ at 
stage $t$, and $w$ and $h$ are the width and the height of the image, respectively. Each stage of the network 
sequentially refines the 2D pose estimates by utilizing the output of the preceding stage and also the features extracted 
from the raw input image. The final 2D pose $\bold{x}$ is obtained as 

\begin{equation}
\bold{x} = \argmax_{\bold{x'} = \{x'_{j}\}_{j\in \mathcal{J}}}  \sum_{j \in \mathcal{J}}^{} s^j_{T}(x'_j).
\end{equation}.

In our experiments we will show that training the network on publicly available dataset for 2D pose estimation in-the-wild, 
such as MPII Human Pose dataset \cite{andriluka14cvpr}, is sufficient to obtain state-of-the-art results with our proposed method.

\section{3D Pose Estimation}
While the CNN for 2D pose estimation is trained on the images with 2D pose annotations as shown in Fig.~\ref{fig:sysflow}, 
we now describe an approach that makes use of a second dataset with 3D poses in order to predict the 3D pose from an image. 
Since the two sources are independent, we first have to establish relations between 2D poses and 3D poses.
This is achieved by using an estimated 2D pose as query for 3D pose retrieval (Section \ref{sec:similarity}). 
The retrieved poses, however, contain many wrong poses due to errors in 2D pose estimation, 2D-3D ambiguities
and differences of the skeletons in the two training sources. It is therefore necessary to fit the 3D poses to the 2D observations. 
This will be described in Section~\ref{sec:posrec}.
\subsection{3D Pose Retrieval}\label{sec:similarity}
In order to efficiently retrieve 3D poses for a 2D pose query, we preprocess the motion capture data.
We first normalize the poses by discarding orientation and translation information from the poses in our
motion capture database. We denote a 3D normalized pose with \ptdn~and the 3D normalized pose space with \nspacet.
As in \cite{yasin-2013}, we project the normalized poses $\ptdn \in \nspacet$ to 2D using orthographic projection.
We use 144 virtual camera views with azimuth angles spanning 360 degrees and elevation angles in the range of 0 and
90 degree. Both angles are uniformly sampled with a step size of 15 degree. We further normalize the projected 2D 
poses by scaling them such that the y-coordinates of the joints are within the range of $[-1, 1]$.
The normalized 2D pose space is denoted by \nspace\ and does not depend on a specific camera model
or coordinate system. This step is illustrated in Fig.~\ref{fig:sysflow}. After a 2D pose is estimated by the 
approach described in Section~\ref{sec:2DPose}, we first normalize it according to \nspace, \ie, we 
translate and scale the pose such that the y-coordinates of the joints are within the range of $[-1, 1]$,  
then use it to retrieve 3D poses. The distance between two normalized 2D poses is given by the average Euclidean distance
of the joint positions. The $\mathsf{K}$-nearest neighbors in \nspace\ are efficiently retrieved by a
\kdtree~\cite{krueger-2010}. The retrieved normalized 3D poses are the corresponding poses in \nspacet. 

\subsection{3D Pose Estimation}\label{sec:posrec}
In order to obtain the 3D pose $\mathbf{X}$, we have to estimate the unknown projection 
$\mathcal{M}$ from the normalized pose space \nspacet\ to the image. To this end, we minimize the energy
\begin{equation}
E(\mathbf{X},\mathcal{M}) = E_{p}(\mathbf{X},\mathcal{M}) + \alpha E_{r}(\mathbf{X})
\label{eq:energyMin}
\end{equation}
consisting of the two terms $E_{p}$ and $E_{r}$, where $\alpha$ is a weighting parameter. 

The first term $E_{p}(\mathbf{X},\mathcal{M},s)$ measures the projection error of the 3D pose $\mathbf{X}$ and the projection $\mathcal{M}$:
\begin{equation}\label{eq:energyControl}
E_{p}(\mathbf{X},\mathcal{M}) = \left( \sum_{j \in \mathcal{J}} \| \mathcal{M}\left(X_{j}\right) - x_j\|^2 \right)^{\frac{1}{2}},
\end{equation}
where $x_j$ is the joint position of the predicted 2D pose and $X_{j}$ is the 3D joint position of the unknown 3D pose.

The second term enforces that the pose $\mathbf{X}$ is close to the retrieved 3D poses $\mathbf{X}^{\mathsf{k}}$:
\begin{equation}
E_{r}(\mathbf{X}) = \sum_{\mathsf{k}} \left( \sum_{j \in \mathcal{J}} \| X^{\mathsf{k}}_{j} - X_j\|^2 \right)^{\frac{1}{2}}.
\label{eq:energyMotion}
\end{equation}

Minimizing the energy $E(\mathbf{X},\mathcal{M})$ \eqref{eq:energyMin} over the continuous parameters $\mathbf{X}$ and $\mathcal{M}$ would be expensive.
We therefore propose to obtain an approximate solution where we estimate the projection $\mathcal{M}$ first.
For the projection, we assume that the intrinsic parameters are given and only estimate the global orientation and translation.
The projection $\hat{\mathcal{M}}$ is estimated by minimizing
\begin{equation}
\hat{\mathcal{M}} = \argmin_{\mathcal{M}} \left\{ \sum_{\mathsf{k}=1}^{\mathsf{K}} E_{p}(\mathbf{X}^{\mathsf{k}},\mathcal{M}) \right\}
\label{eq:projerr1}
\end{equation}
using non-linear gradient optimization. Given the estimated projections $\hat{\mathcal{M}}$, we minimize
\begin{equation}
\hat{\mathbf{X}} = \argmin_{\mathbf{X}} \left\{ E(\mathbf{X},\hat{\mathcal{M}}) \right\}
\label{eq:energyMin2}
\end{equation}
to obtain the 3D pose $\mathbf{X}$. In our experiments, we will also evaluate the case when camera orientation and translation are also known.
In this case the projection $\mathcal{M}$ reduces to a rigid transformation of 3D pose $\mathbf{X}$ from normalized pose space $\nspacet$ 
to the camera coordinate system. 

The dimensionality of $\mathbf{X}$ can be reduced by applying PCA to the retrieved 3D poses $\mathbf{X}^{\mathsf{k}}$.
Reducing the dimensions of $\mathbf{X}$ helps to decrease the optimization time without loss in accuracy, as we will show in the experiments.

\section{Experiments}\label{sec:exp}
We evaluate the proposed approach on two publicly available datasets, namely  Human3.6M~\cite{h36m_pami} and
HumanEva-I~\cite{Sigal_2010}. Both datasets provide accurate 3D poses for each image and camera parameters.
For all cases, 2D pose estimation is performed by convolutional pose machines \cite{wei2016convolutional} trained 
on the MPII Human Pose dataset \cite{andriluka14cvpr}, and no fine-tuning is performed, unless stated otherwise. 

\subsection{Evaluation on Human3.6M Dataset}
For evaluation on the Human3.6M dataset, a number of protocols have been proposed in the literature. The protocol originally proposed for the Human3.6M dataset~\cite{h36m_pami}, which we denote by \emph{Protocol-III}, uses the annotated bounding boxes
and the training data only from the action class of the test data. This simplifies the task due to the small
pose variations for a single action class and the known person bounding box. Other protocols have been therefore proposed in \cite{Ilya_2014} and \cite{bogo2016keep}. 
In order to compare with other existing approaches, we report results for all three protocols~\cite{Ilya_2014,bogo2016keep} and \cite{h36m_pami}.
\subsubsection{Human3.6M Protocol-I}
\emph{Protocol-I}, which was proposed by \cite{Ilya_2014}, is the most unconstrained protocol.
It does not 
make any assumption about the location and activity labels during testing, and the training data comprises all
action classes. The training set consists of six subjects (S1, S5, S6, S7, S8 and S9), whereas the testing 
is performed on every $64^{th}$ frame taken from the sequences of S11. For evaluation, we use the \textit{3D pose error} 
as defined in~\cite{SimoSerraCVPR2012}. The error measures the accuracy of the relative pose up to a rigid transformation.
To this end, the estimated skeleton is aligned to the ground-truth skeleton by a rigid transformation and the average 3D
Euclidean joint error after alignment is measured, where the body skeleton consists of 14 body joints namely head, neck,
ankles, knees, hips, wrists, elbows and shoulders. In order to comply with the protocol, we do not use ground truth person
bounding boxes, but instead use an off-the-shelf person detector~\cite{RenHG0CoRR15} to detect the person's 
bounding box required for 2D pose estimation. We consider two sources for the motion capture data, namely the Human3.6M and
the CMU motion capture dataset. 

We first evaluate the impact of the parameters of our approach and the impact of different MoCap datasets. We then compare our approach with the state-of-the-art and evaluate the impact of the 2D pose estimation accuracy.    

\subsubsection*{Parameters}

\begin{figure}[t]
\begin{center}
\includegraphics[width=1\linewidth]{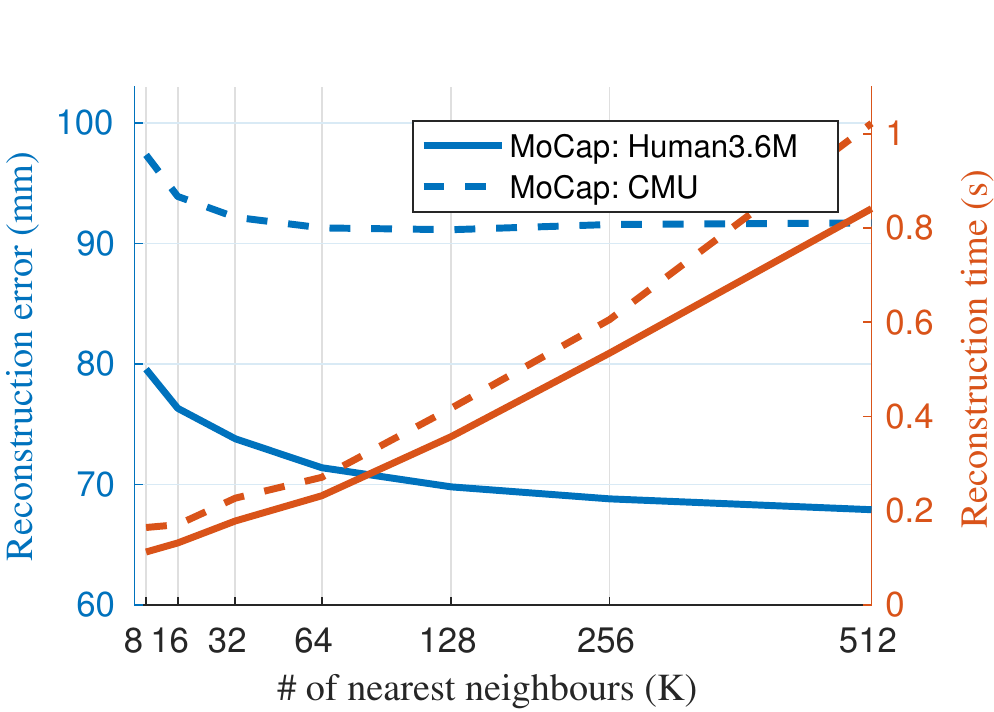}
\end{center}
   \caption{Impact of the number of nearest neighbors $K$.}
\label{fig:ImpactOfK}
\end{figure}

\noindent\textbf{Nearest Neighbors.} The impact of the number of nearest neighbors $K$ used during
3D pose reconstruction is evaluated in Fig.~\ref{fig:ImpactOfK}. Increasing the number of nearest neighbors
improves 3D pose estimation. This, however, also increases the reconstruction time. In the rest of this paper, we use a default value of $K=256$ that provides a good trade-off between accuracy and run-time. We can see 
that using the CMU MoCap dataset results in a higher error as compared to the Human3.6M dataset. We will evaluate
the impact of different MoCap datasets in more details later in this section. \\

\noindent\textbf{PCA.} PCA can be used to reduce the dimension of $\mathbf{X}$. While
in \cite{Yasin_2016_CVPR} a fixed number of principal components is used, we use a more adaptive approach and set the number of principal components based on the captured variance. The number of principal components therefore varies for each image.   

\begin{figure}[t]
\begin{center}
\includegraphics[width=1\linewidth]{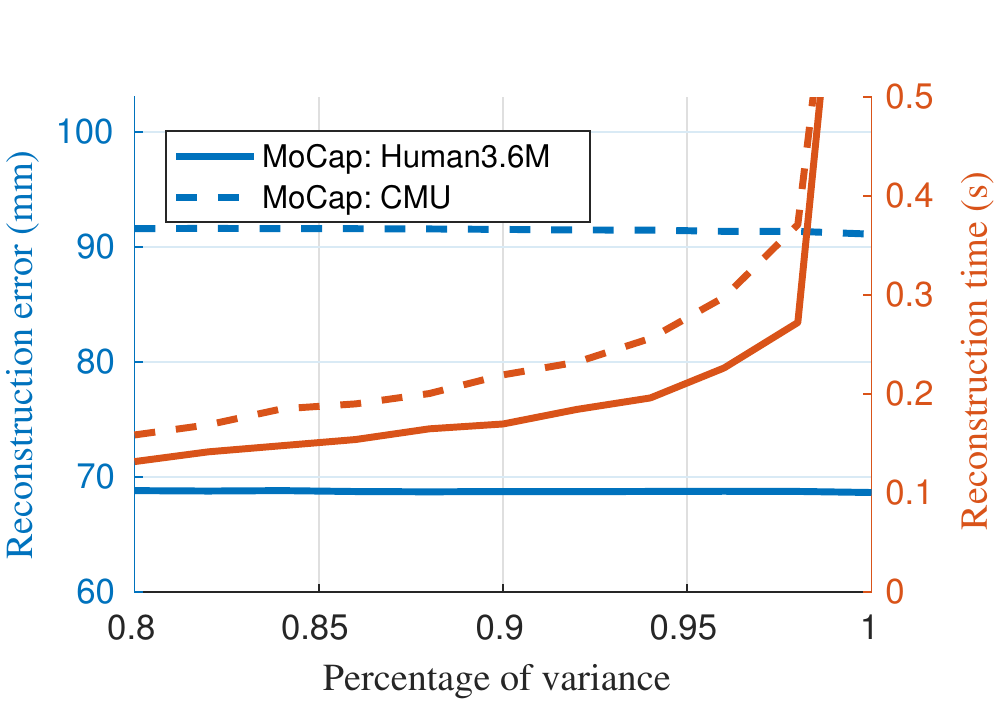}
\end{center}
   \caption{Impact of PCA. The number of principle components are selected based on the minimum number of components that explain a given percentage of variation. The x-axis corresponds to the threshold for the cumulative amount of variation.}
\label{fig:PCAVariancePlot}
\end{figure}

The impact of the threshold on the minimum amount of variation can be seen in Fig.~\ref{fig:PCAVariancePlot}. If the threshold is within a reasonable range, \ie between $0.8$ and $1$, the accuracy is barely reduced while the runtime decreases significantly compared to $1$, \ie without PCA. In this work, we use the minimum number of principle components that explain at least $80\%$ of the variance of the retrieved 3D poses $\mathbf{X}^{\mathsf{k}}$.\\

\begin{figure}[t]
\begin{center}
\includegraphics[width=1\linewidth]{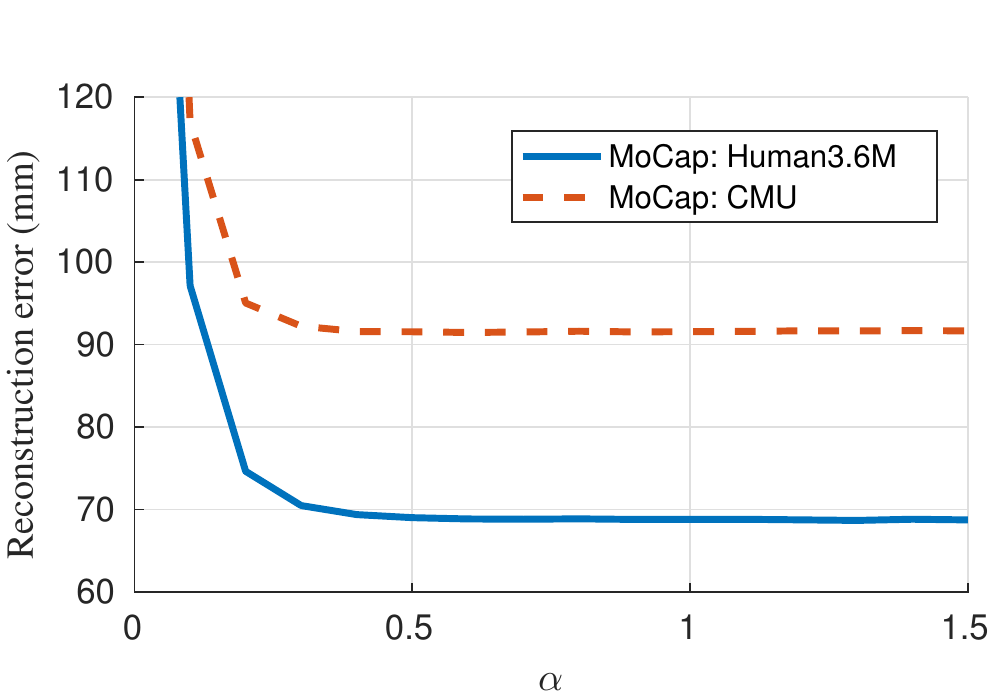}
\end{center}
   \caption{Impact of $\alpha$.}
\label{fig:ImpactOfAlpha}
\end{figure}

%
\noindent\textbf{Energy Terms.} The impact of the weight $\alpha$ in~\eqref{eq:energyMin} is reported
in Fig.~\ref{fig:ImpactOfAlpha}. If $\alpha=0$, the term $E_{r}$ is ignored and the error is very high. This is expected
since $E_{r}$ constrains the possible solution while $E_{p}$ ensures that the estimated 3D pose projects onto the estimated 2D pose.  
In our experiments, we use $\alpha=1$. \\

\subsubsection*{MoCap Data}\label{sec:mocap_data}

\begin{figure}[t]
\begin{center}
\includegraphics[width=1\linewidth]{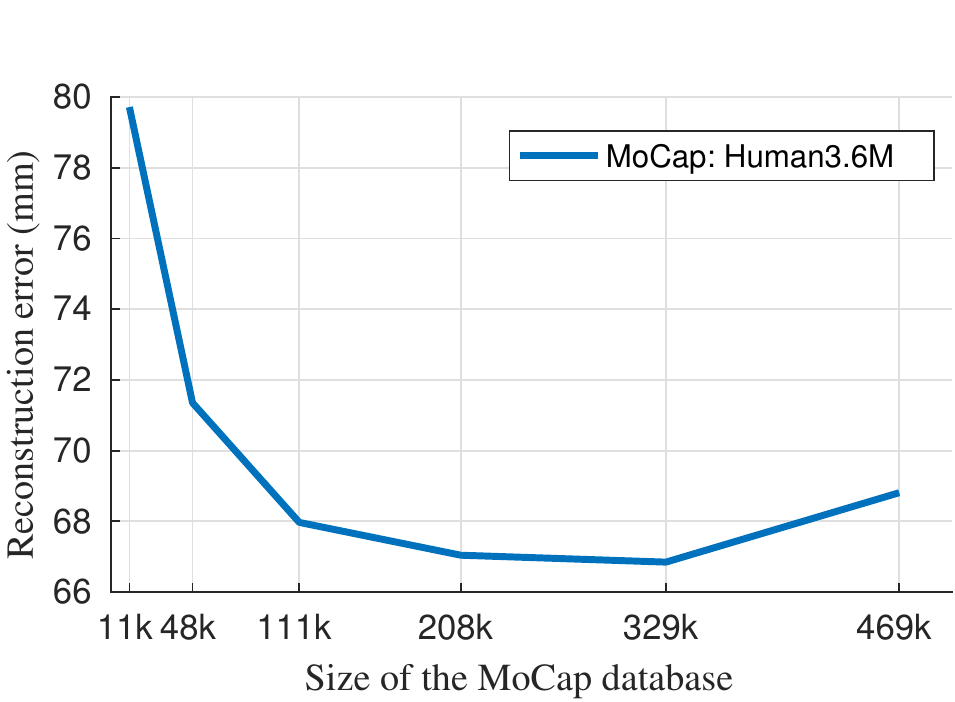}
\end{center}
   \caption{Impact of the size of the MoCap dataset. }
\label{fig:impactOfMocapH}
\end{figure}

\textbf{Impact of MoCap dataset size.}
We evaluate the impact of the size of the MoCap dataset in Fig.~\ref{fig:impactOfMocapH}. Using the entire 469K 3D poses of the 
Human3.6M training set as motion capture data results in a 3D pose error of 68.8mm. Reducing the size of the MoCap data to 329K by
removing similar poses reduces the error to 66.85mm. We consider two poses as similar when the Euclidean distance between 
both poses is less than a certain threshold, which is 20mm in this case. Removing similar poses ensures that the retrieved nearest neighbors for each
test image embody a sufficient variety of 3D poses. However, decreasing the size of the MoCap dataset even further 
degenerates the performance. In the rest of our experiments, we use the MoCap dataset from Human3.6M with 329K 3D poses, 
where a threshold of 20mm is used to remove similar poses.\\

\begin{table*}[!]
\centering
\begin{tabularx}{\linewidth}{lsssssssss}
\toprule \midrule
MoCap data 				& Direction		& Discuss		& Eating	& Greeting 		& Phoning 		& Posing		& Purchases 		& Sit 			& SitDown	\\
\midrule \midrule
Human3.6M				& 59.5			&52.4			&75.5		&67.0			&58.8			&64.9			&58.2			&68.4			& 89.7		\\	
Human3.6M  $\setminus$ Activity		& 61.2			&52.3			&92.6		&70.2			&61.1			&66.5			&59.3			&85.6			& 122.2		\\ 
Human3.6M  $\in$ Activity		& 68.8			&57.6			&70.8		&73.7			&62.9			&66.7			&63.4			&73.4			& 99.4		\\ 
Human3.6M + GT 3D Poses			& 52.9			&45.7			&59.9		&60.1			&50.4			&54.1			&51.6			&56.3			& 71.7		\\
CMU					& 73.3			&64.7			&95.9		&80.2			&85.7			&81.8			&77.1			&110.5			& 138.8	 	\\ 
\midrule \midrule
MoCap data  				& Smoking 		& Photo 		& Waiting 	& Walk 			& WalkDog 		& WalkTogehter 		& Mean 			& Median 		& -\\
\midrule
\midrule
Human3.6M				& 73.0			&88.5			&67.7		&52.1			&73.0			&54.1			&66.9			&61.5		 	&-\\
Human3.6M  $\setminus$ Activity 	&74.8			&92.6			&72.4		&64.5			&74.6			&69.0			&74.5			&67.3			&-\\
Human3.6M  $\in$ Activity 		&74.8			&89.5			&77.4		&49.3			&70.8			&55.9			&70.4			&65.3			&-\\
Human3.6M + GT 3D Poses			&64.2			&69.2			&60.4		&47.8			&60.6			&44.9			&56.7			&51.3			&-\\
CMU					&100.9			&95.3			&90.6		&82.9			&87.6			&91.3			&91.0			&83.3			&-\\
\bottomrule
\end{tabularx}
\caption{Impact of the MoCap dataset. While for Human3.6M  $\setminus$ Activity we removed all poses from the dataset that correspond to the activity of the test sequence, Human3.6M  $\in$ Activity only contains the poses of the activity of the test sequence. For Human3.6M + GT 3D Poses, we include the ground-truth 3D poses of the test sequences to the MoCap dataset.}
\label{tab:impactOfMoCap}
\end{table*}

\begin{figure}[t]
\begin{center}
\includegraphics[width=1\linewidth]{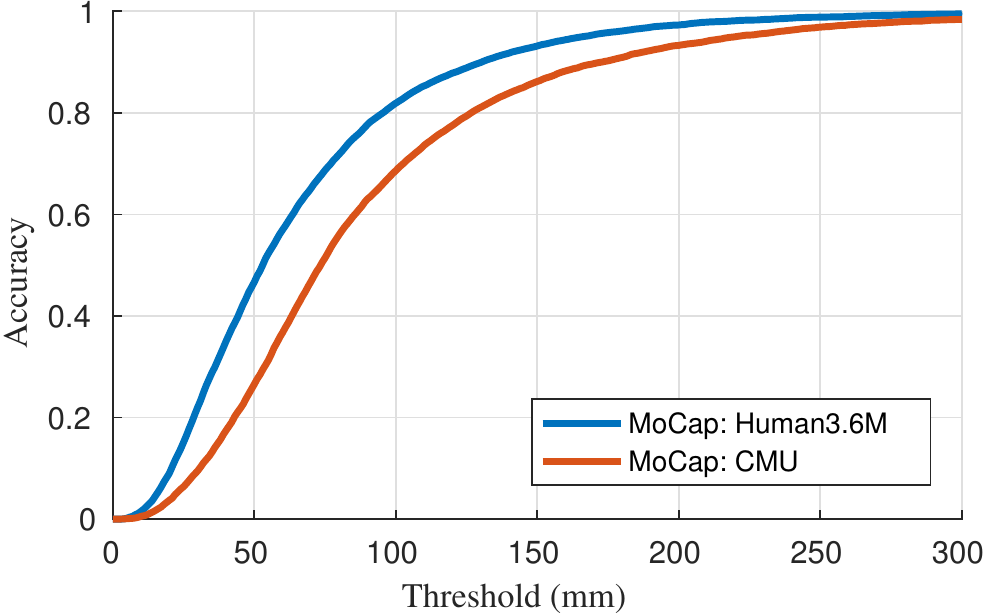}
\end{center}
   \caption{Comparison of 3D pose error using different MoCap datasets. The plot represent the percentage of estimated 3D poses with an error below a specific threshold.}
\label{fig:H36M_VS_CMU_AccuracyPlot}
\end{figure}

\noindent\textbf{CMU Motion Capture Dataset.}
Since we do not assume that the images are annotated by 3D poses but use motion capture data as a second training source.
We therefore evaluate our approach using the CMU motion capture dataset~\cite{cmu_mocap} for our 3D pose retrieval. 
We use one third of the CMU dataset and downsample the CMU dataset from 120Hz to 30Hz, resulting in 360K 3D poses. 
We remove similar poses using the same threshold (20mm) as used for Human3.6M which results in a final MoCap dataset with 303K 3D poses.
Fig.~\ref{fig:H36M_VS_CMU_AccuracyPlot} compares the pose estimation accuracy using both datasets, while the results for each activity 
can be seen in Tab.~\ref{tab:impactOfMoCap}. As expected the error is higher due to the differences of the datasets. 

To analyze the impact of the motion capture data more in detail, we have evaluated the pose error for various modifications of the MoCap data in Tab.~\ref{tab:impactOfMoCap}. 
First, we remove all poses of a selected activity from the MoCap data and evaluate the 3D pose error 
for the test images corresponding to the removed activity. The error increases since the dataset 
does not contain poses related to the removed activity anymore. While the error still stays comparable for many activities, \eg Direction, Discussion, etc., a significant increase in error can be seen for activities that do not share similar 
poses with other activities \eg SitDown. However, even if all poses related to the activity of the test images are removed, the results are still good and better compared to the CMU dataset. This indicates that the error increase for the CMU dataset cannot only be explained by the difference of poses, but also other factors like different motion capture setups seem to influence the result. We will investigate the impact of the skeleton structure between two datasets in Section~\ref{sec:EvaI}.

We also evaluate the case when MoCap data only consist of the poses of a specific activity. 
This also results in an increased mean pose estimation error and shows that having a diverse MoCap dataset is helpful to obtain good performance. 
Finally, we also report the error when the 3D poses of the test sequences are added to the MoCap dataset. 
In this case, the error is reduced from 61.53mm to 51.29mm. 

\begin{table*}[htbp]
\centering
\begin{tabularx}{\linewidth}{lsssssssss}
\toprule \midrule
Method 					& Direction		& Discuss		& Eating	& Greeting 		& Phoning 		& Posing		& Purchases 		& Sit 			& SitDown\\
\midrule \midrule
Kostrikov \& Gall \cite{Ilya_2014}	& -			&-			&-		&-			&-			&-			&-			&-			&-\\
Yasin \etal \cite{Yasin_2016_CVPR}	& 88.4 			& 72.5 			& 108.5 	& 110.2 		& 97.1 			& 81.6 			& 107.2 		& 119.0 		& 170.8 \\
Rogez \& Schmid \cite{rogez2016mocap}	& -			&-			&-		&-			&-			&-			&-			&-			&-\\
Chen \& Ramanan \cite{chen2017matching}	& 71.6 			& 66.6	 		&{74.7} 	&{79.1} 		&{70.1} 		&{67.6} 		&{89.3} 		&{90.7} 		&195.6 \\
Moreno-Noguer \cite{Moreno_arxiv2016}	& 67.4 			& 63.8 			& 87.2 		& 73.9 			& 71.5 			& 69.9 			& 65.1 			& 71.7 			&98.6 \\
Tome \etal \cite{tome2017lifting}	& -			&-			&-		&-			&-			&-			&-			&-			&-\\
\textbf{Ours}				& \textbf{59.5}		&\textbf{52.4}		&\textbf{75.5}	&\textbf{67.0}		&\textbf{58.8}		&\textbf{64.9}		&\textbf{58.2}		&\textbf{68.4}		& \textbf{89.7}	\\	
\midrule
\multicolumn{10}{c}{(MoCap from CMU dataset)} \\  
\midrule
Yasin \etal \cite{Yasin_2016_CVPR}	& 102.8			&80.4			& 133.8		&120.5			& 120.7			& 98.9			&117.3 			&150.0 			& 182.6 \\
\textbf{Ours}				& \textbf{73.3}		&\textbf{64.7}		&\textbf{95.9}	&\textbf{80.2}		&\textbf{85.7}		&\textbf{81.8}		&\textbf{77.1}		&\textbf{110.5}		&\textbf{138.8}	 \\ 
\midrule \midrule
Method 					& Smoking 		& Photo 		& Waiting 	& Walk 			& WalkDog 		& WalkTogehter 		& Mean 			& Median 		& -\\
\midrule
\midrule
Kostrikov \& Gall \cite{Ilya_2014}	&-			&-			&-		&-			&-			&-			& 115.7 		& - 			&-\\
Yasin \etal \cite{Yasin_2016_CVPR}	& 108.2 		& 142.5 		& 86.9 		& 92.1 			& 165.7 		& 102.0 		& 108.3 		& - 			&-\\
Rogez \& Schmid \cite{rogez2016mocap}	&-			&-			&-		&-			&-			&-			& 88.1 		& - 			&-\\
Chen \& Ramanan \cite{chen2017matching} &{83.5} 		&{93.3} 		&{71.2} 	&{55.7} 		&{85.9} 		&{62.5}			& {82.7} 		& {69.1}		&-\\
Moreno-Noguer \cite{Moreno_arxiv2016} 	& 81.33 		& 93.3 			& 74.6 		& 76.5 			& 77.7 			& 74.6  		& 76.5			&- 			&-\\
Tome \etal \cite{tome2017lifting}	&-			&-			&-		&-			&-			&-			& 70.7 			& - 			&-\\
\textbf{Ours}				& \textbf{73.0}		&\textbf{88.5}		&\textbf{67.7}	&\textbf{52.1}		&\textbf{73.0}		&\textbf{54.1}		&\textbf{66.9}		&\textbf{61.5} 		&-\\
\midrule 
\multicolumn{10}{c}{(MoCap from CMU dataset)} \\  
\midrule
Yasin \etal \cite{Yasin_2016_CVPR} 	& 135.6 		&  140.1 		& 104.7 	& 111.3 		&167.0 			&116.8			& 124.8 		& - 			&-\\
\textbf{Ours}				&\textbf{100.9}		&\textbf{95.3}		&\textbf{90.6}	&\textbf{82.9}		&\textbf{87.6}		&\textbf{91.3}		&\textbf{91.0}		&\textbf{83.3}		&-\\
\bottomrule
\end{tabularx}
\caption{Comparison with the state-of-the-art on the Human3.6M dataset using \emph{Protocol-I}.}
\label{tab:s11_compare_p1}
\end{table*}

\subsubsection*{Comparison with State-of-the-art}\label{exp:comp_p1}
Tab.~\ref{tab:s11_compare_p1} compares the performance of the proposed method with the state-of-the-art 
approaches \cite{Ilya_2014, Yasin_2016_CVPR, rogez2016mocap, chen2017matching, Moreno_arxiv2016, tome2017lifting} using both MoCap datasets. 
Our approach outperforms the other approaches. In particular, the 3D pose error reported in \cite{Yasin_2016_CVPR} is reduced from 108.3mm to 66.9mm when using the Human3.6M 
MoCap dataset. A similar decrease in error can also be seen for the CMU dataset (124.8mm vs.\ 91.0mm). The main improvement compared to \cite{Yasin_2016_CVPR} stems from the better 2D pose estimation model. Our approach also outperforms the recent methods \cite{Moreno_arxiv2016, tome2017lifting}.
While \cite{Moreno_arxiv2016} relies on pair of images and 3D poses as training data, \cite{tome2017lifting} learns a deep CNN model 
using the 2D pose data from Human3.6M. We on the other hand do not use pairs of images and 3D pose for training and only utilize a 
pre-trained model trained on the MPII Human Pose Dataset \cite{andriluka14cvpr} for 2D pose estimation.

\subsubsection*{Impact of 2D Pose}\label{exp:comp_p2}
We also investigate the impact of the accuracy of the estimated 2D poses. 
If we initialize the approach with the 2D ground-truth poses, the 3D pose error is significantly reduced as 
shown in Tab.~\ref{tab:s11_compare_gt2d}. This indicates that the 3D pose error can be further reduced by improving
the used 2D pose estimation method. We also report the 3D pose error when both 3D and 2D ground-truth poses are available. In this case 
the error reduces even further which shows the potential of further improvements for the proposed method. We also compare our approach to \cite{Yasin_2016_CVPR} and \cite{chen2017matching}, which also report the accuracy for ground-truth 2D poses.     

{\small
\begin{table*}[htbp]
\begin{center}
\begin{tabularx}{\linewidth}{lsssssssss}
\midrule \midrule
Method 						& Direction 	& Discuss 	& Eat 		& Greet 	& Phone 	& Pose 		& Purchase 	& Sit 		& SitDown\\
\midrule \midrule
\textbf{Ours}					& 59.5		&52.4		&75.5		&67.0		&58.8		&64.9		&58.2		&68.4		& 89.7	\\	
\textbf{Ours}  + GT 2D				&51.9		&45.3		&62.4		&55.7		&49.2		&56.0		&46.4		&56.3		&76.6	\\
\textbf{Ours}  + GT 2D + GT 3D			& 40.9		&35.3		&41.6		&44.3		&36.6		&43.7		&38.0		&40.3		&53.4	\\
Yasin \cite{Yasin_2016_CVPR} + GT 2D		& 60.0 		& 54.7 		& 71.6 		& 67.5 		& 63.8 		& 61.9 		& 55.7 		& 73.9 		& 110.8 \\
Chen \& Ramanan \cite{chen2017matching} + GT 2D &53.3		&46.8 		&58.6 		&61.2		&56.0		&58.1 		&48.9 		&55.6 		&73.4\\

\midrule 
\multicolumn{10}{c}{(MoCap from CMU dataset)} \\  
\midrule
\textbf{Ours} + GT 2D				&67.8		&58.7		&90.3		&72.1		&78.2		&75.7		&71.9		&103.2		&132.8	\\
\midrule \midrule
Method 						& Smoke 	& Photo 	& Wait 		& Walk 		& WalkDog 	& WalkPair 	& Mean 		& Median 	& -\\
\midrule \midrule
\textbf{Ours}					& 73.0		&88.5		&67.7		&52.1		&73.0		&54.1		&66.9		&61.5		 &-\\
\textbf{Ours} + GT 2D 				&58.8		& 79.1		&58.9		&35.6		& 63.4		&46.3		&56.1		& 51.9		&-\\
\textbf{Ours}  + GT 2D + GT 3D			&44.2		&56.6		&45.9		&26.9		& 45.8		&31.4		& 41.6		& 39.1		&-\\
Yasin \cite{Yasin_2016_CVPR} + GT 2D		& 78.9 		& 96.9 		& 67.9 		& 47.5 		& 89.3 		& 53.4 		& 70.5 		&  - 		&- \\
Chen \& Ramanan \cite{chen2017matching} + GT 2D &60.3		&76.1		&62.2		&35.8 		&61.9		&51.1 		&57.5 		& 51.9		&-\\
\midrule 
\multicolumn{10}{c}{(MoCap from CMU dataset)} \\  
\midrule
\textbf{Ours} + GT 2D				&91.3		&91.6		&84.7		&70.9		&81.2		&76.7		&83.7		& 75.6		&-\\
\bottomrule
\end{tabularx}
\end{center}
\caption{Impact of 2D pose estimation. GT 2D denotes that the ground-truth 2D pose is used. GT 3D denotes that the 3D poses of the test images are added to the MoCap dataset as in Tab.~\ref{tab:impactOfMoCap}.}
\label{tab:s11_compare_gt2d}
\end{table*}
}

\subsubsection{Human3.6M Protocol-II} 
The second protocol, \emph{Protocol-II}, has been proposed in \cite{bogo2016keep}. The dataset is split using five subjects (S1, S5, S6, S7, S8)
for training and two subjects (S9 and S11) for testing. We follow \cite{lassner2017unite} and perform testing
on every $5^{th}$ frame of the sequences from the frontal camera (cam-3) and trial-1 of each activity. 
The evaluation is performed in the same way as in \emph{Protocol-I} with a body skeleton consisting 14 joints. 
In contrast to \emph{Protocol-I}, the \emph{ground-truth bounding boxes} are, however, used during testing. 
Tab.~\ref{tab:compare_p2} reports the comparison of the proposed method with the state-of-the-art approaches 
\cite{Akhter:CVPR:2015, Ramakrishna_2012, zhou2015sparse, bogo2016keep, lassner2017unite, tome2017lifting, Moreno_arxiv2016}.
Our approach achieves comparable results to the state-of-the-art, while also outperforming other methods on some 
activities.    \\

\begin{table*}[t]
\begin{tabularx}{\linewidth}{lsssssssss}
\toprule \midrule
						& Directions 			 & Discussion			  & Eating			& \multicolumn{1}{c}{Greeting} & \multicolumn{1}{c}{Phoning} & \multicolumn{1}{c}{Photo} & \multicolumn{1}{c}{Posing} & \multicolumn{1}{c}{Purchases} & \multicolumn{1}{c}{Sit}    \\ 
\midrule \midrule
Akhter \& Black \cite{Akhter:CVPR:2015}       	& 199.2                          & 177.6                          & 161.8                       & 197.8                        & 176.2                       & 186.5                     & 195.4                      & 167.3                         & 160.7                      \\
Ramakrishna et al. \cite{Ramakrishna_2012} 	& 137.4                          & 149.3                          & 141.6                       & 154.3                        & 157.7                       & 158.9                     & 141.8                      & 158.1                         & 168.6                      \\
Zhou et al. \cite{zhou2015sparse}       	& 99.7                           & 95.8                           & 87.9                        & 116.8                        & 108.3                       & 107.3                     & 93.5                       & 95.3                          & 109.1                      \\
SMPLify    \cite{bogo2016keep}        		& \textbf{62.0}                  & \textbf{60.2}                  & \textbf{67.8}		& \textbf{76.5}                & \textbf{92.1}               & \textbf{77.0}             & \textbf{73.0}              & 75.3	                      & 100.3             	   \\ 
Lassner \etal \cite{lassner2017unite}		&   -				 &  -				  & -				& -			       &			     &   -			 &   -			      &   -			      &  -			   \\
Tome \etal \cite{tome2017lifting}		&   -				 &  -				  & -				& -			       &			     &   -			 &   -			      &   -			      &  -			   \\
Moreno-Noguer \cite{Moreno_arxiv2016}		& 66.1	 			 & 77.9 			  & 72.6 			& 84.7 		       	       & 99.7			     & 98.5			 &  74.8 		      &65.3 			      & \textbf{93.4}			   \\
\textbf{Ours}					&75.3 				 & 75.8 			  & 70.9 			& 92.8 			       & 89.0		     	     & 101.5 			 & 78.1 		      & \textbf{61.4}		      & 97.9		   	   \\
\midrule 
\multicolumn{10}{c}{(MoCap from CMU dataset)} \\  
\midrule 
\textbf{Ours}					& 89.7 			 	& 88.6 			  	  & 94.1 			& 101.1 			& 106.3 		     & 104.1 			 & 85.9 		      & 81.0 			      & 121.7 			   \\

\midrule \midrule
						& \multicolumn{1}{c}{SitDown}    & \multicolumn{1}{c}{Smoking}    & \multicolumn{1}{c}{Waiting} & \multicolumn{1}{c}{WalkDog}  & \multicolumn{1}{c}{Walk}    & \multicolumn{1}{c}{WalkTogether}                       & \multicolumn{1}{c}{Mean}      & \multicolumn{1}{l}{Median} &-\\ 
\midrule \midrule
Akhter \& Black \cite{Akhter:CVPR:2015} 	& 173.7                          & 177.8                          & 181.9                       & 176.2                        & 198.6                       & \multicolumn{1}{c}{192.7}                              & 181.1                         & 158.1                      &-\\
Ramakrishna et al. \cite{Ramakrishna_2012} 	& 175.6                          & 160.4                          & 161.7                       & 150.0                        & 174.8                       & \multicolumn{1}{c}{150.2}                              & 157.3                         & 136.8                      &-\\
Zhou et al. \cite{zhou2015sparse} 		& 137.5                          & 106.0                          & 102.2                       & 106.5                        & 110.4                       & \multicolumn{1}{c}{115.2}                              & 106.7                         & 90.0                       &-\\
SMPLify  \cite{bogo2016keep} 			& 137.3                		 & \textbf{83.4}                  & \textbf{77.3}               & 79.7 		               & 86.8   	             & \multicolumn{1}{c}{81.7}		                      & 82.3                 	      & \textbf{69.3}              &-\\ 
Lassner \etal \cite{lassner2017unite}		&   -				 &  -				  & -				& -			       & -			     & \multicolumn{1}{c}{-}		      		      & 80.7  			      &  -			   &-\\
Tome \etal \cite{tome2017lifting}		&   -				 &  -				  & -				& -			       & -			     & \multicolumn{1}{c}{-}		      		      & \textbf{79.6}  		      &  -			   &-\\
Moreno-Noguer \cite{Moreno_arxiv2016}		& \textbf{103.1} 		 & 85.0				  & 98.8			& 80.1			       & 78.1			     & 74.8						      & 83.5			      & -			   &-\\
\textbf{Ours}					& 121.6		 		 & 84.2 			  & 85.8 			& \textbf{75.8}	 	       & \textbf{67.8}		     & \textbf{65.0} 					      & 83.8 			      & 75.3			   &-\\ 
\midrule 
\multicolumn{10}{c}{(MoCap from CMU dataset)} \\  
\midrule
\textbf{Ours}					& 146.1	 			& 98.9 				  & 101.7 			& 92.7 			       & 84.4 			     & 99.0 						      & 100.5 			      & 92.3			   &-\\
\bottomrule
\end{tabularx}
\caption {Comparison with the state-of-the-art on the Human3.6M dataset using \emph{Protocol-II}.}
\label{tab:compare_p2}
\end{table*}

\subsubsection{Human3.6M Protocol-III}
The third protocol, \emph{Protocol-III}, is the most commonly used protocol for Human3.6M. Similar to 
\emph{Protocol-II}, the dataset is split by using subjects S1, S5, S6, S7 and S8 for training and subjects S9 
and S11 for testing. The sequences are downsampled from the original frame-rate of 50fps to 10fps, and testing is 
performed on the sequences from all cameras and trials. The evaluation is performed without a rigid transformation, but both the ground-truth and estimated 3D poses are centered with respect to the root joint. We therefore have to use the provided camera parameters such that the estimated 3D pose is in the coordinate system of the camera. 
The training and testing is often performed on the same activity. However, some recent approaches also report results by training
only once for all activities. In this work, we report results under both settings. In this protocol, a body skeleton
with 17 joints is used and the ground-truth bounding boxes are used during testing. Note that even though the 3D poses
contain 17 joints, we still use the 2D poses with 14 joints for nearest neighbor retrieval and only use the corresponding 
joints for optimizing objective (\ref{eq:energyMin}).  Tab.~\ref{tab:compare_p3}
provides a detailed comparison of the proposed approach with the state-of-the-art methods 
\cite{h36m_pami, LiC14, tekin2015predicting, tekin2016structured, zhou2016sparseness, zhou2016deep, sanzari2016bayesian, tome2017lifting, Moreno_arxiv2016, chen2017matching}.  \\

\begin{table*}
\begin{center}
\begin{tabularx}{\linewidth}{lsssssssss}
\toprule \midrule
						& Directions     & Discussion		& Eating	 & Greeting       & Phoning        & Photo           & Posing         & Purchases      	& Sit\\
\toprule \midrule
LinKDE \cite{h36m_pami}*           		& 132.7          & 183.6      		& 132.4       	 & 164.4          & 162.1          & 205.9           & 150.6          & 171.3         	& 151.6\\ 
Li \etal \cite{LiC14}*            		& -              & 136.9 	      	& 96.9        	 & 124.7          & -              & 168.7 	     & -              & -              	& - \\
Tekin \etal \cite{tekin2015predicting}*    	& 102.4   	 & 158.5 	      	& 88.0        	 & 126.8          & 118.4          & 185.0           & 114.7          & 107.6         	& 136.2\\
Tekin \etal \cite{tekin2016structured}*    	& -              & 129.1 	      	& 91.4        	 & 121.7          & -              & 162.2           & -              & -              	& - \\
Tekin \etal \cite{tekin2016fusing}*        	& 85.0           & 108.8 	      	& 84.4        	 & 98.9       	  & 119.4          & \textbf{95.7}   & 98.5           & 93.8          	& \textbf{73.8}\\
Zhou \etal \cite{zhou2016sparseness}*      	& 87.4           & 109.3 	      	& 87.1        	 & 103.2      	  & 116.2          & 143.3           & 106.9          & 99.8          	& 124.5\\
Zhou \etal \cite{zhou2016deep} 	 		& 91.8 		 & 102.4		& 97.0 		 & 98.8 	  & 113.4 	   & 125.2	     & 90.0 	      & 93.9 	        & 132.2\\
Du \etal \cite{du2016marker}*			& 85.1		 & 112.7		& 104.9		 & 122.1	  & 139.1	   & 135.9	     & 105.9	      & 166.2		& 117.5\\ 
\textbf{Ours}*			 		& 97.0	 	 & 107.4 	   	& 97.2 		 & 128.9 	  & 126.2 	   & 137.0 	     & 102.5 	      & 116.5 		& 149.3\\ \midrule 
Sanzari \etal \cite{sanzari2016bayesian}  	& \textbf{48.8}  & \textbf{56.3} 	& 96.0          & \textbf{84.8}   & 96.5           & 105.6           & \textbf{66.3}  & 107.4           & 116.9 \\
Tome \etal \cite{tome2017lifting} 		& 65.0           & 73.5          	& \textbf{76.8}  & 86.4           & \textbf{86.3}  & 110.7           & 68.9           & \textbf{74.8}   & 110.2\\
Moreno-Noguer \cite{Moreno_arxiv2016}		& 69.5   	 & 80.2 		& 78.2  	 & 87.0 	  & 100.8	   & 102.7           & 76.0 	      &  69.7 		& 104.7\\
Chen \& Ramanan \cite{chen2017matching}		& 89.9 		 &97.6 			& 90.0 		 &107.9 	  & 107.3	   & 139.2	     &93.6	       &136.1 		& 133.1\\
\textbf{Ours}					& 90.9 		 & 98.4 		& 98.2 		 & 118.3 	  & 118.0 	   & 130.5 	     & 95.9 	      & 112.1 		& 146.1\\
\midrule 
\multicolumn{10}{c}{(MoCap from CMU dataset)} \\  
\midrule
\textbf{Ours}				 	& 139.4 	 & 148.0 	   & 148.3 	 & 165.2 	  & 161.7 	   & 170.1 	     & 138.6	      & 168.2 		& 168.5\\
\midrule \midrule
						& SitDown    & Smoking        	   & Waiting        & WalkDog       & Walk        & WalkTogether   & Mean		& Media		& -    \\
\midrule \midrule
LinKDE \cite{h36m_pami}*           		& 243.0           & 162.1          & 170.7          & 177.1          & 96.6           & 127.9         	& 162.1         & -		&- 	\\
Li \etal \cite{LiC14}*            		& -               & -              & -              & 132.2          & 70.0           & -              	& -             & -		&- 	 \\
Tekin \etal \cite{tekin2015predicting}*    	& 205.7          & 118.2         & 146.7        & 128.1         & 65.9          & 77.2          	& 125.3        & -		&- 	 \\
Tekin \etal \cite{tekin2016structured}*    	& -               & -              & -              & 130.5         & 65.8          & -              	& -             & -		&- 	 \\
Tekin \etal \cite{tekin2016fusing}*        	& 170.4           & 85.1           & 116.9          & 113.7          & \textbf{62.1}  & 94.8          	& 100.1         & -		&- 	 \\
Zhou \etal \cite{zhou2016sparseness}*      	& 199.2           & 107.4          & 118.1          & 114.2          & 79.4           & 97.7          	& 113.0         & -		&- 	 \\
Zhou \etal \cite{zhou2016deep} 			& 159.0	 	  &  106.9 	   & 94.4 	    &  126.1	     & 79.0 	      &  99.0         	&  107.3  	& -		&- 	\\
Du \etal \cite{du2016marker}*			& 226.9		  & 120.0	   & 117.7	    & 137.4	     & 99.3	      & 106.5		& 126.5		& -		&-	\\
\textbf{Ours}* 					& 155.4 	  & 111.4	   & 128.4 	    & 115.9 	     & 84.2 	      & 90.8 		& 117.1 	& 98.5		&-	\\ \midrule 
Sanzari \etal \cite{sanzari2016bayesian}  	& 129.6	 	  & 97.8           & \textbf{65.9}  & 130.5          & 92.6  	      & 102.2         	& 93.2          & -		&- 	 \\
Tome \etal \cite{tome2017lifting} 		& 173.9           & \textbf{85.0}  & 85.8           & 86.3 	     & 71.4           & \textbf{73.1} 	& 88.4		& -		&- 	 \\
Moreno-Noguer \cite{Moreno_arxiv2016}	 	& \textbf{113.9}  & 89.7	   & 98.5	    & \textbf{82.4}  & 79.2	      & 77.2		& \textbf{87.3} & -		&- 	 \\
Chen \& Ramanan \cite{chen2017matching}		& 240.1 	  & 106.7 	   & 106.2	    & 114.1	     & 87.0	      & 90.6		& 114.2		& 93.1		&- 	\\
\textbf{Ours}				 	& 150.1 	  & 112.4 	   & 113.5 	    & 109.2 	     & 89.1 	      & 88.4 		& 111.8		& 95.3		&- 	 \\
\midrule 
\multicolumn{10}{c}{(MoCap from CMU dataset)} \\  
\midrule
\textbf{Ours}					& 186.7 	  & 154.8 	   & 154.4 	    & 163.7 	     & 140.9 	      & 160.3 		& 157.3 	& 141.7	&-	\\ 
\bottomrule
\end{tabularx}
\end{center}
\caption {Comparison with the state-of-the-art on the Human3.6M dataset using \emph{Protocol-III}. (*perform testing and training on the same activity.) }
\label{tab:compare_p3}
\end{table*}

\begin{figure*}
\centering
\scalebox{0.95}{

\begin{tabular}{c c c | c c c}
2D Pose & View-1 & View-2 & 2D Pose & View-1 & View-2 \\
\includegraphics[height=0.13\linewidth]{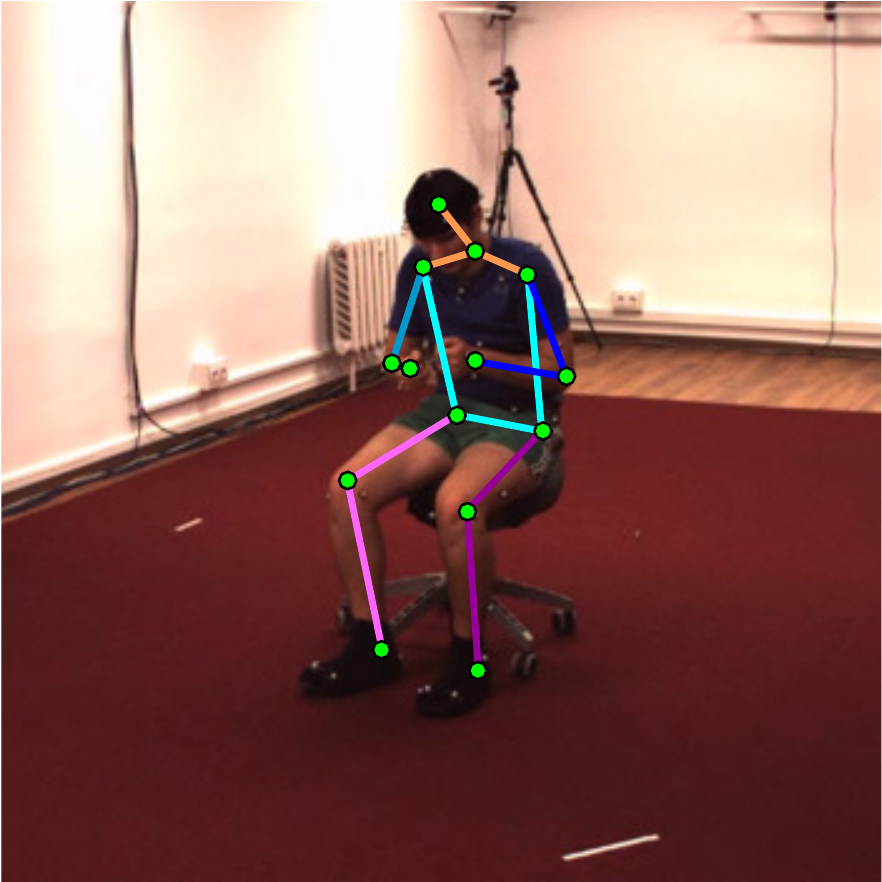} &
\includegraphics[height=0.15\linewidth]{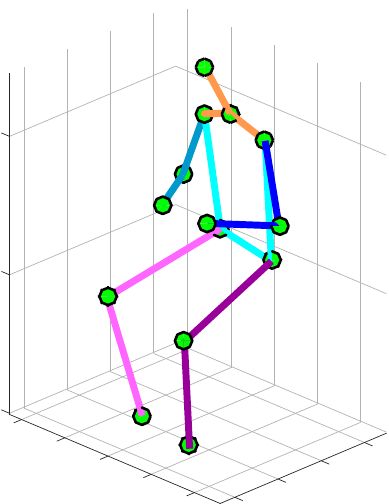} &
\includegraphics[height=0.15\linewidth]{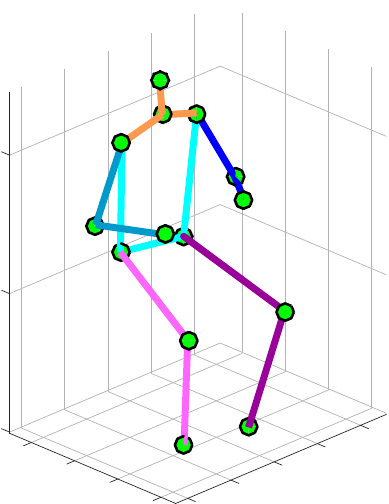} & 
\includegraphics[height=0.13\linewidth]{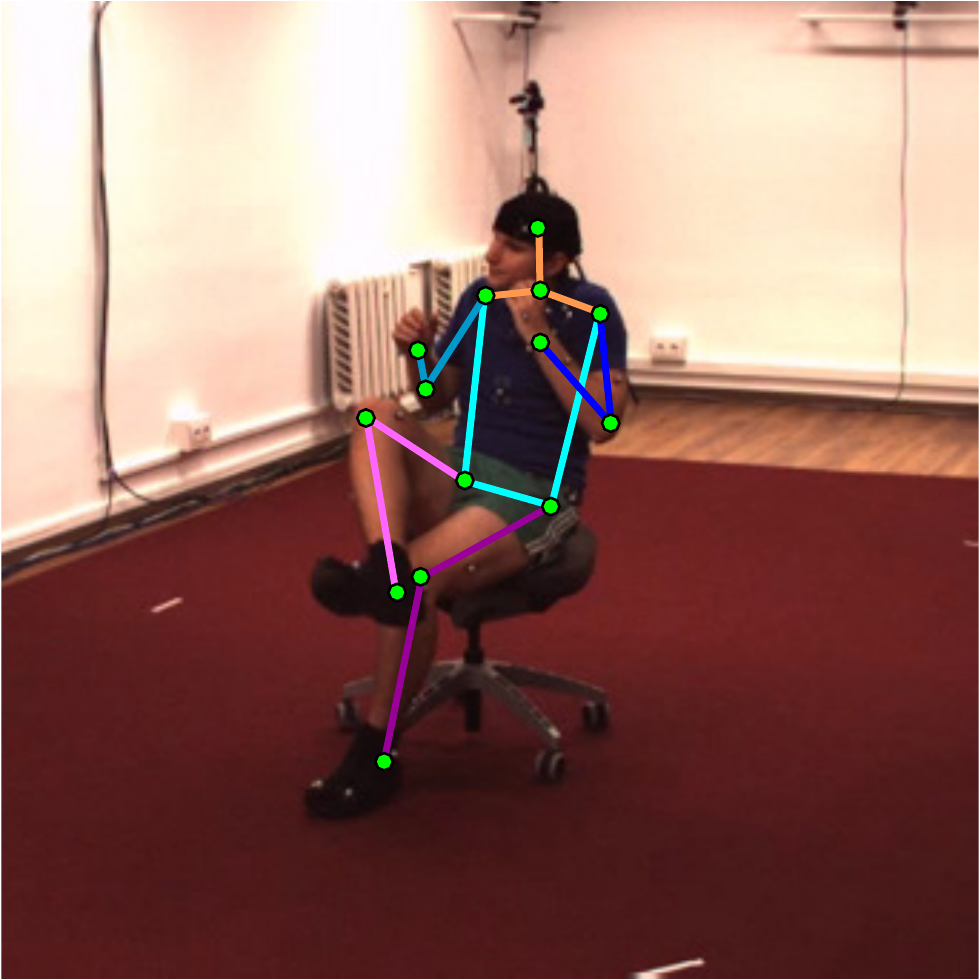} &
\includegraphics[height=0.15\linewidth]{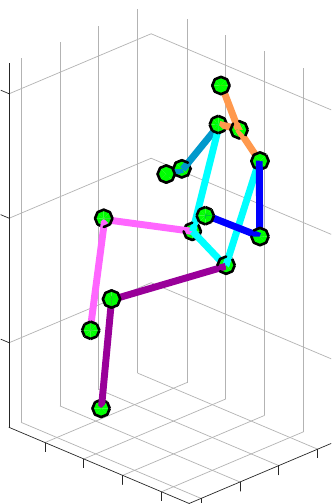} &
\includegraphics[height=0.15\linewidth]{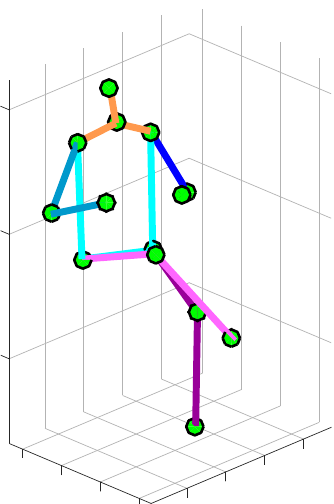} \\
\includegraphics[height=0.13\linewidth]{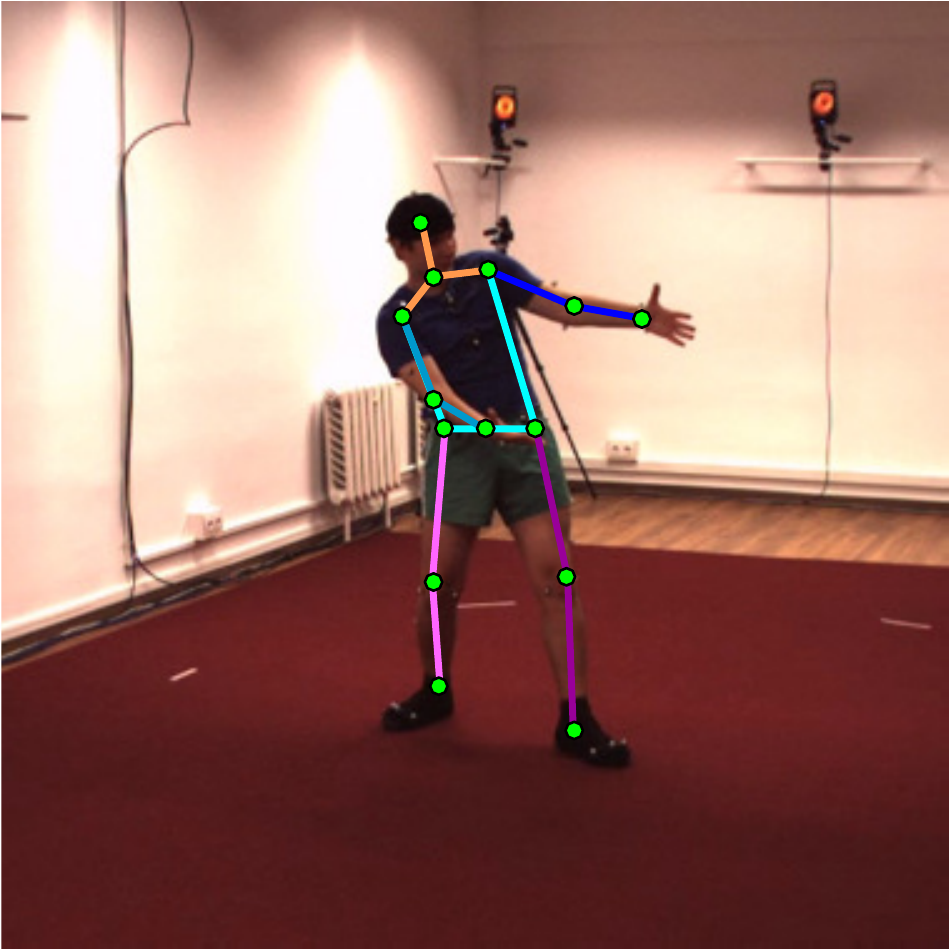} &
\includegraphics[height=0.15\linewidth]{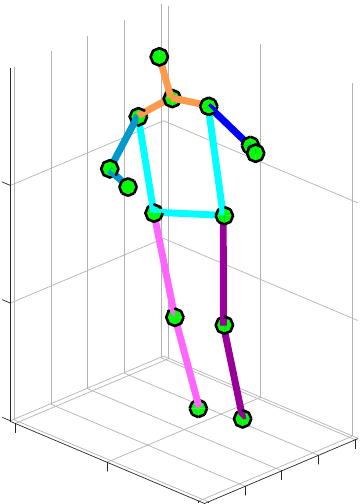} &
\includegraphics[height=0.15\linewidth]{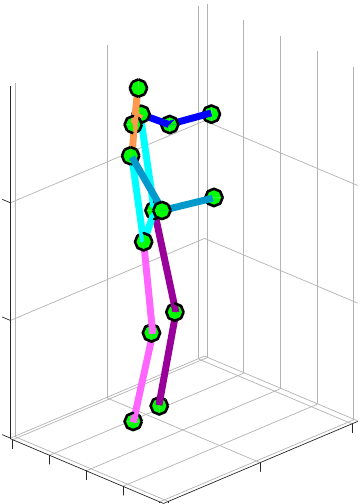} &
\includegraphics[height=0.13\linewidth]{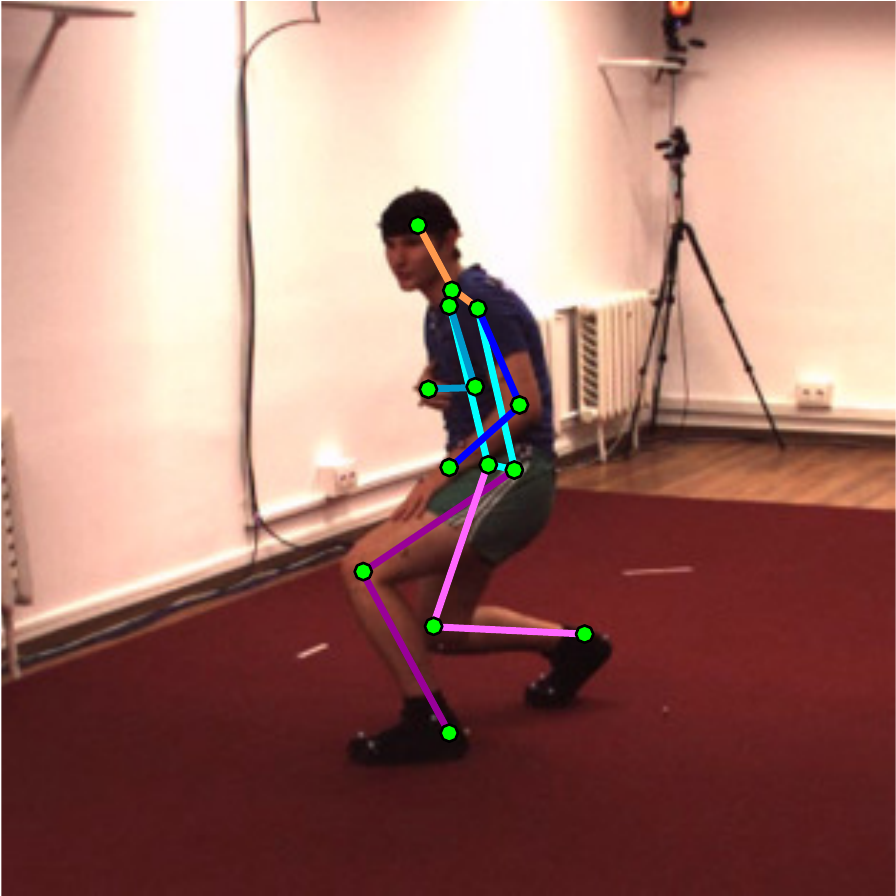} &
\includegraphics[height=0.15\linewidth]{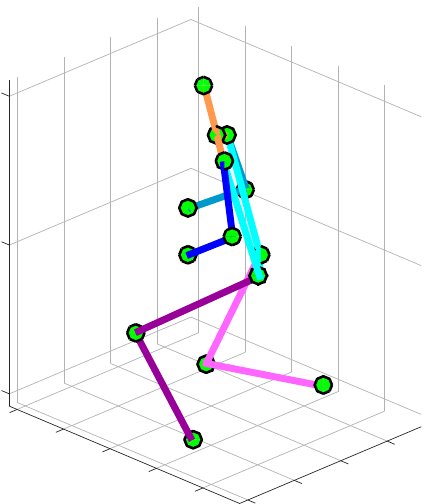} &
\includegraphics[height=0.15\linewidth]{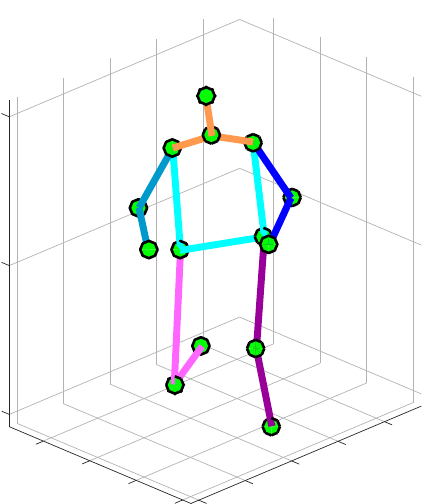} \\
\includegraphics[height=0.13\linewidth]{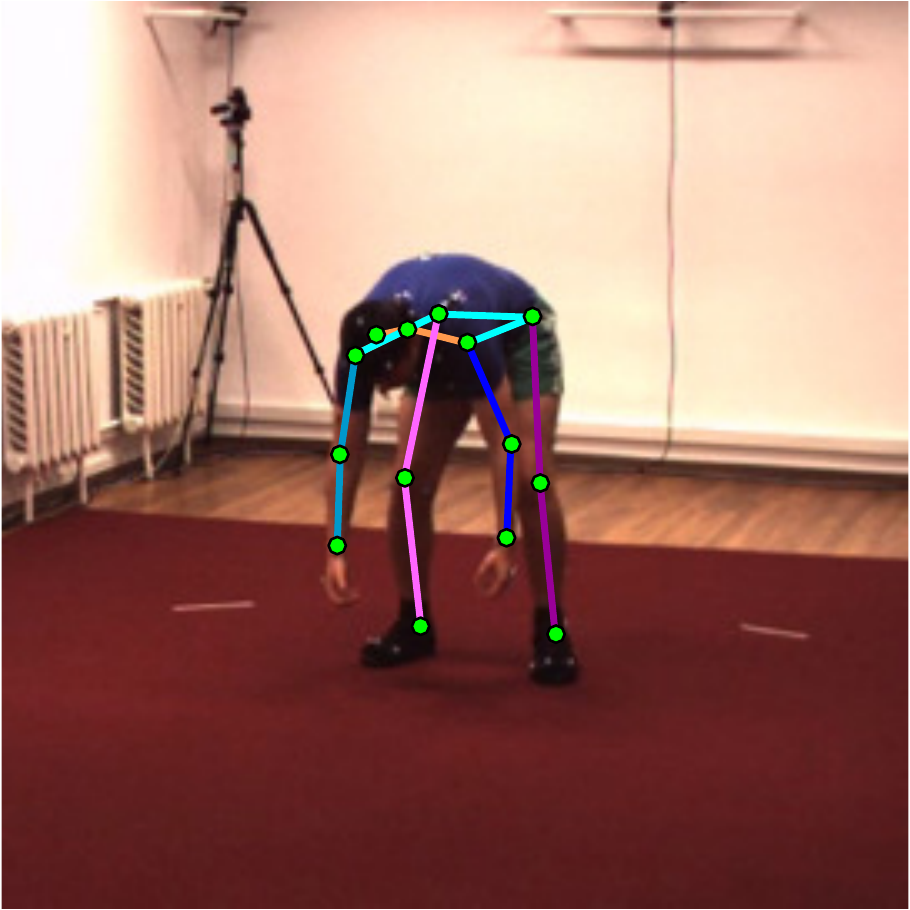} &
\includegraphics[height=0.15\linewidth]{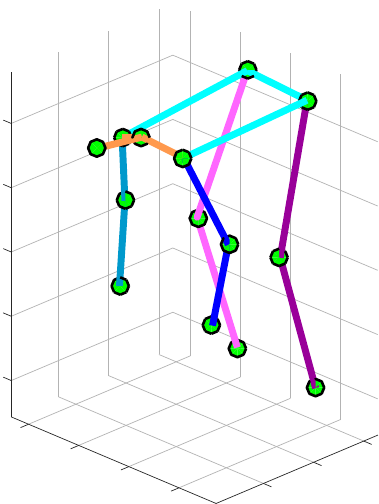} &
\includegraphics[height=0.15\linewidth]{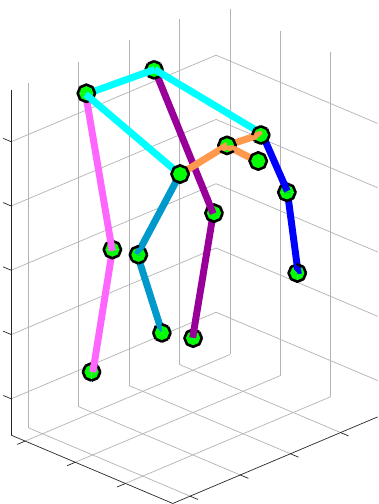} &
\includegraphics[height=0.13\linewidth]{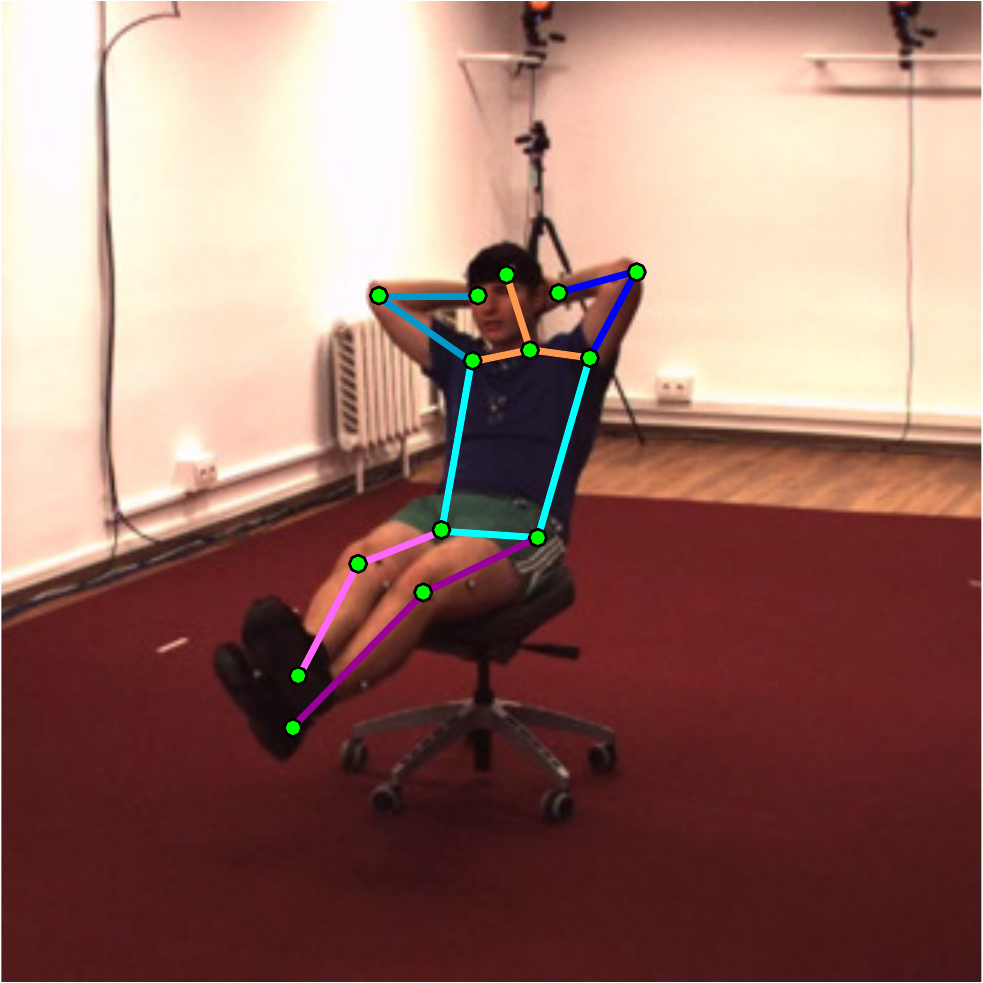} &
\includegraphics[height=0.15\linewidth]{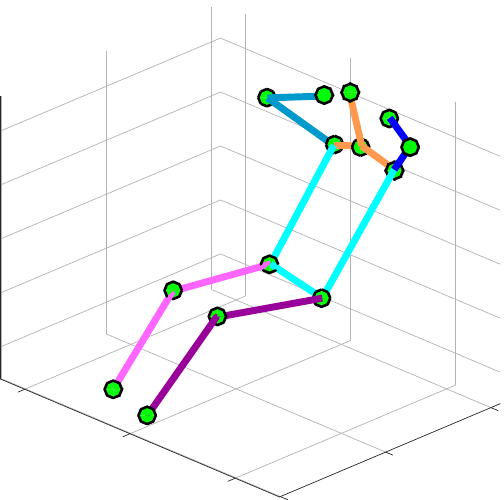} &
\includegraphics[height=0.15\linewidth]{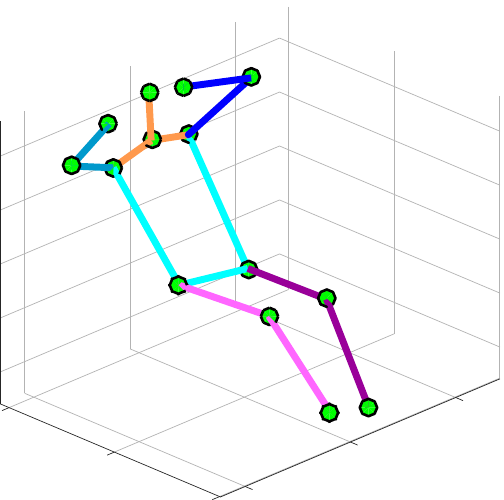} \\
\includegraphics[height=0.13\linewidth]{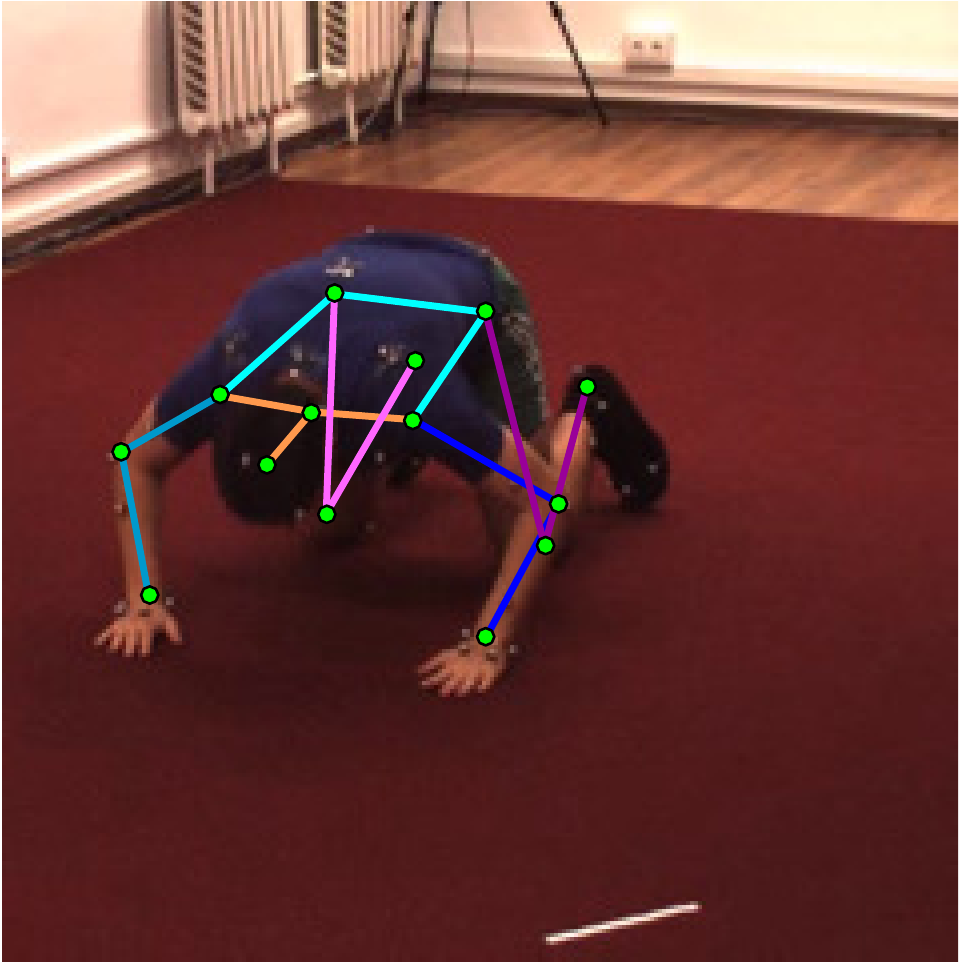} &
\includegraphics[height=0.15\linewidth]{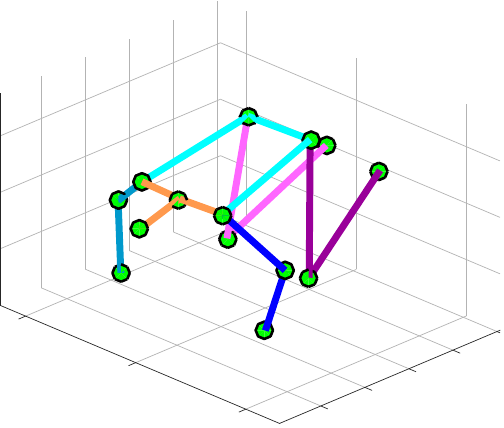} &
\includegraphics[height=0.15\linewidth]{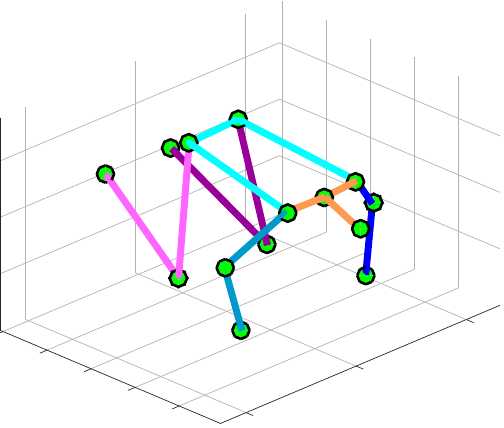} &
\includegraphics[height=0.13\linewidth]{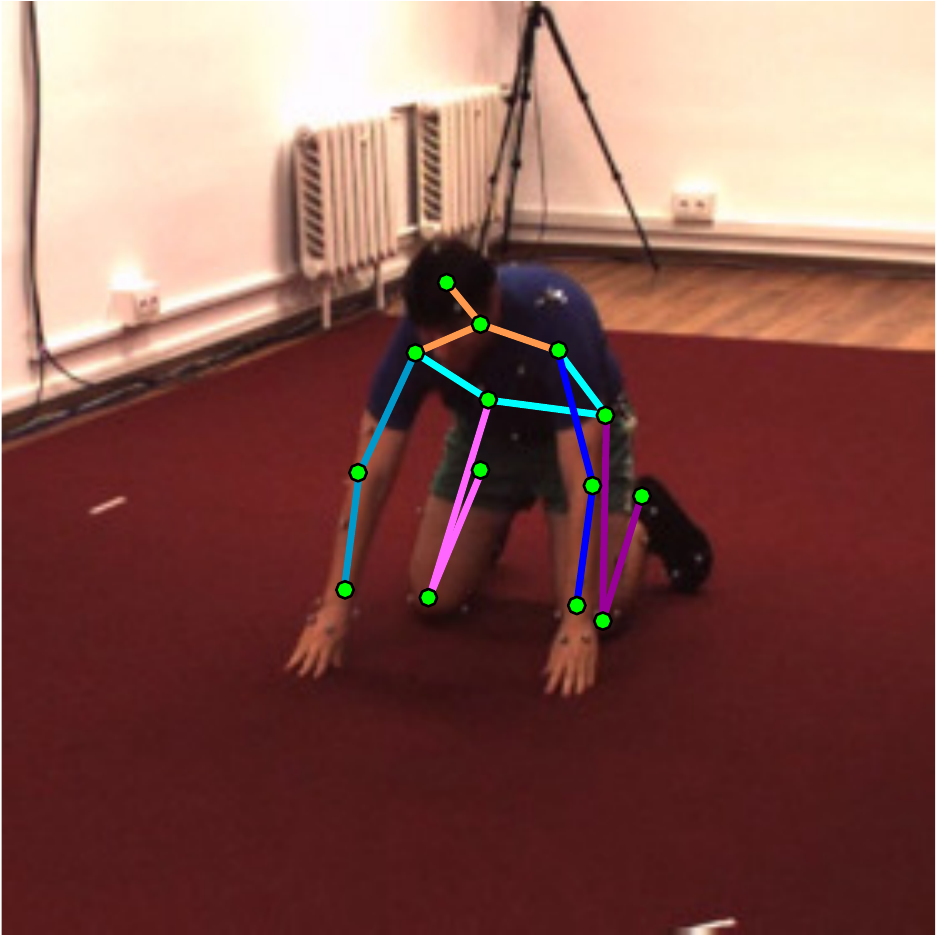} &
\includegraphics[height=0.15\linewidth]{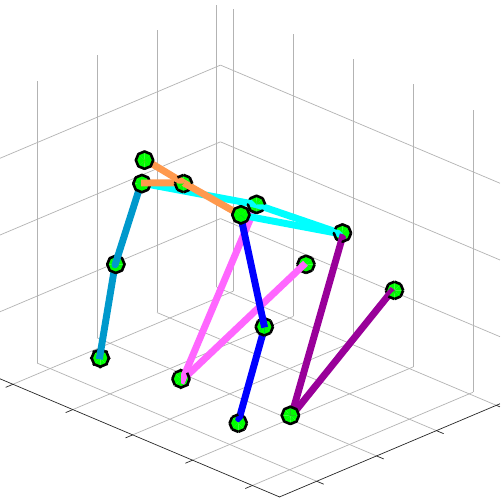} &
\includegraphics[height=0.15\linewidth]{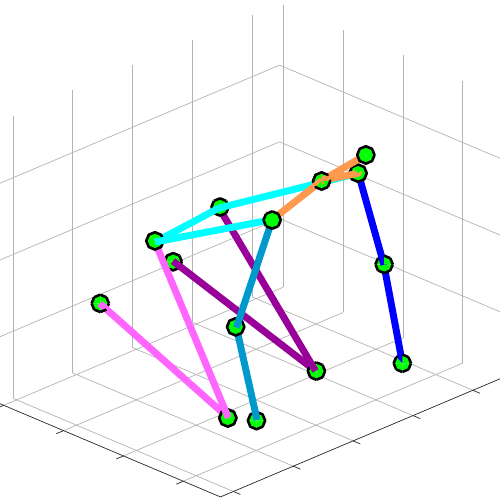} \\
\includegraphics[height=0.13\linewidth]{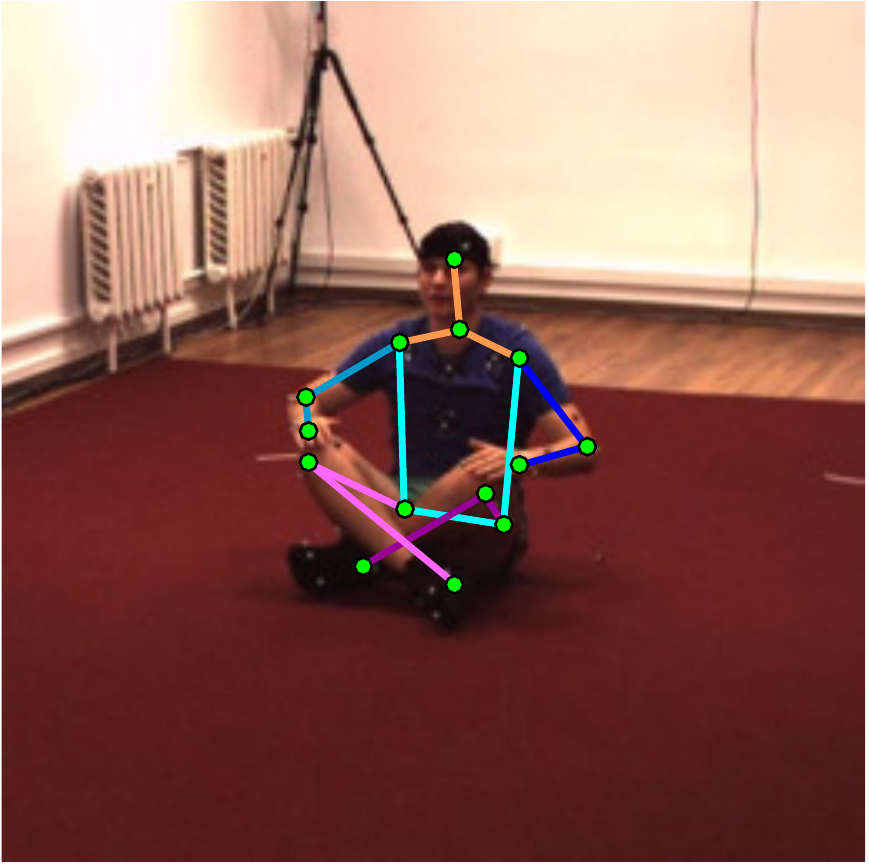} &
\includegraphics[height=0.15\linewidth]{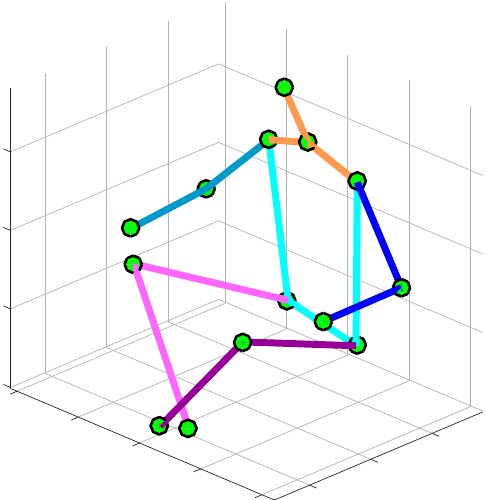} &
\includegraphics[height=0.15\linewidth]{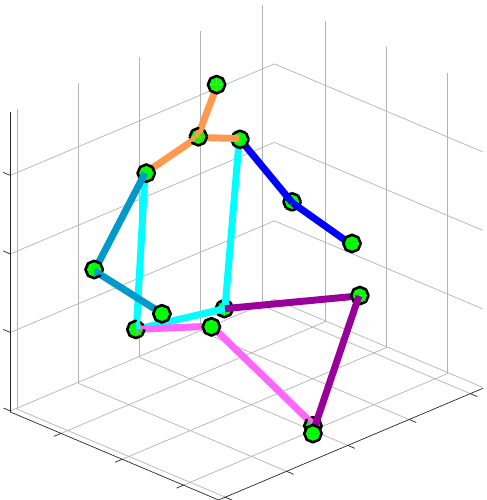} &
\includegraphics[height=0.13\linewidth]{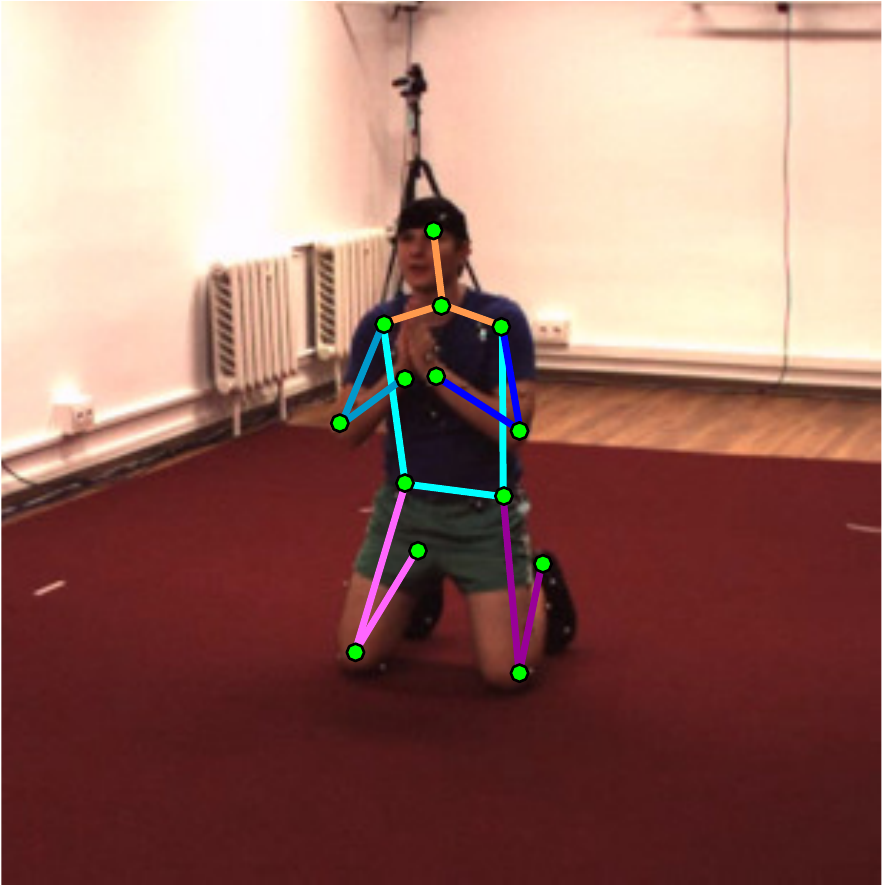} &
\includegraphics[height=0.15\linewidth]{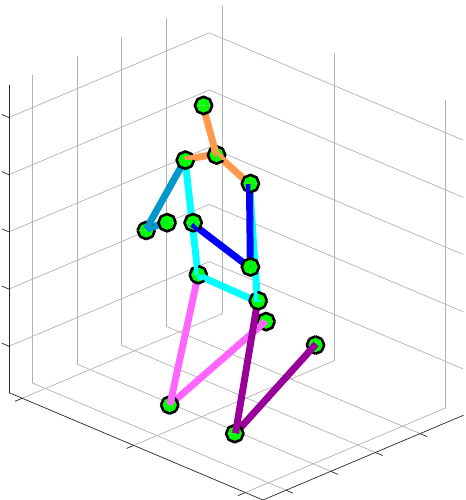} &
\includegraphics[height=0.15\linewidth]{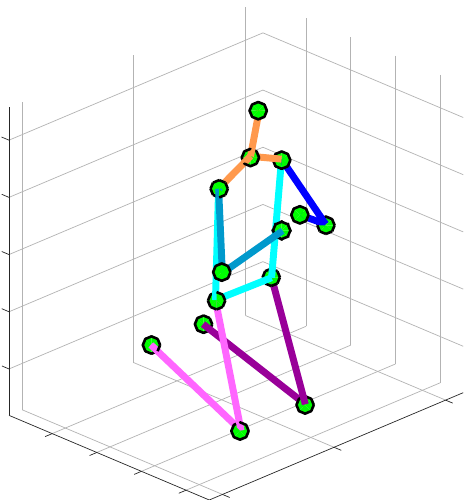} \\
\includegraphics[height=0.13\linewidth]{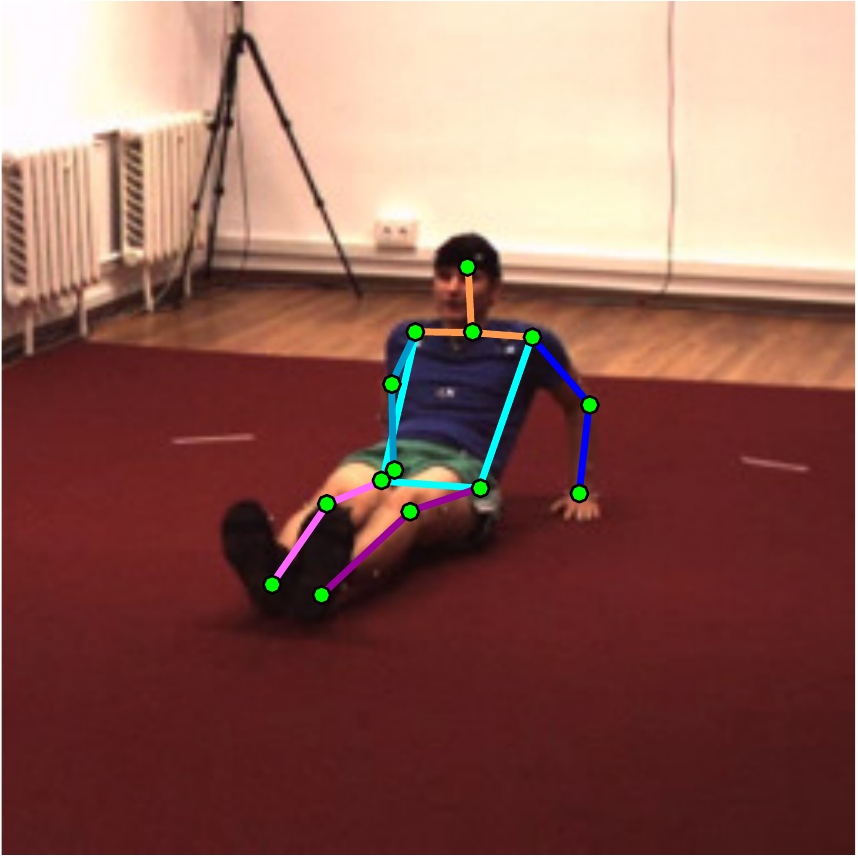} &
\includegraphics[height=0.15\linewidth]{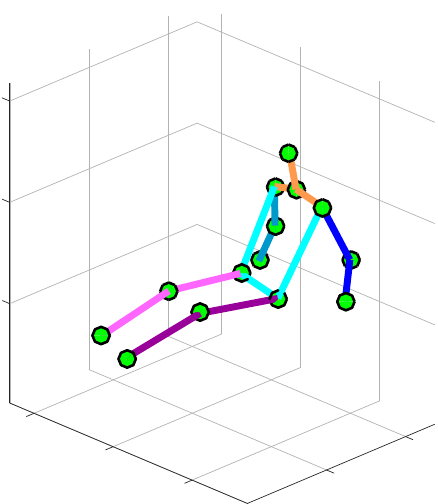} &
\includegraphics[height=0.15\linewidth]{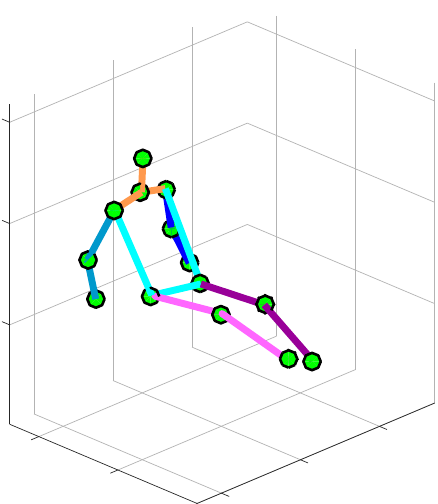} &
\includegraphics[height=0.13\linewidth]{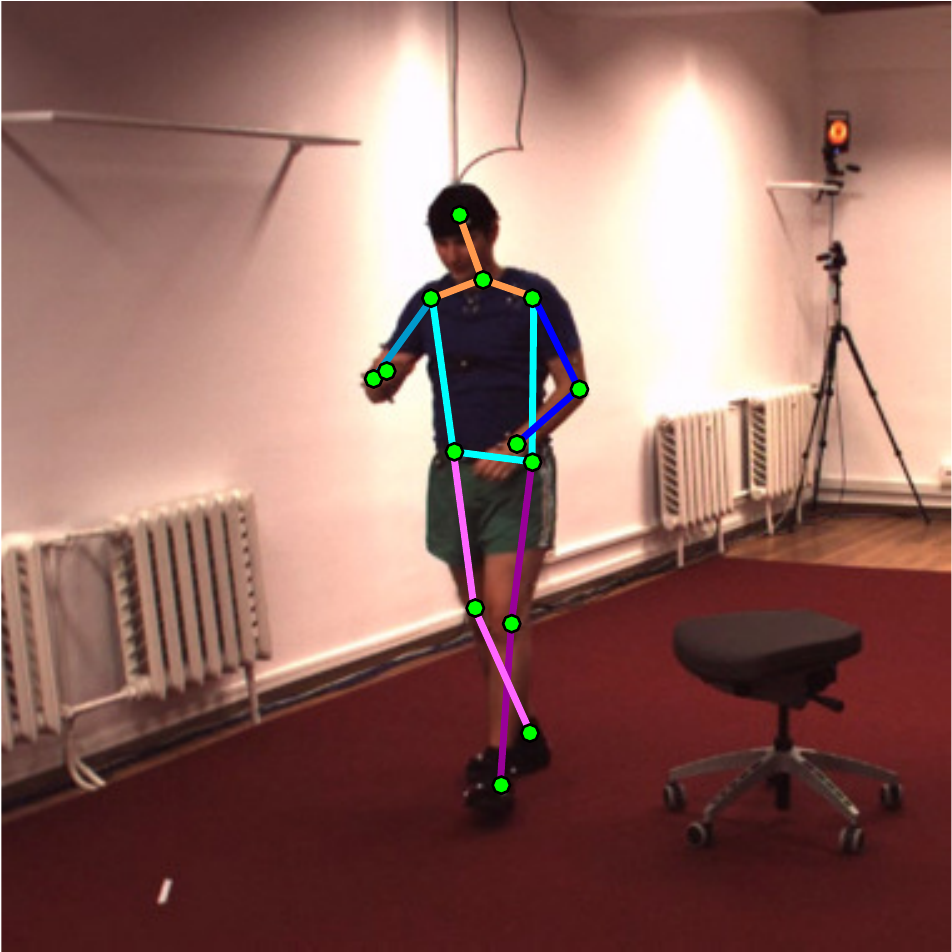} &
\includegraphics[height=0.15\linewidth]{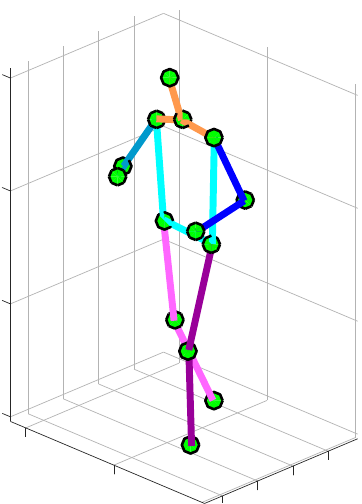} &
\includegraphics[height=0.15\linewidth]{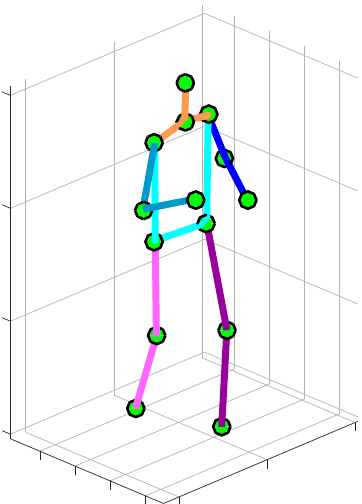} \\
\includegraphics[height=0.13\linewidth]{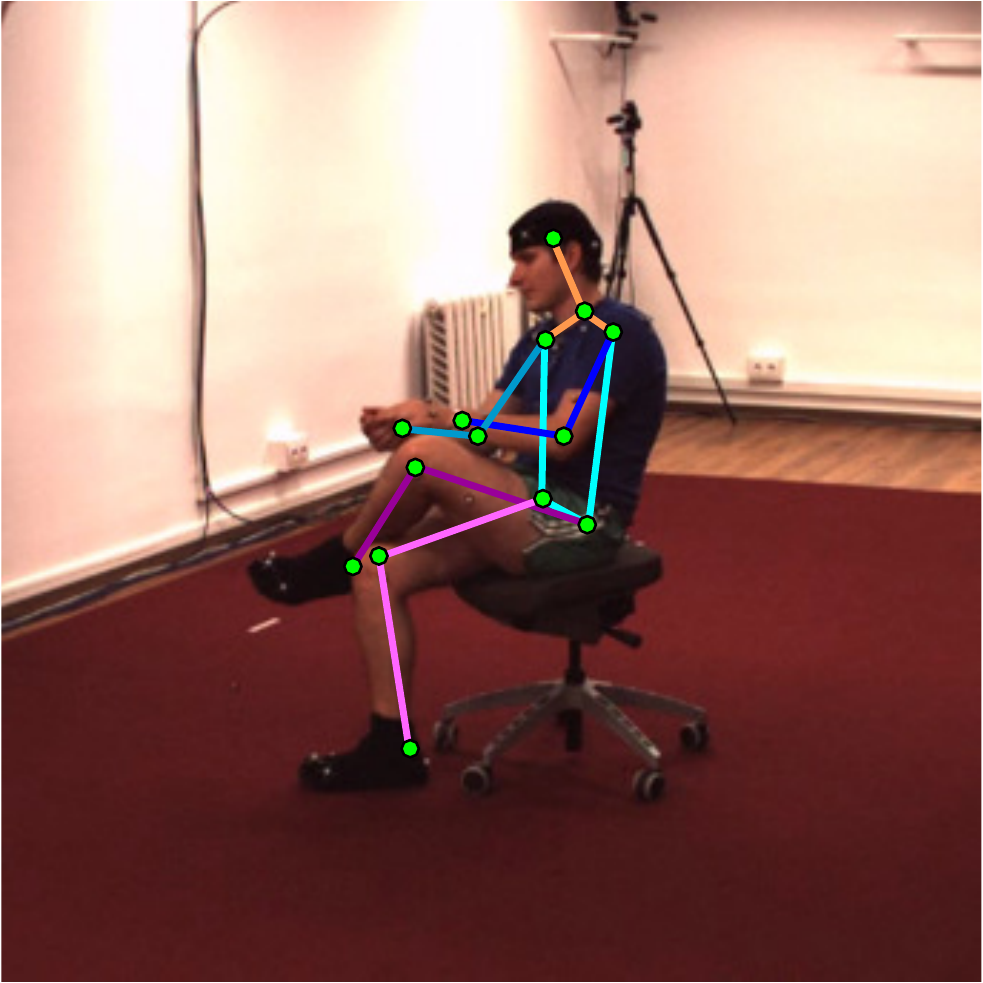} &
\includegraphics[height=0.15\linewidth]{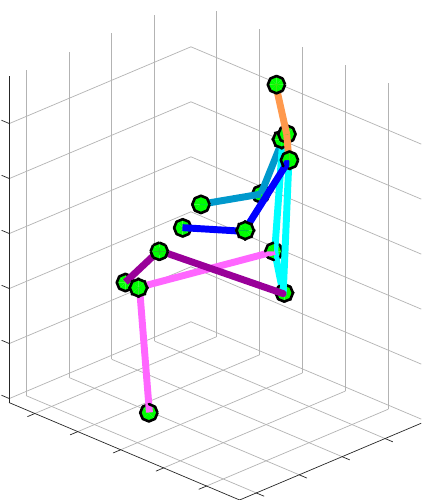} &
\includegraphics[height=0.15\linewidth]{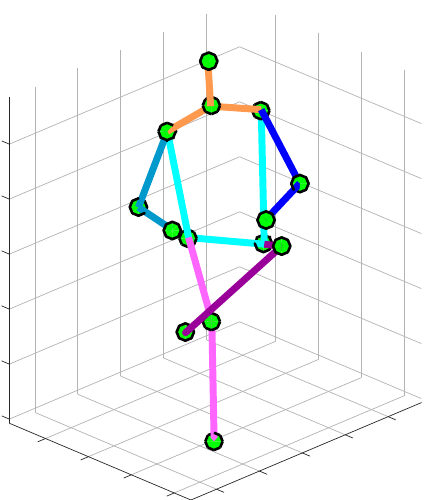} &
\includegraphics[height=0.13\linewidth]{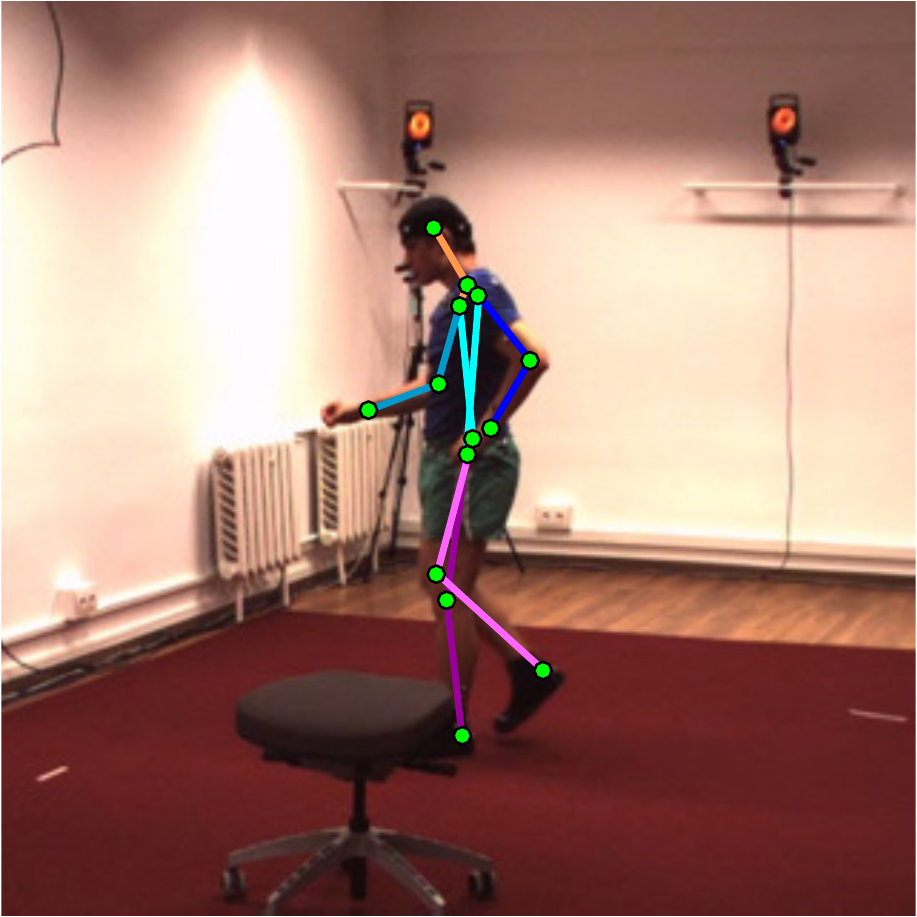} &
\includegraphics[height=0.15\linewidth]{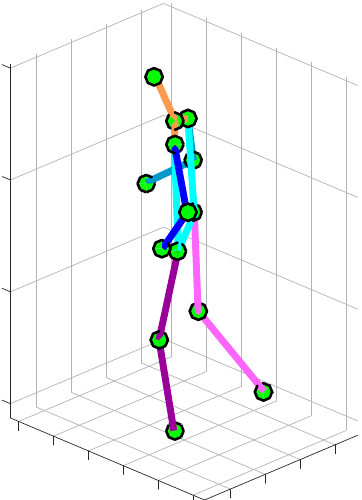} &
\includegraphics[height=0.15\linewidth]{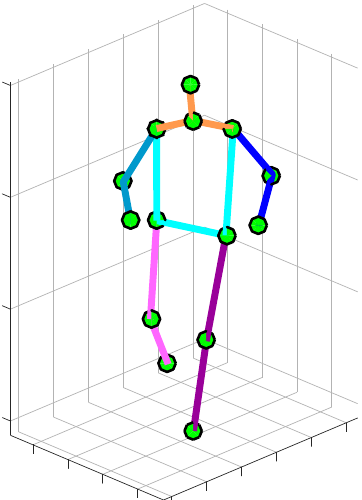} \\
\includegraphics[height=0.13\linewidth]{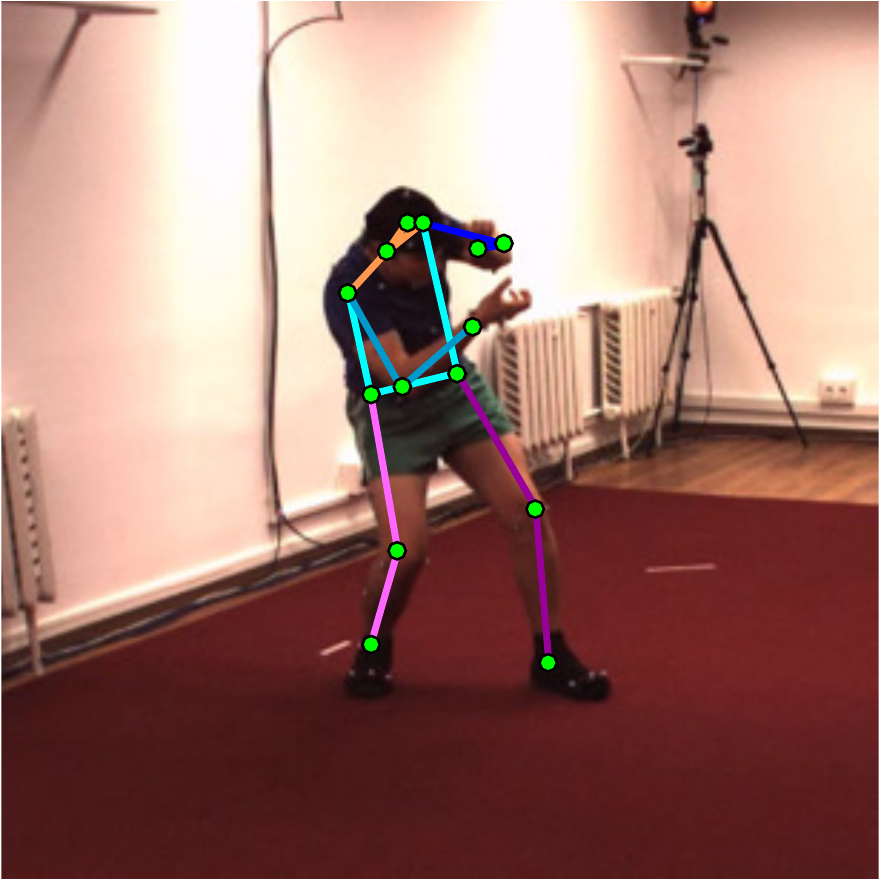} &
\includegraphics[height=0.15\linewidth]{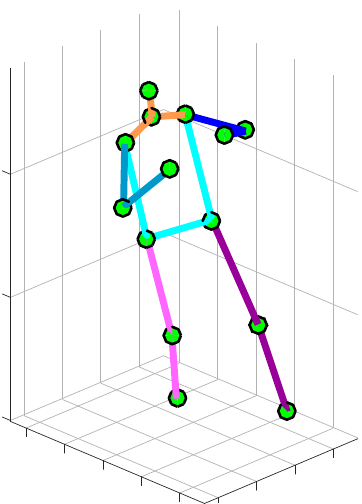} &
\includegraphics[height=0.15\linewidth]{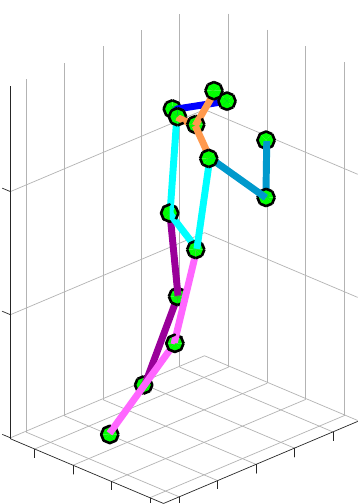} &
\includegraphics[height=0.13\linewidth]{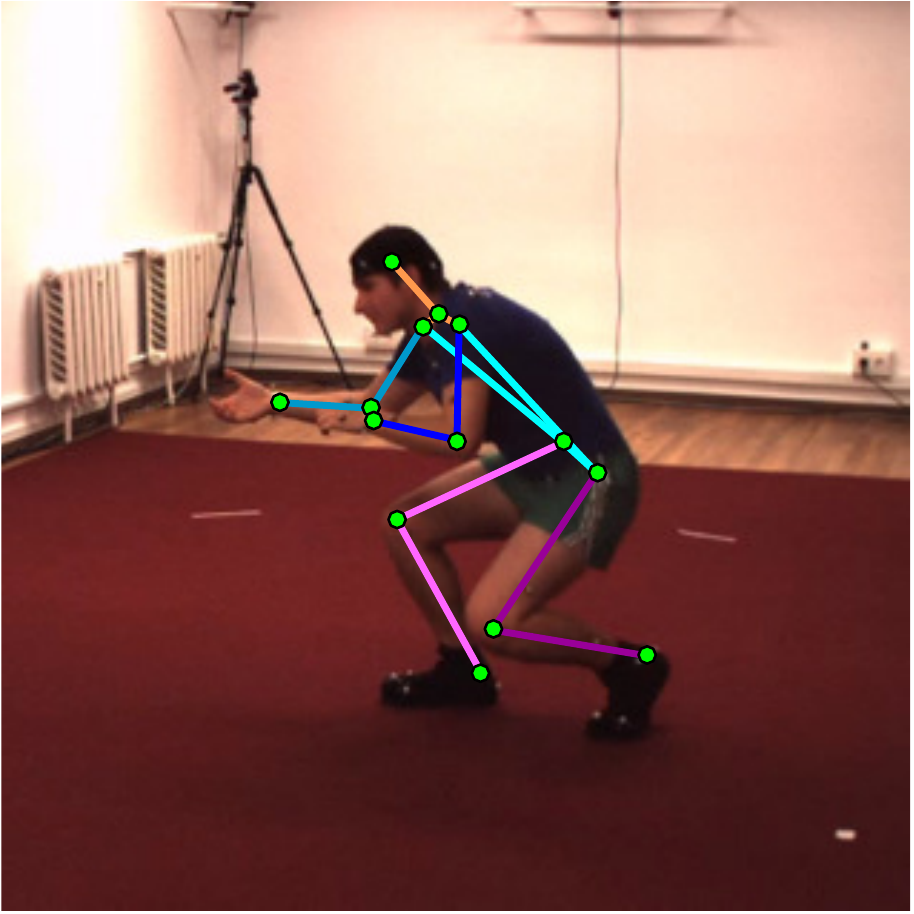} &
\includegraphics[height=0.15\linewidth]{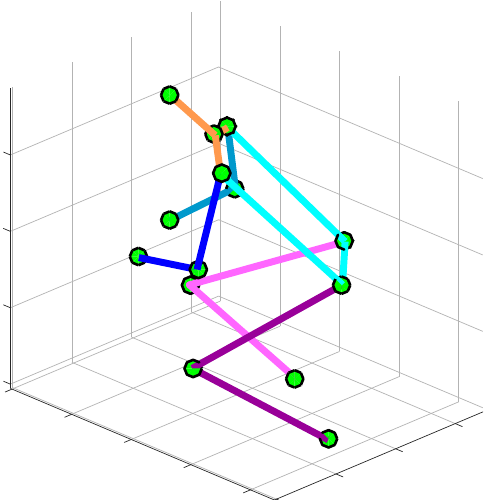} &
\includegraphics[height=0.15\linewidth]{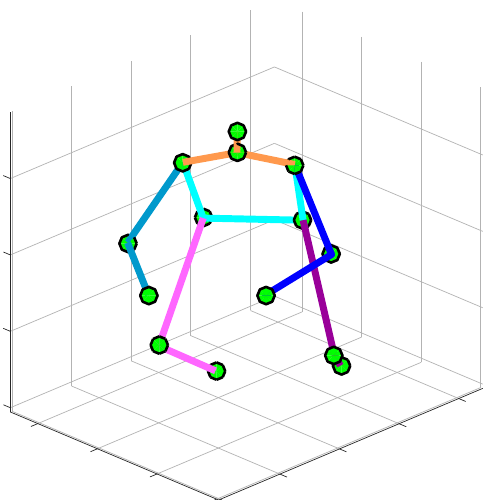} \\
\end{tabular}
}
\vspace{1em}
\caption{Some qualitative results from the Human3.6M \cite{h36m_pami} dataset.}
\label{fig:qualitative_h36m}
\end{figure*}

Finally, we present some qualitative results in Fig. \ref{fig:qualitative_h36m}. As it can be seen, our approach shows very good
performance even for highly articulated poses and under severe occlusions. 

%
%

\subsection{Evaluation on HumanEva-I Dataset}\label{sec:EvaI}
We follow the same protocol as described in~\cite{SimoSerraCVPR2013,Ilya_2014} and use the provided training data to train our approach while 
using the validation data as test set. As in~\cite{SimoSerraCVPR2013,Ilya_2014}, we report our results on every $5^{th}$ frame of the 
sequences \textit{walking} (A1) and \textit{jogging} (A2) for all three subjects (S1, S2, S3) and camera C1. The 3D pose error is computed as in \emph{Protocol-I} for the Human3.6M dataset. 

We perform experiments with the 3D pose data from the HumanEva and CMU MoCap datasets. For HumanEva, we use the entire 49K 3D poses of the training data as MoCap dataset. Since the joint positions of skeleton used for HumanEva differs from the marked joint positions used for the MPII Human Pose dataset, we fine-tune the 2D pose estimation model on the HumanEva dataset using the provided 2D pose data.  
For fine-tuning, we run $500$ iterations with a learning rate of $0.00008$. 

We also have to adapt the skeleton structure of the CMU dataset to the skeleton structure of the HumanEva dataset. As in~\cite{Yasin_2016_CVPR}, we re-target the 3D poses in the CMU dataset to the skeleton of the HumaEva dataset using linear regression. 
To this end, we first search for each 3D pose in the CMU dataset the nearest neighbor in the HumanEva dataset and then select pairs of nearest neighbors that have a distance less than a certain threshold. The selected pairs are then used to learn a linear regressor for each joint. We analyze the impact of the difference between the skeletons of both datasets in Tab.~\ref{tab:results_impact_of_skeletons}. 
Using HumanEva as MoCap dataset results in a 3D pose error of $31.5$mm, whereas using CMU as MoCap dataset increases the error significantly to $80.0$mm.
Re-targeting the skeletons of the CMU dataset to the skeleton of HumanEva reduces the error from $80.0$mm to $50.5$mm, and re-targeting the skeleton of HumanEva to CMU increases the 
error from $31.5$mm to $58.4$mm. This shows that the difference of the skeleton structure between the the two sources can have a major impact on the evaluation. This is, however, not an issue for an application where the MoCap dataset defines the skeleton structure.   

We also compare our approach with the state-of-the-art approaches 
\cite{Ilya_2014,Wang_2014_CVPR,Radwan-2013iccv,SimoSerraCVPR2013,SimoSerraCVPR2012,Bo-2010,Yasin_2016_CVPR,Moreno_arxiv2016} in Tab.~\ref{tab:results_HE}. 
Our method outperforms all other methods except of the two recent approaches~\cite{Moreno_arxiv2016,popa2017CVPRmultitask}. Both approaches use pairs of images and 3D poses to learn a neural network model, while our approach considers them as independent sources and is therefore trained with less supervision.     

\begin{table}[t]
\centering
\scalebox{1}{
\begin{tabularx}{\columnwidth}{l|sss|sss|t}
\toprule
\multicolumn{1}{c}{\multirow{2}{*}{MoCap Data}} & \multicolumn{3}{c}{Walking (A1, C1)}          & \multicolumn{3}{c}{Jogging (A2, C1)}  & \multicolumn{1}{c}{\multirow{2}{*}{Average}}   \\ 
\multicolumn{1}{c}{}            & \multicolumn{1}{c}{S1} & \multicolumn{1}{c}{S2} & \multicolumn{1}{c}{S3} & \multicolumn{1}{c}{S1} & \multicolumn{1}{c}{S2} & \multicolumn{1}{c}{S3} & \\ 
\midrule
HumanEva			& 27.4  	&  28.6		& 32.5  &  39.9  & 29.4 &	31.4  &	31.5 \\
CMU 	 			& 68.4  	&	81.6  	&	88.3 &	70.1  &	81.6  &	89.9  &	80.0   \\
CMU $\rightarrow$ HumanEva	& 39.5  	 	&	47.3  	&	61.4  &	53.5  &	48.3 &	53.1  &	50.5  \\
HumanEva $\rightarrow$ CMU	& 45.1  		&	54.9  &	59.1  	&	58.6  &	63.1 &	69.7 &	58.4  \\

\bottomrule
\end{tabularx}
}
\caption{Impact of different skeleton structures. The symbol $\rightarrow$ indicates retargeting of the skeleton structure of one dataset to the skeleton of another dataset.}
\label{tab:results_impact_of_skeletons}
\end{table}

\begin{table*}[t]
\centering
\scalebox{1}{
\begin{tabularx}{0.8\linewidth}{l|sss|sss|s}
\toprule
\multicolumn{1}{c}{\multirow{2}{*}{Methods} } & \multicolumn{3}{c}{Walking (A1, C1)}            & \multicolumn{3}{c}{Jogging (A2, C1)}  & \multicolumn{1}{c}{\multirow{2}{*}{Average}}   \\ 

\multicolumn{1}{c}{}                  & \multicolumn{1}{c}{S1} & \multicolumn{1}{c}{S2} & \multicolumn{1}{c}{S3} & \multicolumn{1}{c}{S1} & \multicolumn{1}{c}{S2} & \multicolumn{1}{c}{S3} & \\ 
\midrule

Kostrikov \etal \cite{Ilya_2014}           & 44.0   & 30.9     & 41.7      & 57.2      & { 35.0}      & { 33.3}  & 40.3       \\ 
Wang \etal \cite{Wang_2014_CVPR}           & 71.9    & 75.7      & 85.3      & 62.6      & 77.7      & 54.4    & 71.3          \\ 
Radwan \etal \cite{Radwan-2013iccv}     & 75.1   & 99.8  & 93.8  & 79.2   & 89.8      & 99.4  & 89.5 \\ 
Simo-Serra \etal \cite{SimoSerraCVPR2013}  & 65.1    & 48.6      & 73.5      & 74.2     & 46.6     & 32.2   & 56.7           \\ 
Simo-Serra \etal \cite{SimoSerraCVPR2012}  & 99.6   & 108.3     & 127.4     & 109.2     & 93.1     & 115.8 & 108.9         \\ 
Bo \etal \cite{Bo-2010}*  & 38.2    & 32.8      & 40.2      & 42.0      & 34.7      & 46.4   & 39.1   \\ 
Yasin \etal \cite{Yasin_2016_CVPR} & { 35.8}     &  32.4    &  { 41.6}   & { 46.6}   &  41.4     &  35.4           & { 38.9}  \\ 
Popa \etal \cite{popa2017CVPRmultitask} &  27.1 	& 18.4 	    &   39.5 	& 37.6 &   28.9 &  27.6 & 29.9 \\
Moreno-Noguer \cite{Moreno_arxiv2016} 	& \bf19.7 	& \bf13.5   & 	\bf26.5 & \bf34.6 &	 \bf17.9  & \bf20.1   & \bf22.0\\
\textbf{Ours} 			   	& 27.4  	& 28.6  &	32.5  &	39.9  &	29.4  &	31.4  &	31.5  \\

\midrule

\multicolumn{8}{c}{MoCap from CMU dataset} \\ 
\midrule
Yasin \etal \cite{Yasin_2016_CVPR}                   & 52.2    &  51.0    &  62.8    & 74.5   &  72.4     & 56.8            & 61.6   \\ 
Ours &	\bf39.5  &	\bf47.3  &	\bf61.4  &	\bf53.5  &	\bf48.3 &	\bf53.1  &	\bf50.5  \\

\bottomrule
\end{tabularx}
}
\caption{Comparison with other state-of-the-art approaches on the HumanEva-I dataset. The average 3D pose error (mm) are reported for all three subjects (S1, S2, S3) and camera C1. * denotes a different evaluation protocol.
}
\label{tab:results_HE}
\end{table*}

\begin{figure*}
\centering
\scalebox{0.95}{ 
\begin{tabular}{c c c | c c c}
2D Pose & View-1 & View-2 & 2D Pose & View-1 & View-2 \\
\includegraphics[height=0.13\linewidth]{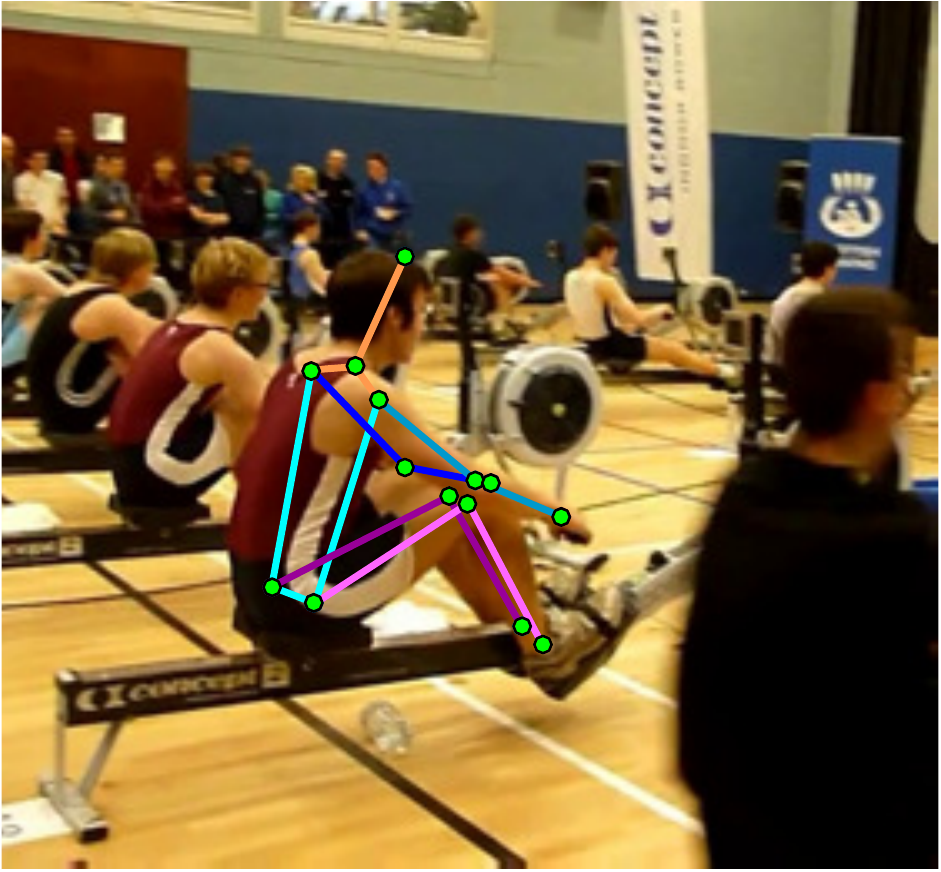} &
\includegraphics[height=0.15\linewidth]{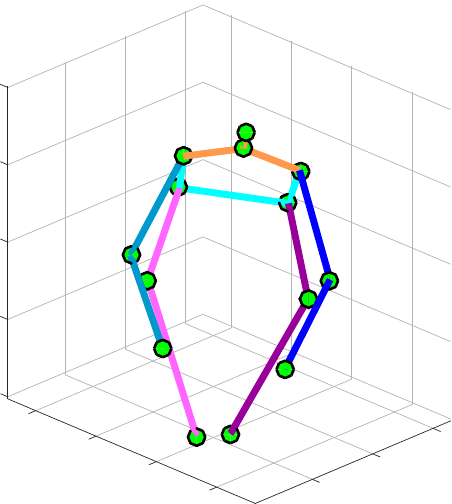} &
\includegraphics[height=0.15\linewidth]{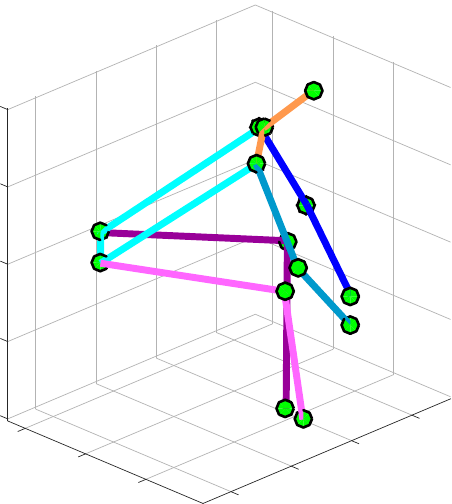} &
\includegraphics[height=0.13\linewidth]{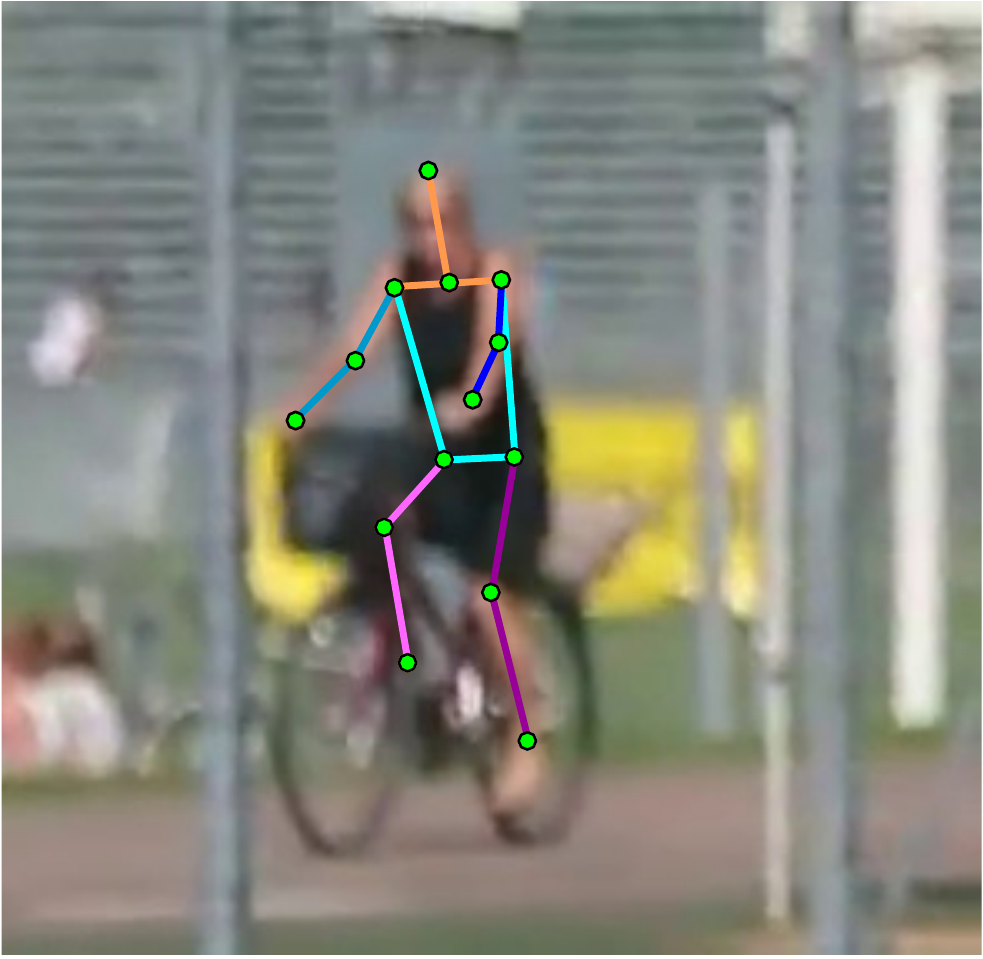} &
\includegraphics[height=0.15\linewidth]{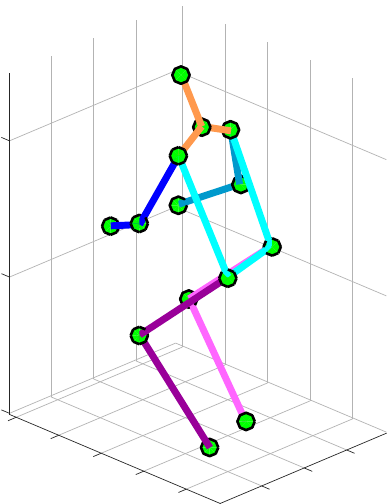} &
\includegraphics[height=0.15\linewidth]{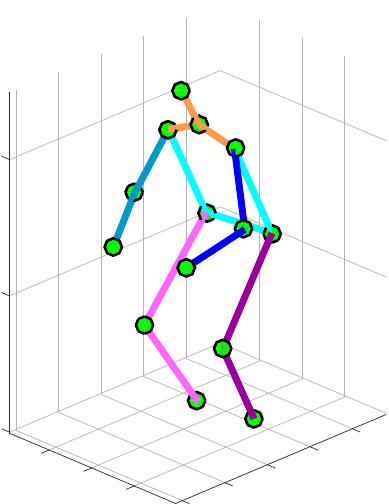} \\ 
\includegraphics[height=0.13\linewidth]{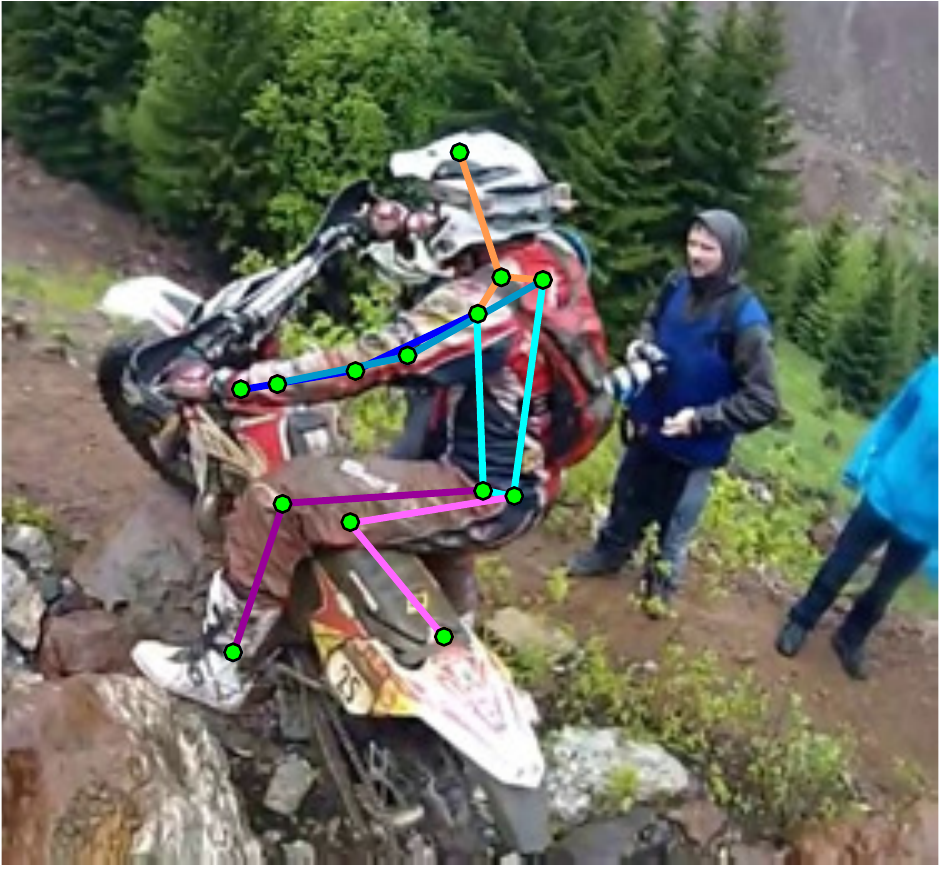} &
\includegraphics[height=0.15\linewidth]{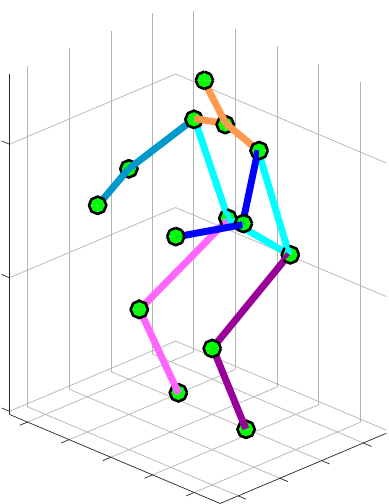} &
\includegraphics[height=0.15\linewidth]{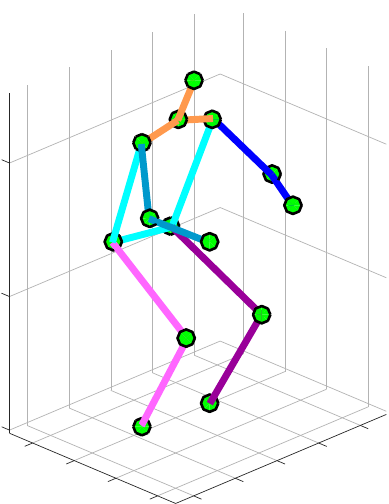} &
\includegraphics[height=0.13\linewidth]{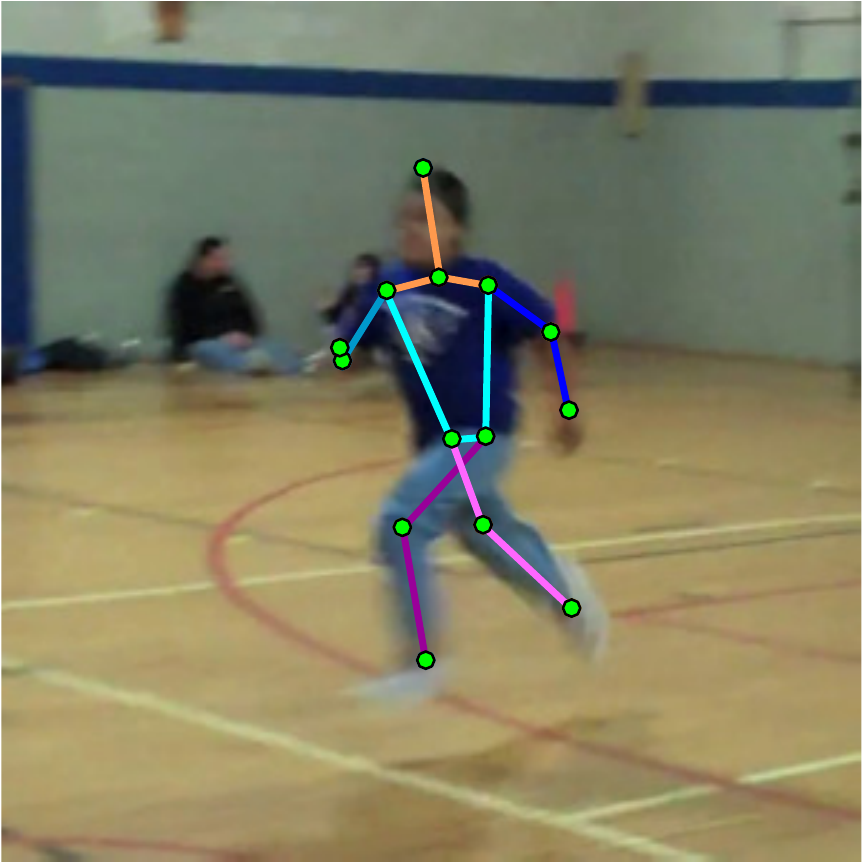} &
\includegraphics[height=0.15\linewidth]{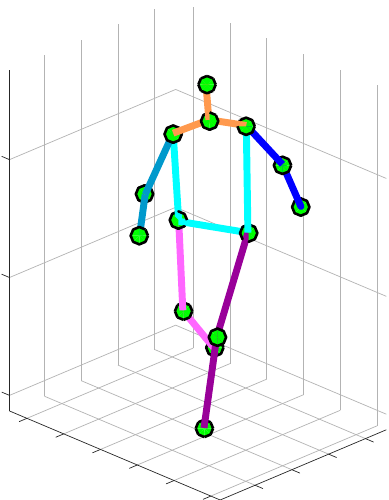} &
\includegraphics[height=0.15\linewidth]{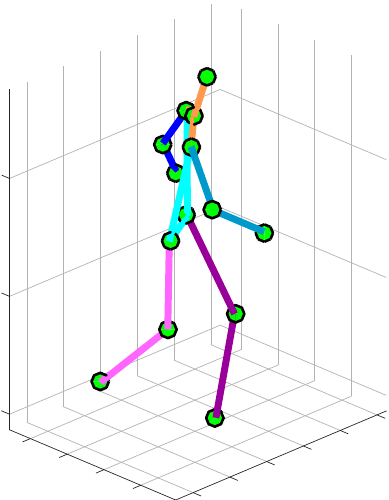} \\ 
\includegraphics[height=0.13\linewidth]{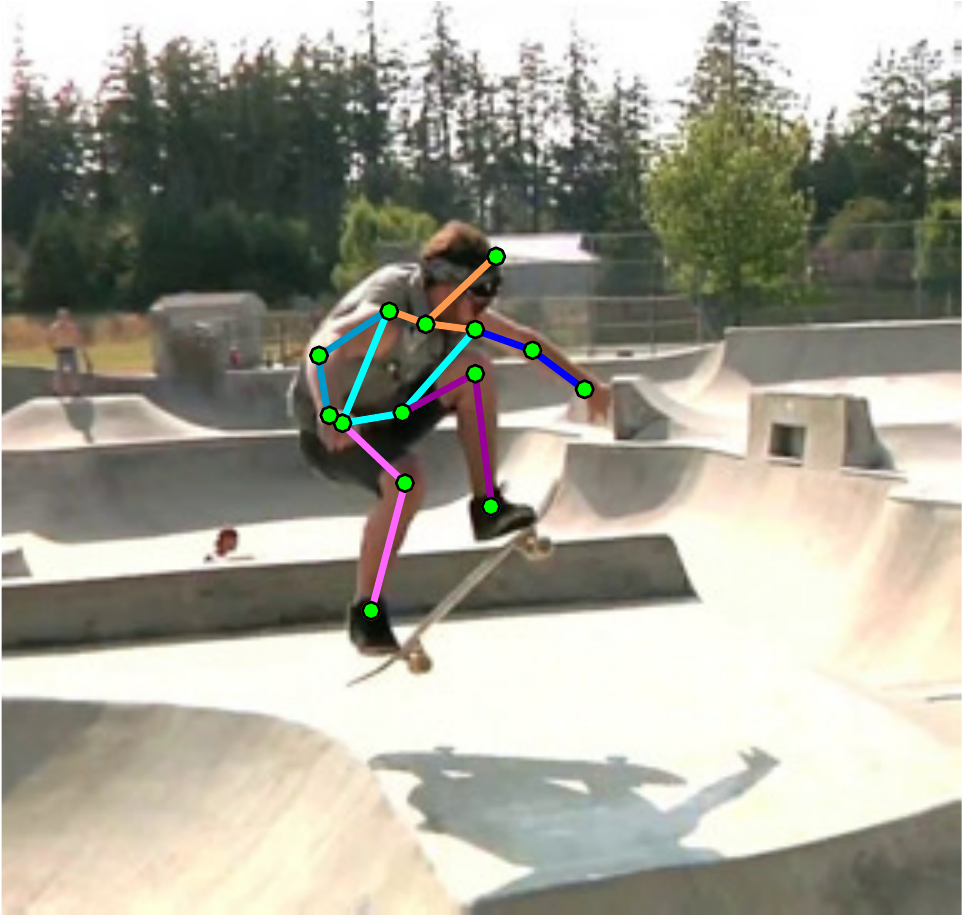} &
\includegraphics[height=0.15\linewidth]{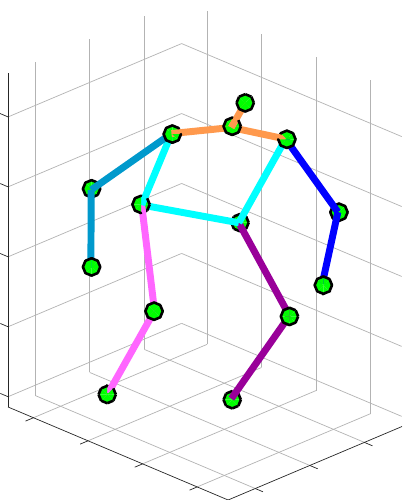} &
\includegraphics[height=0.15\linewidth]{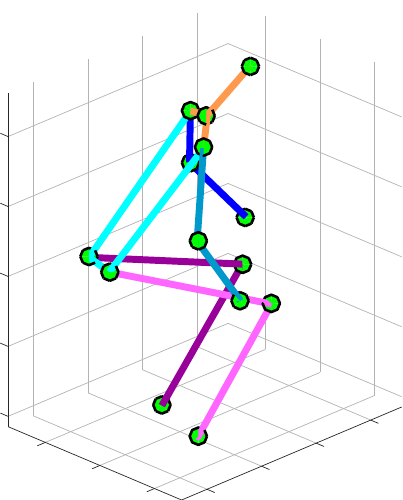} &
\includegraphics[height=0.13\linewidth]{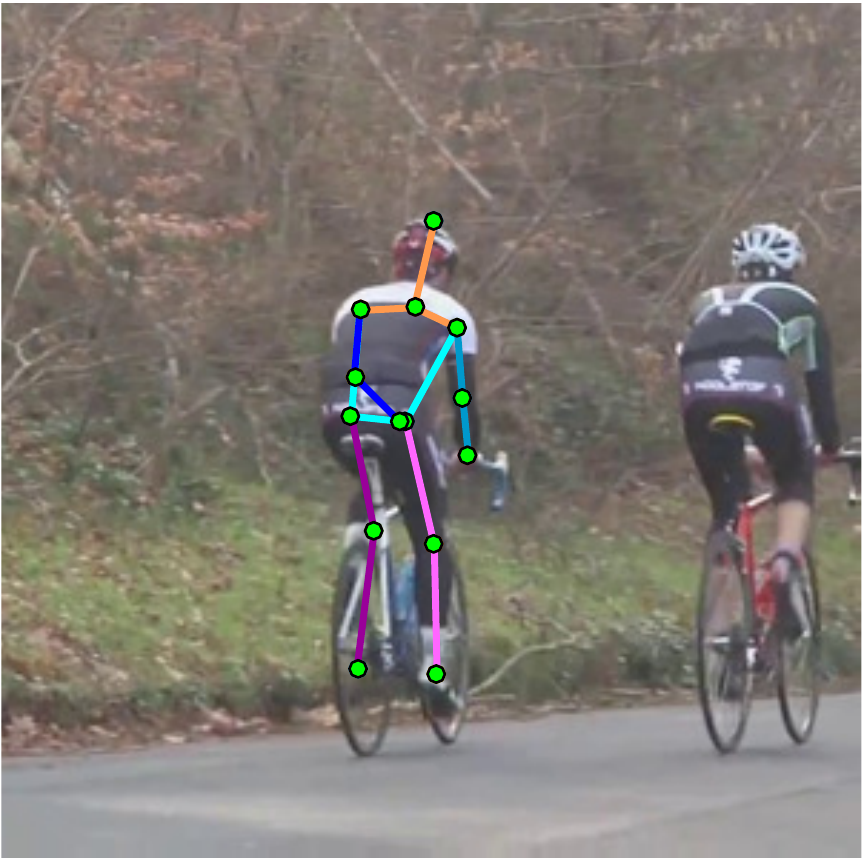} &
\includegraphics[height=0.15\linewidth]{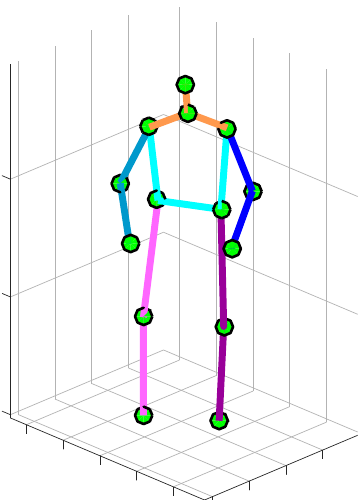} &
\includegraphics[height=0.15\linewidth]{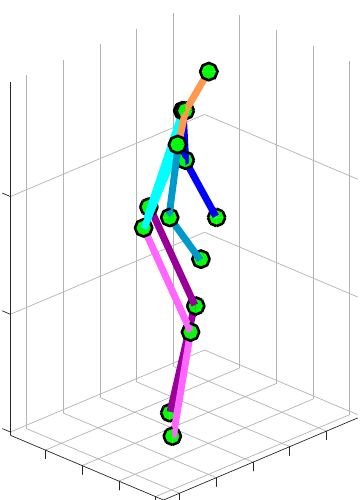} \\ 
\includegraphics[height=0.13\linewidth]{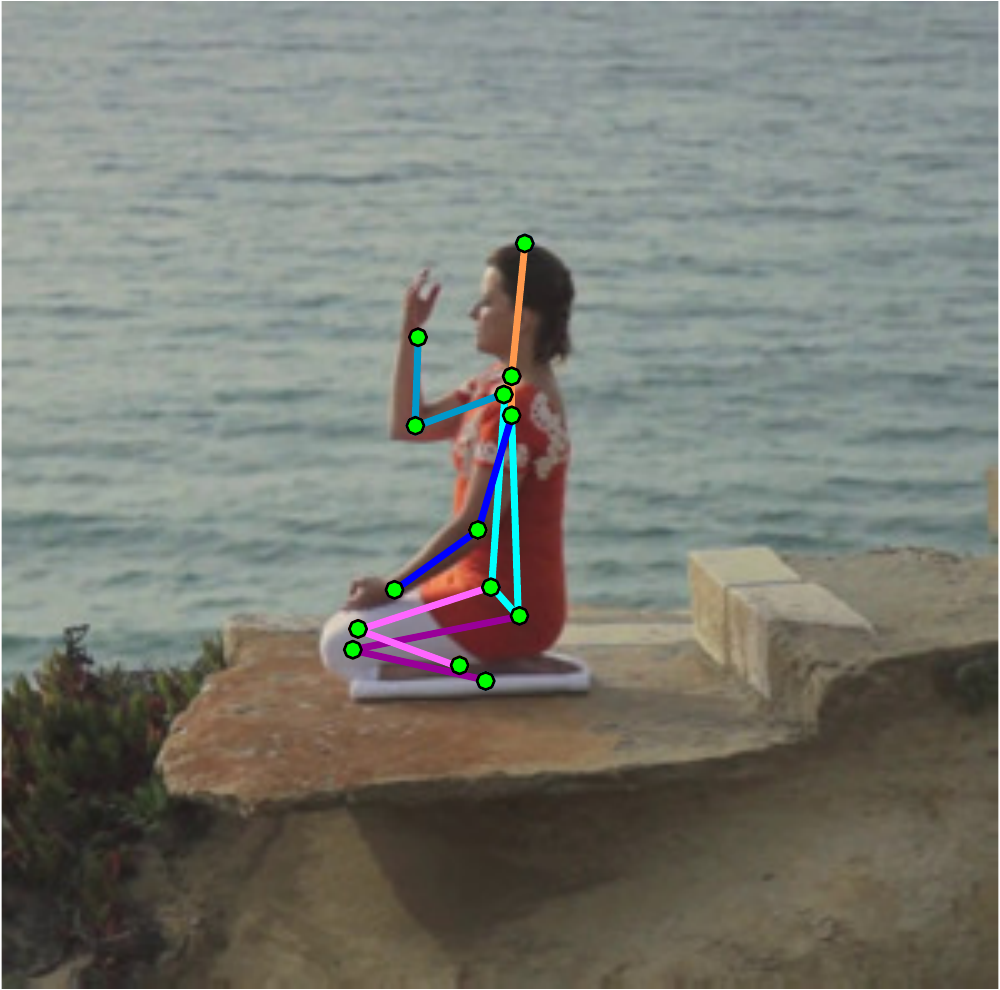} &
\includegraphics[height=0.15\linewidth]{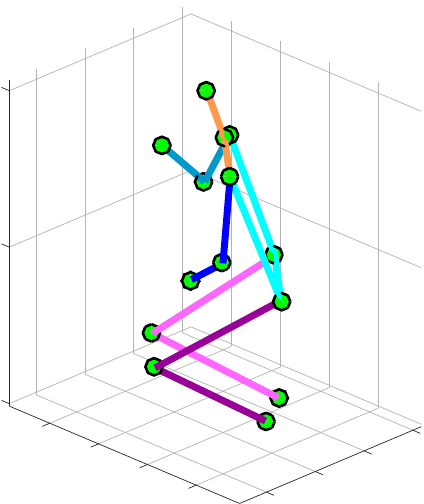} &
\includegraphics[height=0.15\linewidth]{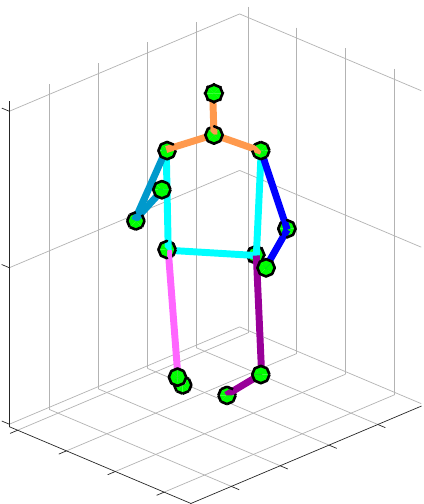} &
\includegraphics[height=0.13\linewidth]{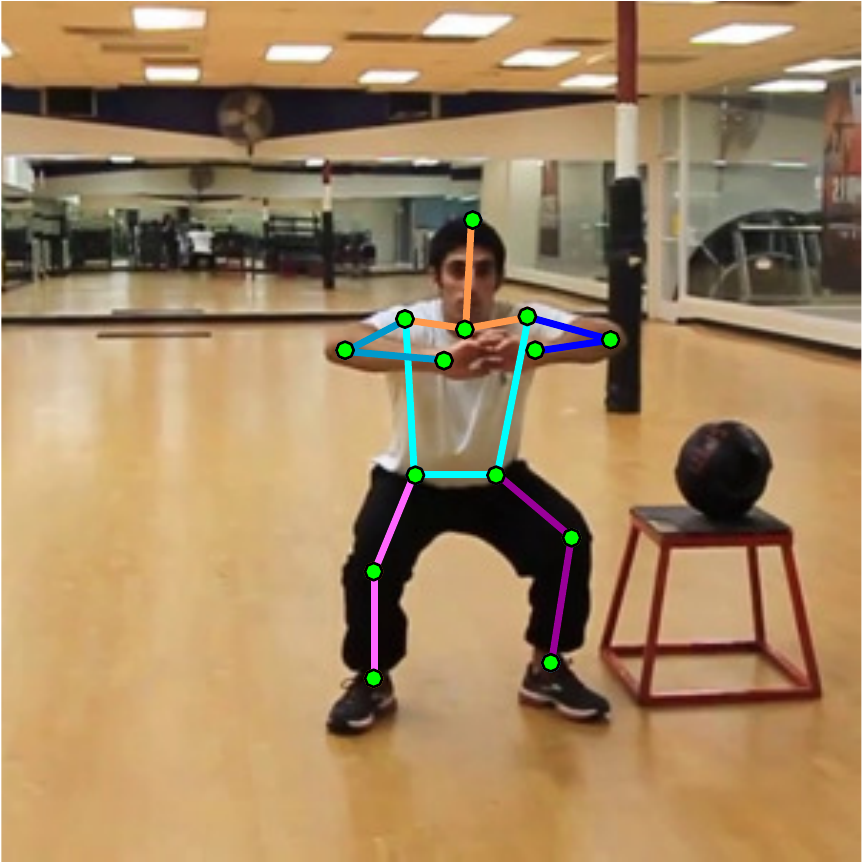} &
\includegraphics[height=0.15\linewidth]{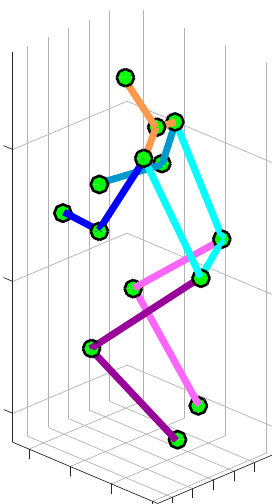} &
\includegraphics[height=0.15\linewidth]{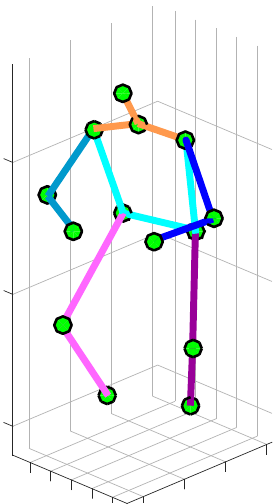} \\ 
\includegraphics[height=0.13\linewidth]{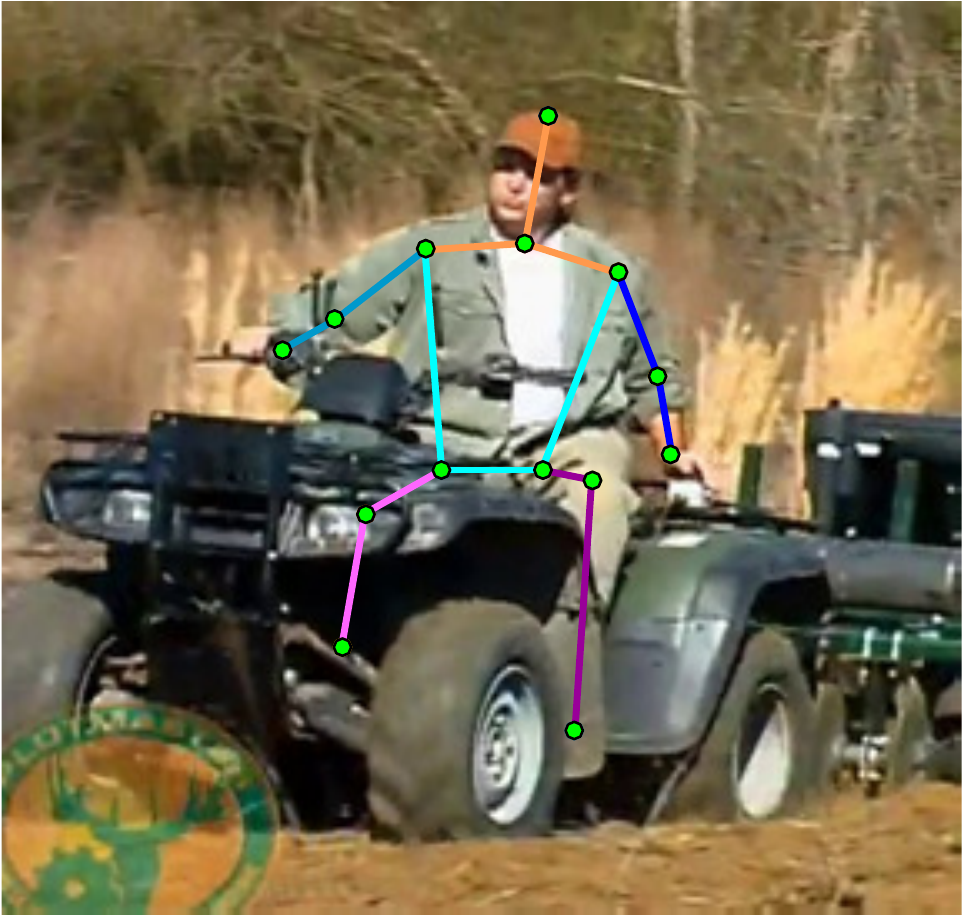} &
\includegraphics[height=0.15\linewidth]{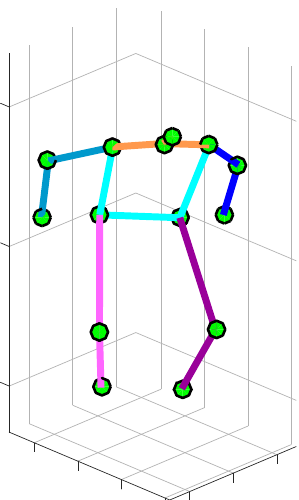} &
\includegraphics[height=0.15\linewidth]{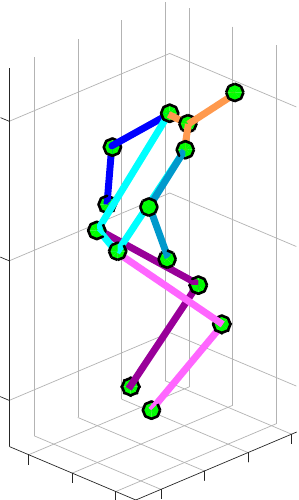} &
\includegraphics[height=0.13\linewidth]{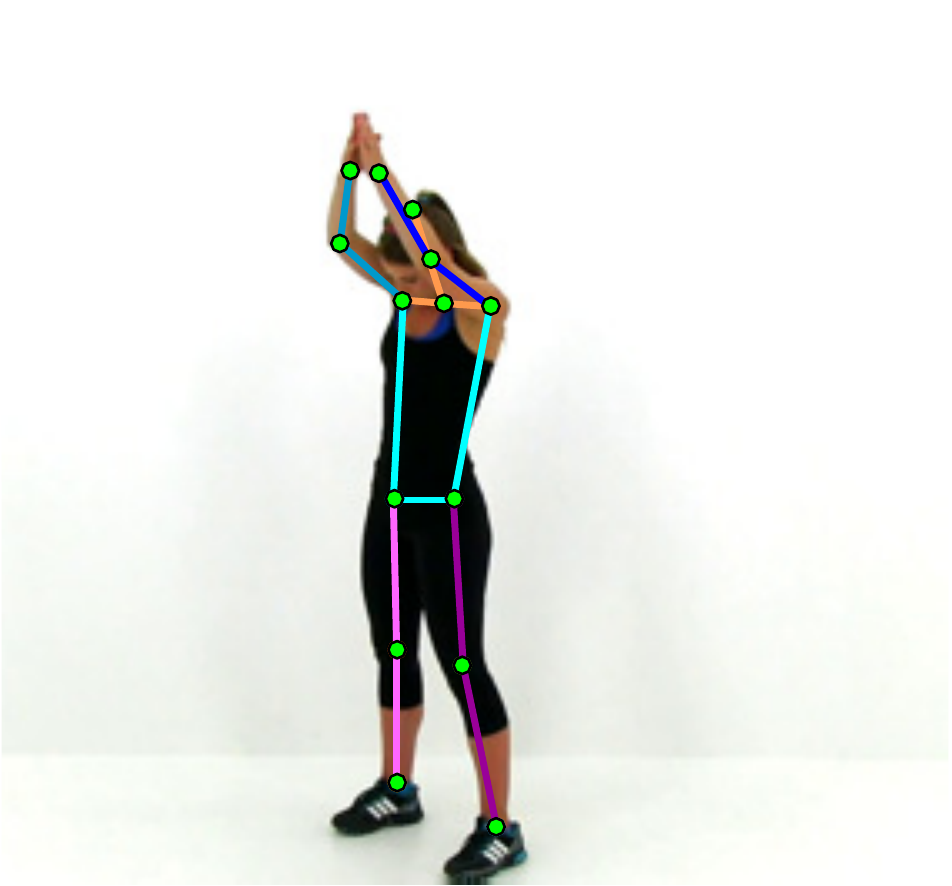} &
\includegraphics[height=0.15\linewidth]{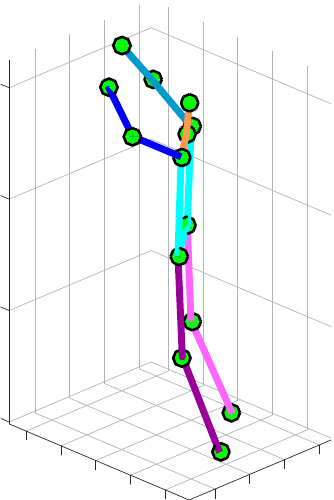} &
\includegraphics[height=0.15\linewidth]{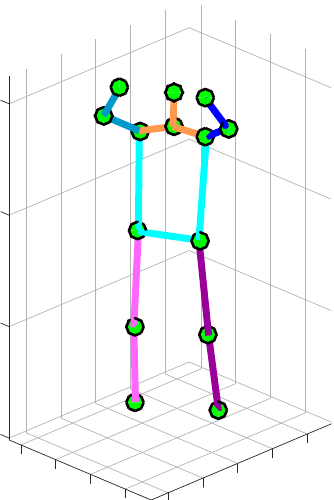} \\ 
\includegraphics[height=0.13\linewidth]{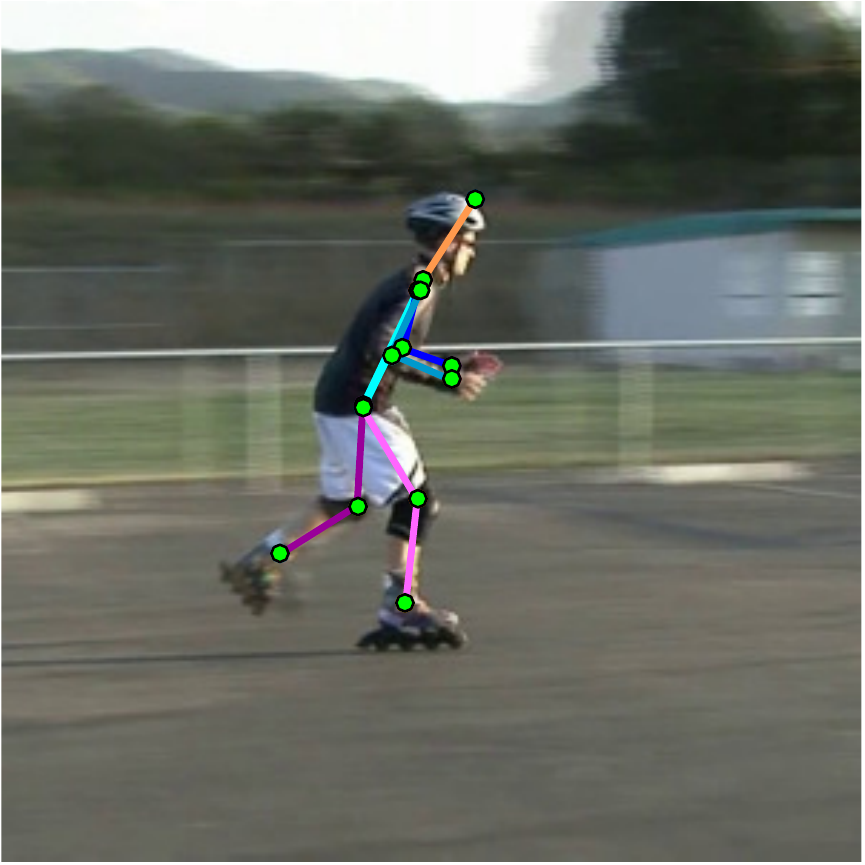} &
\includegraphics[height=0.15\linewidth]{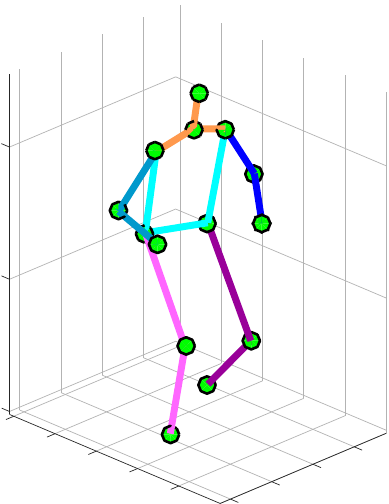} &
\includegraphics[height=0.15\linewidth]{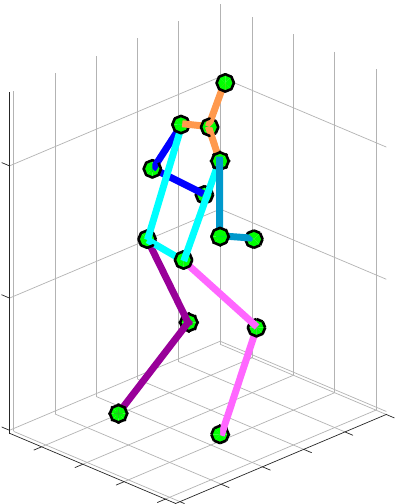} &
\includegraphics[height=0.13\linewidth]{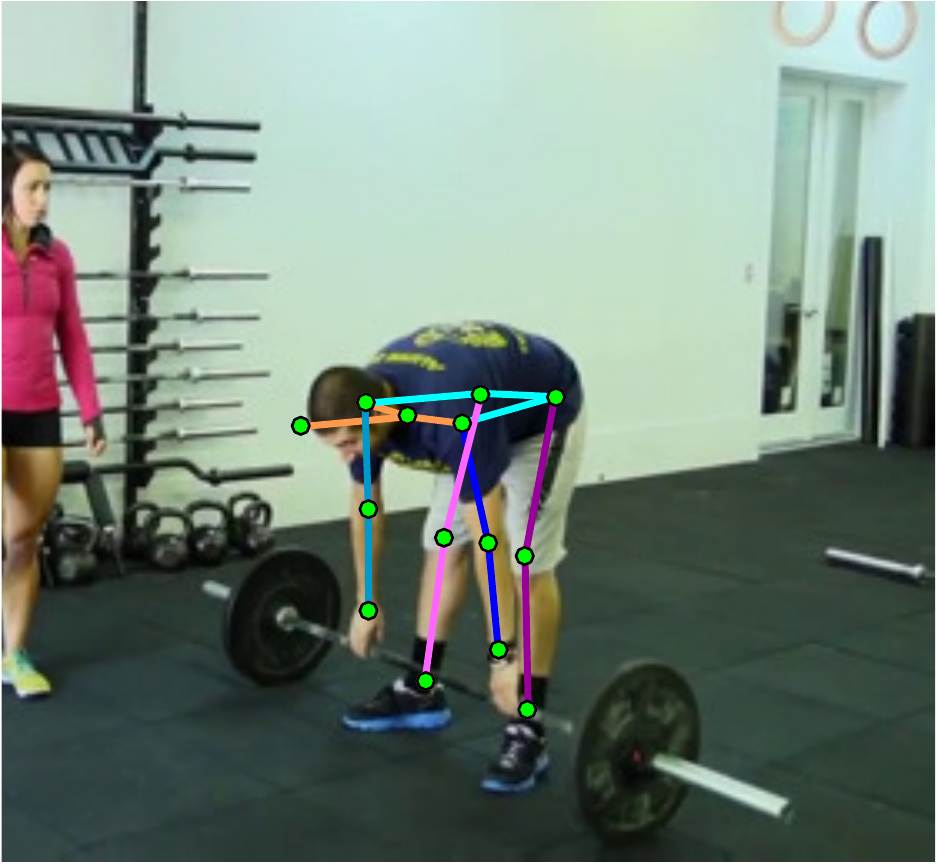} &
\includegraphics[height=0.15\linewidth]{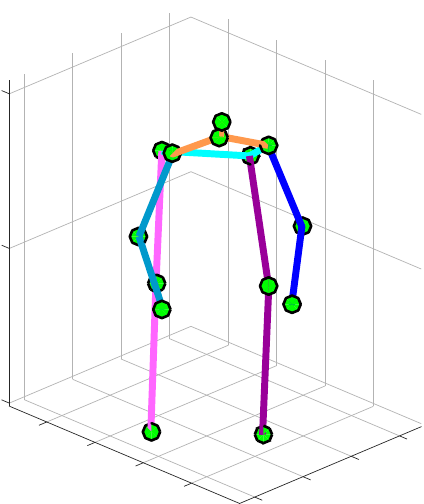} &
\includegraphics[height=0.15\linewidth]{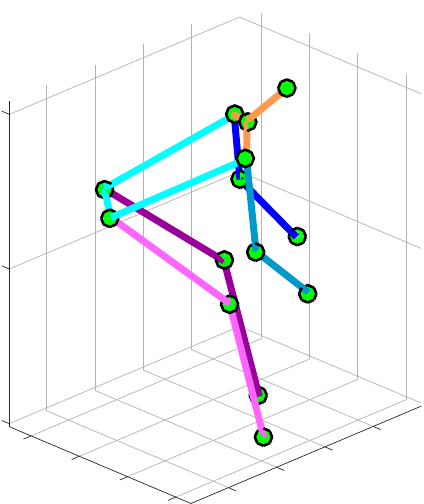} \\ 
\includegraphics[height=0.13\linewidth]{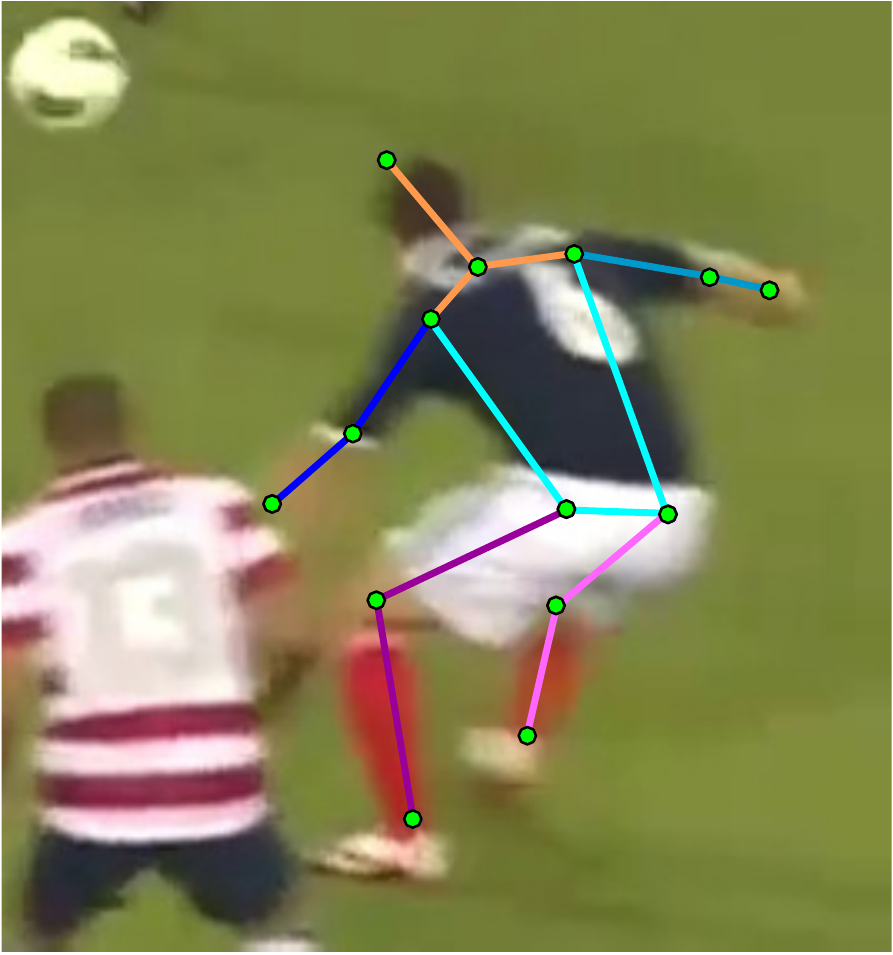} &
\includegraphics[height=0.15\linewidth]{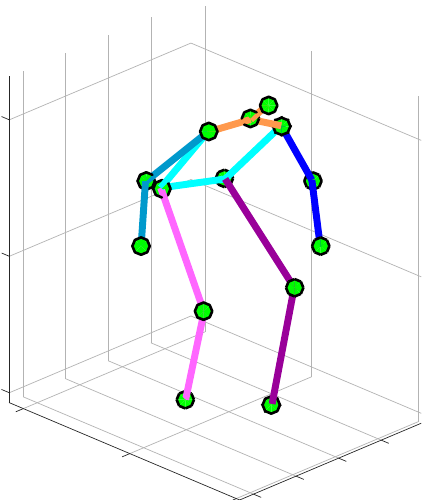} &
\includegraphics[height=0.15\linewidth]{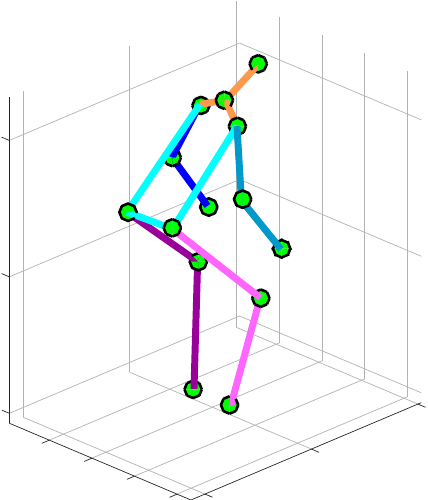} &
\includegraphics[height=0.13\linewidth]{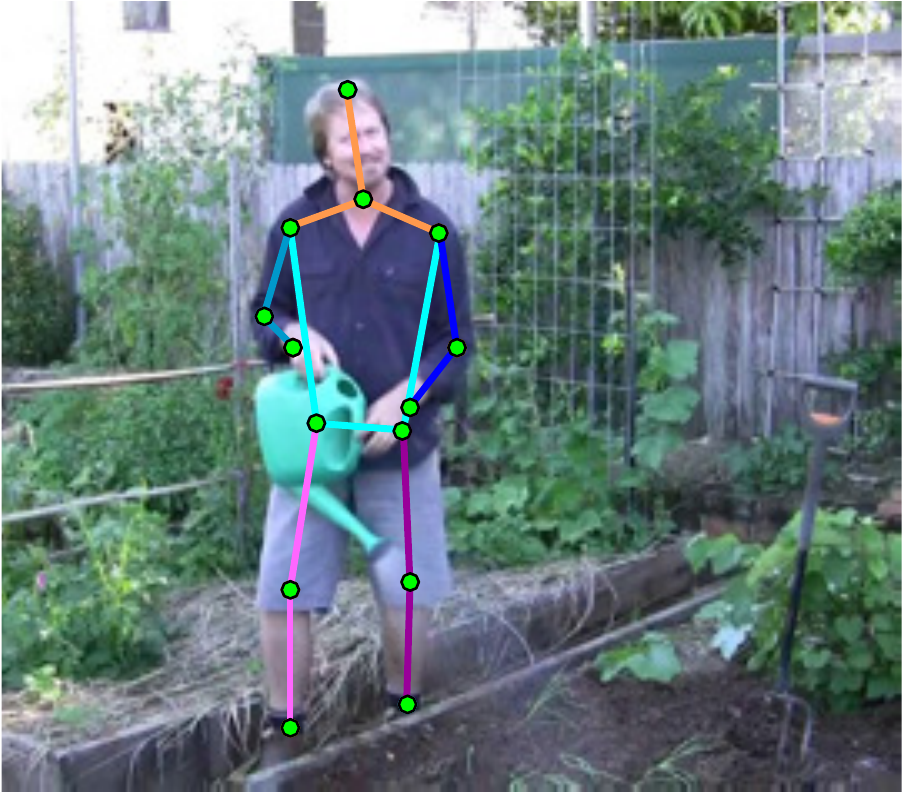} &
\includegraphics[height=0.15\linewidth]{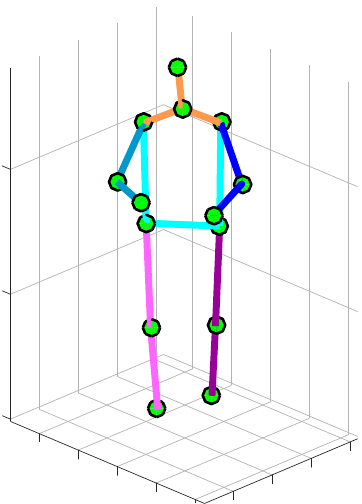} &
\includegraphics[height=0.15\linewidth]{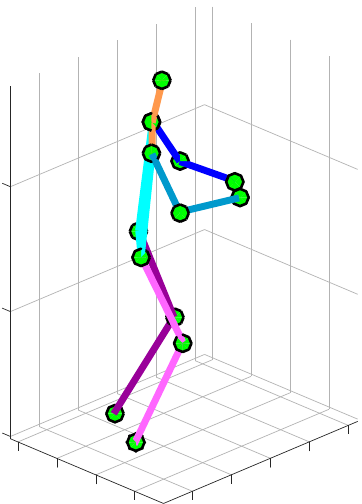} \\ 
\includegraphics[height=0.13\linewidth]{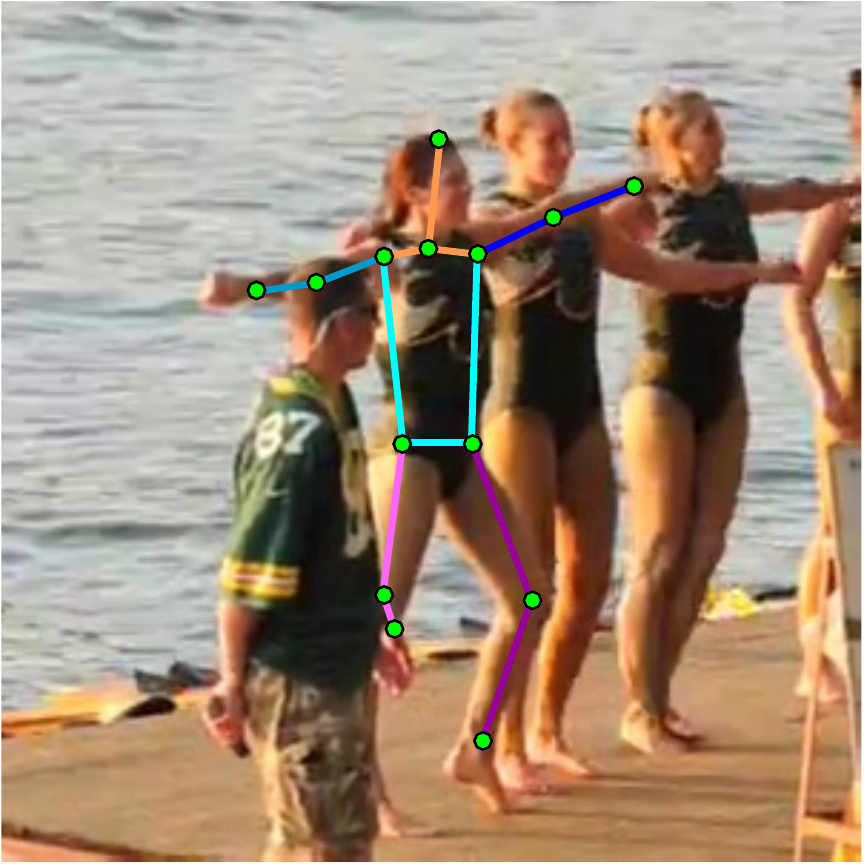} &
\includegraphics[height=0.15\linewidth]{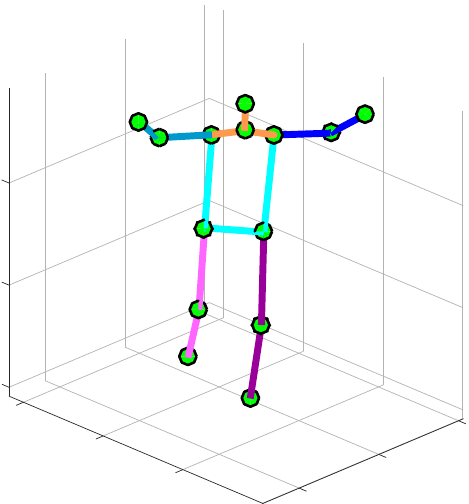} &
\includegraphics[height=0.15\linewidth]{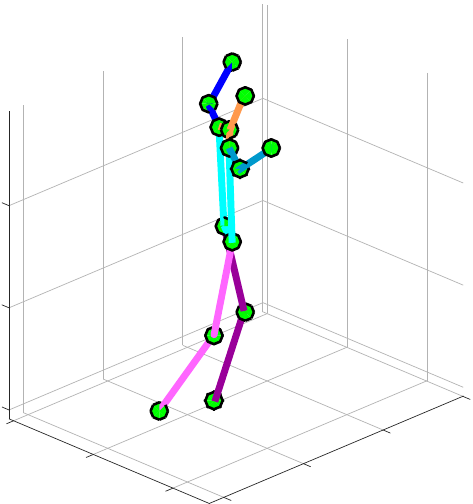} &
\includegraphics[height=0.13\linewidth]{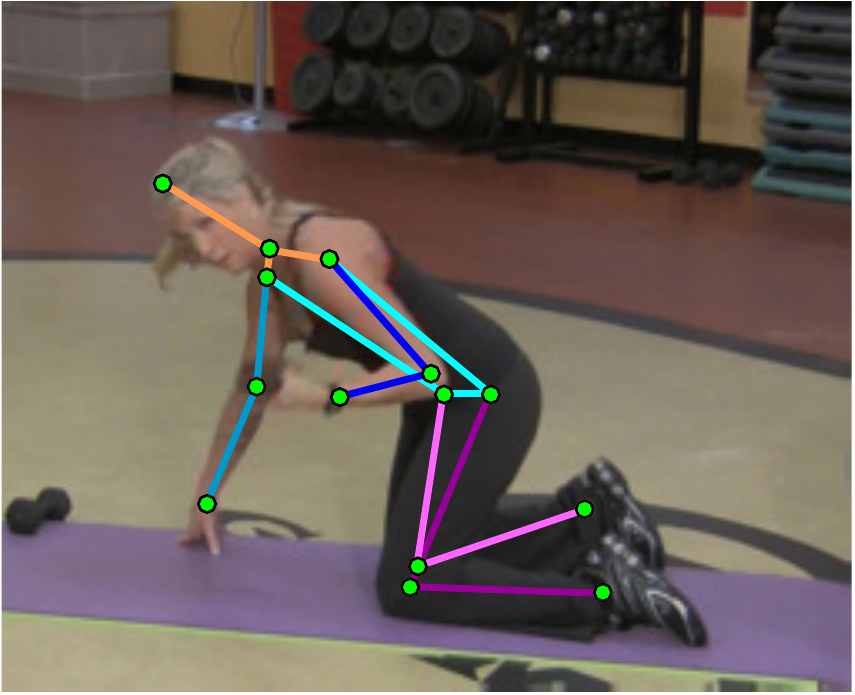} &
\includegraphics[height=0.15\linewidth]{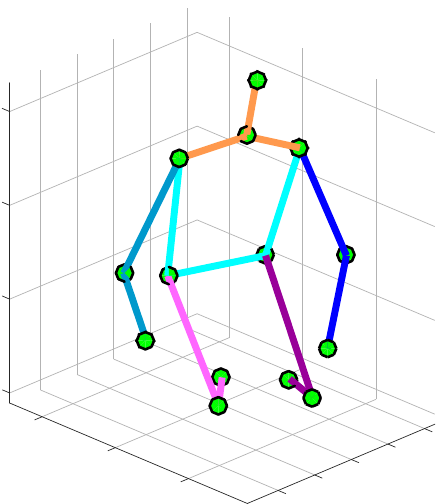} &
\includegraphics[height=0.15\linewidth]{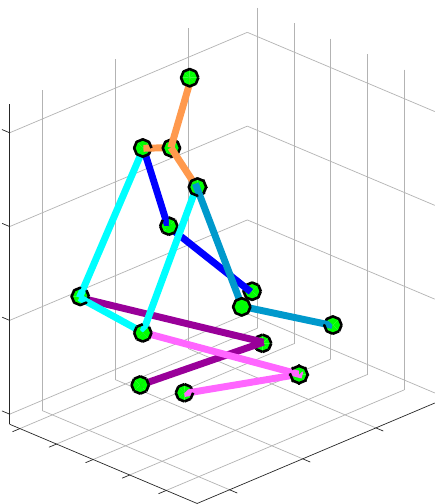} \\ 
\end{tabular}
}
\vspace{1em}
\caption{Some qualitative results from the MPII Human Pose Dataset.}
\label{fig:mpii_qualitative_results}
\end{figure*}

Finally, we present qualitative results for a few realistic images taken from the MPII Human Pose dataset \cite{andriluka14cvpr} in Fig.~\ref{fig:mpii_qualitative_results}. The results show that the proposed approach generalizes very well to complex unconstrained images. 

\section{Conclusion}\label{sec:con}
In this work, we have proposed a novel dual-source method for 3D human pose estimation from monocular images. 
The first source is a MoCap dataset with 3D poses and the other source are images with annotated 2D poses. Due to the separation of the two sources, our approach needs less supervision compared to approaches that are trained from images annotated with 3D poses, which is difficult to acquire under real conditions. The proposed approach therefore presents an important step towards accurate 3D pose estimation in unconstrained images. Compared to the preliminary work, the proposed approach achieves a substantial lower pose estimation error. This is achieved by utilizing the strengths of recent 2D pose estimation methods and combining them with an efficient and 
robust method for 3D pose retrieval. 
We have performed a thorough experimental evaluation and demonstrated that our approach achieves
competitive results in comparison to the state-of-the-art, even when the training data are from very different sources.



\ifCLASSOPTIONcaptionsoff
  \newpage
\fi



%

\bibliographystyle{ieee}
\bibliography{pose3d}
%

\begin{IEEEbiography}
[{\includegraphics[width=1in,height=1.25in,clip,keepaspectratio]{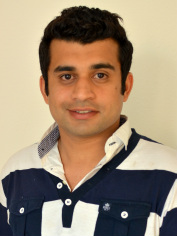}}]{Umar Iqbal}
is a PhD candidate with the Computer Vision Research Group at University of Bonn, Germany. He
received his MS degree in Signal Processing from Tampere University of Technology, Finland, in 2013. He
worked as computer vision research assistant at Computer
Vision Group, COMSATS Institute of Information Technology, Pakistan (2010-2011), Nokia Research Center, Finland
(2011-2013), and Tampere University of Technology, Finland (2013). His research interests include human pose estimation,
activity recognition, multi-target tracking, and person re-identification.
\end{IEEEbiography}

\begin{IEEEbiography}
[{\includegraphics[width=1in,height=1.25in,clip,keepaspectratio]{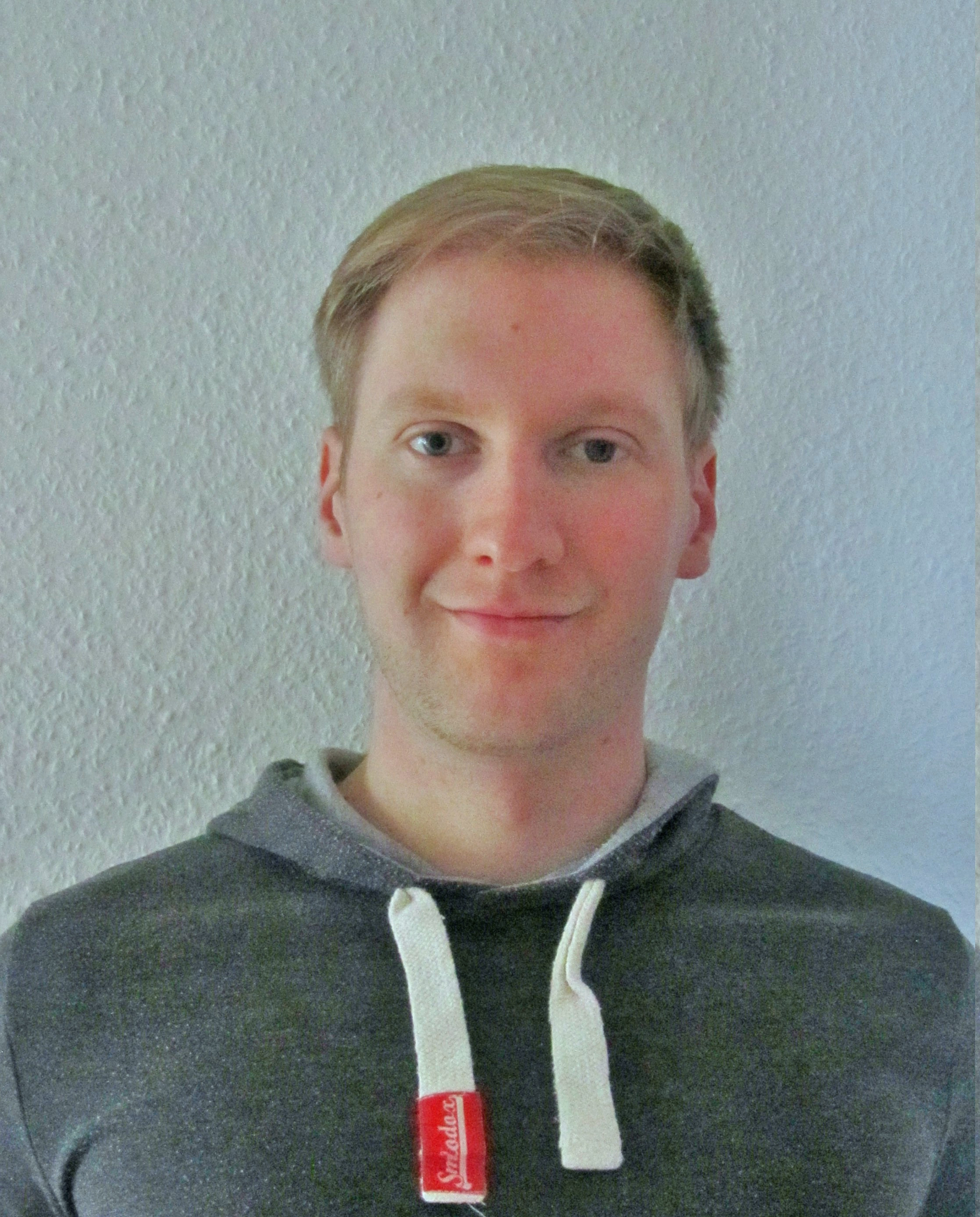}}]
{Andreas Doering} received his B.Sc. degree in computer science from the University of Bonn, Germany, 
in 2015. Currently, he is a master's student and a research assistant at the Computer Vision Group at the University of Bonn. 
His research interests include: human pose estimation, object detection, scene understanding and machine learning.

\end{IEEEbiography}

\begin{IEEEbiography}
[{\includegraphics[width=1in,height=1.25in,clip,keepaspectratio]{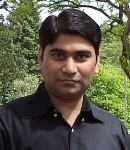}}]
{Hashim Yasin} received his MS and PhD degrees from the Department of Computer Science, University of Bonn, Germany, 
in 2012 and 2016 respectively. Currently, he is an assistant professor at the 
National University of Computer \& Emerging Sciences, Pakistan. His research interests include
vision-based 3D motion retrieval and reconstruction, 3D pose estimation, motion synthesis and analysis etc.
\end{IEEEbiography}

\begin{IEEEbiography}[{\includegraphics[width=1in,height=1.25in,clip,keepaspectratio]{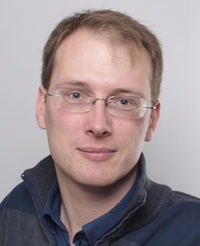}}]{Bj\"orn Kr\"uger}
studied computer science, mathematics and physics at Bonn university. He 
received his MS in computer science (Dipl.-Inform.) in 2006 and his PhD 
(Dr. rer. nat.) in computer science in 2012. From 2012 to 2015 he worked 
as postdoc at Bonn university. Since 2015 he joined the Gokhale Method 
Institute (Stanford, CA) as senior researcher.
His research interests include: computer animation, computer graphics, 
machine learning, and motion capture.
\end{IEEEbiography}

\begin{IEEEbiography}
[{\includegraphics[width=1in,height=1.25in,clip,keepaspectratio]{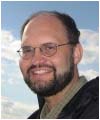}}]
{Andreas Weber} studied mathematics and computer
science at the Universities of Tubingen, Germany and Boulder, Colorado, U.S.A. From
the University of Tubingen he received his MS in Mathematics (Dipl.-Math) in 1990 and his PhD
(Dr. rer. nat.) in computer science in 1993. From 1995 to 1997 he was working with a scholarship
from Deutsche Forschungsgemeinschaft as a postdoctoral fellow at the Computer Science 
Department of Cornell University. From 1997 to 1999 he was a member of the Symbolic Computation
Group at the University of Tubingen, Germany. From 1999 to 2001 he was a member of the research group Animation and Image
Communication at the Fraunhofer Institut for Computer Graphics. Since 2001 he has been professor at the University of Bonn and head of the Multimedia, Simulation and Virtual Reality Group. 
\end{IEEEbiography}


\begin{IEEEbiography}
[{\includegraphics[width=1in,height=1.25in,clip,keepaspectratio]{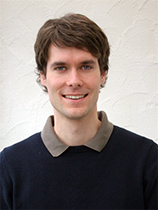}}]{Juergen Gall}
Juergen Gall obtained his B.Sc. and his Masters degree in
mathematics from the University of Wales Swansea (2004)
and from the University of Mannheim (2005). In 2009,
he obtained a Ph.D. in computer science from the Saarland
University and the Max Planck Institut f{\"u}r Informatik. He
was a postdoctoral researcher at the Computer Vision Laboratory, ETH Zurich,
from 2009 until 2012 and senior research scientist at the Max Planck Institute for Intelligent
Systems in T{\"u}bingen from 2012 until 2013. Since 2013,
he is professor at the University of Bonn and head of the
Computer Vision Group. 
\end{IEEEbiography}


\vfill


\end{document}